\documentclass[preprint,12pt]{elsarticle}
\usepackage{subfigure}
\usepackage{makeidx}  % allows for indexgeneration
\usepackage{multirow}
\usepackage{tabularx}
\usepackage{amsmath,amsfonts,amssymb}
\usepackage{multirow,array}
\usepackage{graphicx}
\usepackage{rotating}
\usepackage{MnSymbol,wasysym}
\usepackage{url,soul}
\usepackage[switch]{lineno}
\usepackage{blindtext}
\usepackage{comment}
\usepackage{enumitem}
\usepackage{booktabs}
\usepackage{bm}
\usepackage{url}
\usepackage{makecell}
\usepackage{color}
\usepackage[linesnumbered,ruled,vlined]{algorithm2e}
\usepackage{tikz}

\usepackage{array}
\newcommand{\PreserveBackslash}[1]{\let\temp=\\#1\let\\=\temp}

\newcolumntype{P}[1]{>{\centering\arraybackslash}p{#1}}
\newcolumntype{M}[1]{>{\arraybackslash}m{#1}}
\newcolumntype{L}[1]{>{\centering\arraybackslash}l{#1}}

\newcommand{\tikzrectangle}[2][black,fill=red]{\tikz[]\draw[#1] (0,0) rectangle ++(0.3,0.3);}

\definecolor{myred}{RGB}{251, 128, 114}
\definecolor{myblu}{RGB}{128, 177, 211}
\definecolor{myora}{RGB}{253, 180, 98}
\definecolor{mypink}{RGB}{252, 205, 229}

\begin{document}
\begin{frontmatter}
\title{Exploring the Trade-off between Plausibility, Change Intensity and Adversarial Power in Counterfactual Explanations using Multi-objective Optimization}

\author[a,b]{Javier Del Ser\corref{cor1}}
\ead{javier.delser@tecnalia.com}
\author[a]{Alejandro Barredo-Arrieta}
\author[c]{Natalia D\'iaz-Rodr\'iguez}
\author[c]{Francisco Herrera}
\author[d,e]{and Andreas Holzinger}

\address[a]{TECNALIA, Basque Research and Technology Alliance (BRTA),\\P. Tecnologico, Ed. 700, 48160 Derio, Spain}
\address[b]{Department of Communications Engineering,\\University of the Basque Country (UPV/EHU), 48013 Bilbao, Spain}
\address[c]{DaSCI Andalusian Institute of Data Science and Computational Intelligence,\\University of Granada, 18071 Granada, Spain}
\address[d]{Institute for Medical Informatics, Statistics and Documentation,\\Medical University of Graz, 8036 Graz, Austria}
\address[e]{xAI Lab, Alberta Machine Intelligence Institute,\\University of Alberta, T5J 3B1 Edmonton, Canada}
\cortext[cor1]{Corresponding author. TECNALIA, Basque Research \& Technology Alliance (BRTA), P. Tecnologico, Ed. 700. 48170 Derio (Bizkaia), Spain.}

\begin{abstract}
There is a broad consensus on the importance of deep learning models in tasks involving complex data. Often, an adequate understanding of these models is required when focusing on the transparency of decisions in human-critical applications. Besides other explainability techniques, trustworthiness can be achieved by using counterfactuals, like the way a human becomes familiar with an unknown process: by understanding the hypothetical circumstances under which the output changes. In this work we argue that automated counterfactual generation should regard several aspects of the produced adversarial instances, not only their adversarial capability. To this end, we present a novel framework for the generation of counterfactual examples which formulates its goal as a multi-objective optimization problem balancing three different objectives: 1) plausibility, i.e., the likeliness of the counterfactual of being possible as per the distribution of the input data; 2) intensity of the changes to the original input; and 3) adversarial power, namely, the variability of the model's output induced by the counterfactual. The framework departs from a target model to be audited and uses a Generative Adversarial Network to model the distribution of input data, together with a multi-objective solver for the discovery of counterfactuals balancing among these objectives. The utility of the framework is showcased over six classification tasks comprising image and three-dimensional data. The experiments verify that the framework unveils counterfactuals that comply with intuition, increasing the trustworthiness of the user, and leading to further insights, such as the detection of bias and data misrepresentation.
\end{abstract}

\end{frontmatter}

%\linenumbers

\section{Introduction} \label{sec:intro}

During the last years Deep Neural Networks (also referred to as \emph{Deep Learning}) have started traversing beyond the theoretical analysis of their properties towards being implemented and deployed in a multitude of real-world applications. This has been specially noted in applications dealing with high-dimensional data, over which Deep Learning has delivered promising results to conquer the broad landscape of Machine Learning modeling approaches. As such, their superior performance has been noted in many scenarios dealing with image, video and/or spatial-temporal data, including agriculture \cite{kamilaris2018deep}, transportation \cite{del2019bioinspired} or industry \cite{diez2019data}, to cite a few. Nowadays the prevalence of Deep Learning in such areas is beyond any doubt. 

Unfortunately, some concerns arise from the mismatch between research studies dealing with Deep Learning applied to certain modeling tasks (\textit{let the model perform to its best for the task at hand}) and the real-world use of models to improve an already known solution. Most in-field approaches contemplate attempts at improving an already human-created solution to solve a problem (optimizing a process), whereas the most common Deep Learning approaches are better suited to find their own solutions to a more high-level problem (predicting an outcome). Together with this difference, another concern deals with the difficulty to understand and interpret the mechanisms by which Deep Learning works, particularly when the audience that makes decisions on their outputs lacks any knowledge about Computer Science. This renders Deep Learning a useless modeling choice for real-world scenarios in which models are used to improve decision making in processes that are managed by humans and/or where decisions affect humans. This is the case observed in life sciences \cite{HOLZINGER2022263}, law or social policy making \cite{doshi2017accountability}, among others. In other words, actionability of predictions requires a step beyond a proven good generalization performance of the model issuing them. 

In order to bridge this gap in Machine Learning-based decision making, new frameworks for explainability are required. These frameworks aim at giving insights not only to experts in the field of application, but also to those commonly in charge of the use and maintenance of the deployed models. These two audience profiles differ significantly in what refers to their capabilities to understand explanations generated for a given model. These different capabilities entail that approaches to explain Deep Learning models generate explanations better suited for auditing the models by developers, leaving them far from the cognitive requirements of experts that ultimately make decisions in practice. 

Recent research is profoundly bothered with bridging this gap. To this end, the broad scope of model explainability has been approached from manifold areas, including robustness by adversarial attacks \cite{goodfellow2014generative, arjovsky2017wasserstein}, output confidence estimation \cite{papernot2018deep,pmlr-v48-gal16,1707.07013}, visualization of internal representation \cite{zeiler2010deconvolutional, simonyan2013deep} or attention-based explanations \cite{bach2015pixel}. Even though the research community is thrilling with new advances in explainability, they do not entirely bridge the aforementioned gap between theoretical developments and their practical adoption. Most explainability solutions \cite{arrieta2019explainable} consider an audience with profound knowledge of the inner workings of the models, which eases the understanding of explanations, but does not comply with real-world settings often encountered in model-based decision making processes.

Among the alternatives to reach such a universal understanding of model explanations, counterfactual examples is arguably the one that best conforms to human understanding principles when faced with unknown phenomena. The underlying concept in our paper is human counterfactual thinking, which describes a set of possible alternatives to events that have already occurred, but which contradict the actual events \cite{roese1997counterfactual}. Indeed, discerning what would happen should the initial conditions differ in a plausible fashion is a mechanism often adopted by human when attempting at understanding any unknown \cite{byrne2002mental,van2015cognitive}. Circumscribing the factual boundaries by which a given model works \emph{as usual} can be conceived as a post-hoc explainability method, which grounds on an adversarial analysis of the audited model \cite{stepin2021survey}. From the practical perspective several aspects of the produced counterfactual examples should be considered besides its plausibility, so that the audience of the model can examine the limits of the model from a multi-faceted perspective. It is only by investigating this manifold interplay between the features of the generated counterfactuals when a well-rounded counterfactual analysis can be achieved.

This manuscript joins the research area aimed at making Deep Learning models more usable in practice via counterfactual-based explanations. To this end, we propose an adversarial strategy to produce counterfactual examples for a Deep Learning classifier. This classifier to be audited solves a task defined over a certain dataset (e.g. discriminate male and female images from human faces), so that counterfactuals are generated to explain the boundaries of the model once trained to address the classification task at hand. We further impose that the generated counterfactual examples are \emph{plausible}, i.e., changes made on the input to the classification model have an appearance of credibility without any computer intervention. To ensure plausibility, the proposed method makes use of GANs (Generative Adversarial Networks) in order to learn the underlying probability distribution of each of the features needed to create examples of a target distribution (namely, human faces). Our framework allows searching among samples of the first distribution to find realistic counterfactuals close to a given test sample that could be misclassified by the model (namely, a face of a male being classified as a female). As a result, our framework makes the user of the model assess its limits with an adversarial analysis of the probability distribution learned by the model, yet maintaining a sufficient level of plausibility for the analysis to be understood by a non-expert user. As a step beyond the state of the art, the proposed framework ensures the production of multi-faceted counterfactual examples by accounting for two additional objectives besides plausibility: 1) the \emph{intensity of the modification} made to a original example to produce its counterfactual version; and 2) its \emph{adversarial power}, which stands for the change in the output of the model that is audited. 

In summary, the main contributions of this work beyond our preliminary findings reported in \cite{barredo2020plausible} can be summarized as follows:
\begin{itemize}[leftmargin=*]
    \item We present a novel framework to generate multi-faceted counterfactual explanations for targeted classification models. The framework brings together GAN architectures for generative data modeling and multi-objective optimization for properly balancing among conflicting properties sought for the counterfactuals: plausibility, change intensity and adversarial power.
    \item The framework is described mathematically, and design rationale for each of its compounding blocks is given and justified.
    \item Explanations generated by the framework are showcased for several classifiers and GAN models for image and volumetric data, discussing on the trade-off between the properties of the counterfactual set. 
    \item We argue and show that when inspected from a multi-faceted perspective, counterfactual examples can be a magnificent tool for bias analysis and the discovery of misrepresentations in the data space.
\end{itemize}

The rest of the article is organized as follows: first, Section \ref{sec:background} covers background required for connecting the four core aspects of our proposed framework: Deep Learning for image classification, GANs, model explainability and counterfactual explanations. Section \ref{sec:framework} details the framework proposed in this study, including a mathematical statement of the problem tackled via multi-objective optimization and a discussion on how the output of the framework can be consumed by different audiences. Section \ref{sec:experiments} describes the experimental setup designed to showcase the output of the framework. Section \ref{sec:discussion} presents and discusses the results stemming from the performed experiments. Finally Section \ref{sec:conclusion} draws conclusions and future research lines related to our findings. {\color{black}Table \ref{tb:notation} summarizes the main mathematical symbols used throughout the manuscript.}
\begin{table}[h!]
	\centering
	\caption{{\color{black}Summary of symbols, meaning and their first appearance in the manuscript.}}
	\label{tb:notation}
	\vspace{2mm}
	\resizebox{\columnwidth}{!}{{\color{black}\begin{tabular}{ccp{15cm}}
		\toprule
		\multicolumn{1}{c}{Symbol} & \multicolumn{1}{c}{Appearance} & \multicolumn{1}{c}{Meaning} \\
		\midrule
		$\mathbf{x}$ & Section 3.2 & Input example to the audited model \\
		$P_\mathbf{X}(\mathbf{x})$ & Section 3.2 & Input data distribution followed by $\mathbf{x}$ \\
		$\mathbf{a}$ & Section 3.2 & Original attribute vector \\
		$\mathbf{b}$ & Section 3.2 & Modified attribute vector \\
		$\bm{\delta}$ & Section 3.2 & Perturbation vector ($\bm{\delta}=\mathbf{b}-\mathbf{a}$) \\
		$\mathbf{x}^{\mathbf{b},\prime}$ & Section 3.2 & Generated counterfactual for input $\mathbf{x}^\mathbf{a}$ and attribute vector $\mathbf{b}$ \\
		$\widehat{\mathbf{b}}$ & Section 3.2 & Modified attribute vector predicted by $C(\cdot)$\\
		$\oplus$ & Section 3.2 & Superscript defining the anchor sample for which a counterfactual is produced \\
		$C(\cdot)$ & Section 3.2, Fig. 2 & Classifier that predicts the attribute vector of its input query \\
		$D(\cdot)$ & Section 3.2, Fig. 2 & Discriminator module of a GAN \\
		$G_{enc}(\cdot)$ & Section 3.2, Fig. 2 & Encoder of a GAN generator module \\
		$G_{dec}(\cdot)$ & Section 3.2, Fig. 2 & Decoder of a GAN generator module \\
		$T(\cdot)$ & Section 3.2, Fig. 2 & Target model to be audited \\
		$\mathcal{L}_{rec}(\mathbf{x},\mathbf{x}^{\mathbf{a}\prime})$ & Eqs. (1) and (2) & Reconstruction loss\\
		\smash{$\mathcal{L}_{att}^G(\mathbf{b},\widehat{\mathbf{b}}^\prime)$} & Eqs. (1) and (3) & Attribute loss\\
		$\mathcal{L}_{adv}^G(\mathbf{x}^{\mathbf{b}\prime})$ & Eqs. (1) and (4) & Adversarial loss\\
		\multirow{2}{*}{$\lambda_i$} & \multirow{2}{*}{Eqs. (1) and (5)} & Weights of reconstruction ($i=1$) and attribute terms ($i=2,3$) in the training losses of $G_{enc}(\cdot)$, $G_{dec}(\cdot)$, $D(\cdot)$ and $C(\cdot)$ \\
		$f_{att}(\cdot)$ & Eq. (11) & Function quantifying the \emph{change intensity} of the generated counterfactual \\
		$f_{gan}(\cdot)$ & Eq. (11) & Function quantifying the \emph{plausibility} of the generated counterfactual\\
		$f_{adv}(\cdot)$ & Eq. (11) & Function quantifying the \emph{adversarial power} of the generated counterfactual\\
		\bottomrule
	\end{tabular}}}
\end{table}

\section{Background}\label{sec:background}

As anticipated in the introduction, the proposed framework resorts to GANs for producing realistic counterfactual examples of classification models. Since the ultimate goal is to favor the understanding of the model classification boundaries by an average user, the framework falls within the XAI (Explainable Artificial Intelligence) umbrella. This section briefly contextualizes and revisits the state of the art of the research areas related to the framework: Deep Learning for image classification and generative modeling (Subsection \ref{ssec:image_class_gans}), XAI and counterfactual analysis (Subsection \ref{ssec:xai_counterfactuals}) and multi-objective optimization (Subsection \ref{ssec:multiobjective_optimization}).

\subsection{Deep Learning for Image Classification and Generative Modeling} \label{ssec:image_class_gans}

When it comes to classification tasks over image data, the reportedly superior modeling capabilities of Convolutional Neural Networks (CNNs) are often adopted to capture spatial correlations in image data \cite{russakovsky2015imagenet,krizhevsky2012imagenet}. This is achieved by virtue of trainable convolutional filters which can be trained via gradient backpropagation or even imported from other networks pretrained for similar tasks, giving rise to image classification models of the highest performance. The increasing availability of image datasets and the capability of processing them efficiently have yielded hierarchically stacked CNNs that, despite attaining unprecedented levels of accuracy, come at the cost of more complex, less understandable model structures \cite{lipton2018mythos}. The more complex the model is, the harder is to pinpoint the reasons for its failures. The need for auditing these black-boxes is the core motivation of the study presented in this paper.

Another task for which CNNs are crucial is generative modeling, e.g. the construction of models capable of characterizing the distribution of a given dataset and sampling it to create new, synthetic data instances. When the dataset is composed by images, generative adversarial networks (GANs) are arguably the spearhead in image generative modeling. GANs were first introduced by Goodfellow in \cite{goodfellow2014generative}, bringing the possibility of using neural networks (\emph{function approximators}) to become generators of a desired distribution. Since their inception, GANs have progressively achieved photo-realistic levels of resolution and quality when synthesizing images of different kinds. In general, a GAN architecture consists of two data-based models, which are trained in a mini-max game: one of the players (models) minimizes its error (loss), whereas the other maximizes its gain. In such a setup, multiple models have flourished to date, each governed by its strengths and vulnerabilities \cite{hindupur2017gan}. In connection to the scope of this paper, some of these were conceived with the intention of finding the pitfalls of a certain model and the ways to confuse it \cite{goodfellow2014generative, arjovsky2017wasserstein}. Other GAN approaches aim at generating samples of incredibly complex distributions like photo-realistic human faces \cite{zhang2017stackgan, wu2019gp}. 

As will be later detailed in Section \ref{sec:framework}, the framework proposed in this work hybridizes these two uses of CNNs by optimizing the output generated by a GAN to perform a counterfactual analysis of a given classification model to be audited.

\subsection{Explainable Artificial Intelligence (XAI) and Counterfactual Explanations} \label{ssec:xai_counterfactuals}

Model explainability \cite{arrieta2019explainable} and causability \cite{holzinger2021towards} have recently become topics of capital importance in Machine Learning, giving rise to a plethora of different approaches aimed to explain how decisions are issued by a given model. Most research activity in this area is arguably focused on post-hoc XAI tools that produce explanations for single data instances (what is referred to as \emph{local explanations}). The LIME tool presented in \cite{ribeiro2016should} is one of this kind, visualizing a model's internal activations when processing a given test sample. A similar approach is followed by LRP (Layer-wise Relevance Propagation) embedded in the SHAP suite \cite{lundberg2017unified}, which highlights the parts of an input image that push the output of the model towards one label or another. This provides an understandable interface of the reasons why the model produces its decision. More recently, Grad-CAM \cite{selvaraju2017grad} an its successor Grad-CAM++ \cite{chattopadhay2018grad} can be considered as the \emph{de facto} standard for the explainability of local decisions, particularly in the field of image classification. These two methods implement a gradient-based inspection of the knowledge captured by a neural network, giving rise to a quantitative measure of the importance of parts of the image for the output of the model. Unfortunately, the dependence of such explanations on the gradient of the model restricts the applicability of these techniques to other techniques beyond neural architectures. 

When pursuing model-agnostic local explanations, a common strategy is to analyze the model from a counterfactual perspective. Counterfactual exploration is an innate process for the human being when facing an unknown phenomenon, system or process. The concept behind counterfactual explanations reduces to providing an informed answer to a simple question: \emph{which changes would make the output of the unknown model for a given input vary?}. Such changes constitute a counterfactual example, always related to an input to the process or system under focus. Based on this concept, many contributions have hitherto developed different XAI approaches to generate counterfactual examples that allow understanding how Machine Learning models behave. Some approaches are based on discovering the ability of a given individual to change the model's outcome. One example is the work in \cite{wachter2017counterfactual}, which presents a simple but effective distance-based counterfactual generation approach, that can be used to audit different classifiers (e.g. neural networks and support vector machines). Later, the counterfactual problem is tackled in \cite{ustun2019actionable} departing from the premise that a user should be able to change a model's outcome by actionable variables. This hypothesis is validated over linear classifiers, but also claimed to be extensible to non-linear classifiers by means of local approximations. In a similar fashion, \cite{karimi2020model, poyiadzi2020face} allow the user to guide the generation of counterfactual examples by imposing forbidden changes that cannot be performed along the process. A subset of counterfactual studies are rather focused on the problem of predictive multiplicity \cite{breiman2001statistical, marx2020predictive}. Multiple classifiers may output the same solution while treating the data in different ways, hence the generation of counterfactuals can lead to insights into the question of which of these classifiers is better for the problem at hand. In this research area several contributions \cite{pawelczyk2020counterfactual, rawal2020can, chen2020strategic, kasirzadeh2021use} have developed different schemes to address this problem. Connectedness, proximity, plausibility, stability and robustness are yet other concerns that have pushed the development of techniques for the generation of counterfactuals. In their search for robust interpretability, the work in \cite{alvarez2018towards} came up with a method to generate self-explaining models based on explicitness, faithfulness and stability.

Following the extensive analysis carried out in \cite{chou2021counterfactuals}, it is of utmost importance to recall the \textit{``master theoretical algorithm''} \cite{hasperue2015master}, from which nineteen other algorithms concerning counterfactual explanations can be derived. The nineteen algorithms fall into a categorization of six different counterfactual generation strategies: instance-based, constraint-based, genetic-based, regression-based, game theory-based and case reasoning-based. Instance based approaches are derived from \cite{wachter2017counterfactual,lewis2013counterfactuals}, based on feature perturbance measured by a distance metric. The pitfall of these approaches (when pure) resides on their inability to validate instance plausibility. Constraint-based approaches are, in turn, the methods that modulate their counterfactual search by means of a constraint satisfaction problem. The more general scope of these approaches allows for an easier adaptation to the problems at hand. Genetic-based approaches, as the name conveys, guide the search for counterfactuals as a genetic-oriented optimization problem. Regression-based approaches use the weights of a regression model as a proxy to produce counterfactual examples. However, these approaches again fall short at assuring the plausibility and diversity of the produced counterfactual instances. Game-theory based approaches are driven by game-theoretical principles (e.g. Shapley values), but also disregard important properties of its counterfactual outputs. Finally, case reasoning-based approaches seek past solutions (in the model) that are close to a given instance, and adapt them to produce the counterfactual. Once again, such adaptations may produce counterfactual instances that, even if close to a certain input, cannot be claimed to be plausible nor diverse with respect to the input under consideration. 

\subsection{Multi-objective Optimization} \label{ssec:multiobjective_optimization}

\begin{figure}[!ht]
	\vspace{-1mm}
	\centering
	\includegraphics[width=\columnwidth]{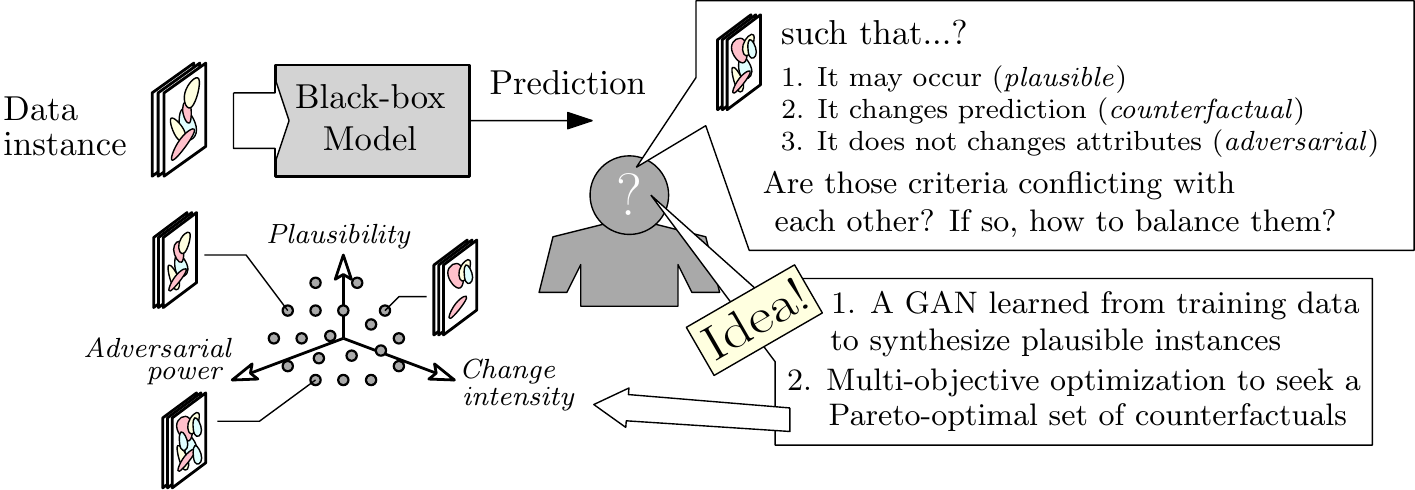}
	\caption{Conceptual representation of the rationale behind the confluence of predictive modeling, generative adversarial learning, explainability and multi/objective optimization that lies at the core of the proposed framework.}
	\label{fig:concept}
	\vspace{-1mm}
\end{figure}

From the previous section it can be inferred that the generation of counterfactual explanations can be mathematically stated as a multi-objective optimization problem comprising different objectives that can be conflicting with each other. Plausibility -- i.e., the likelihood of the counterfactual example to occur in practice -- can be thought of conflicting with the amplitude of the modifications made to the input of the model. Likewise, intense changes in the output of the audited model (namely, its \emph{adversarial power} as introduced in Section \ref{sec:intro}) when fed with the counterfactual example can jeopardize its plausibility. There lies the contribution of the framework proposed in this work: the generation of a portfolio of counterfactual examples to a certain input that optimally trade among these objectives. This portfolio provides richer information for the user to understand the behavior of the audited model, and distinguishes this work from the current research on counterfactual analysis. The conceptual diagram shown in Figure \ref{fig:concept} illustrates this motivational idea.

To this end, the framework presented in this work falls between constraint-based, genetic-based and instance-based counterfactual explanations, combining these three categories to render a set of multi-criteria counterfactuals. The usage of a GAN architecture presents the ability of a bounded search within a target distribution, enabling quantitative measures of the plausibility of the generated counterfactual (via the discriminator) and algorithmic means to sample this distribution (via the generator). The usage of a multi-objective optimization algorithm yields the ability to guide the counterfactual generation process as per the desired objectives (plausibility, intensity of the modifications and adversarial power), giving rise to the aforementioned portfolio of multi-criteria counterfactual explanations. Among them, we will resort to multi-objective evolutionary algorithms \cite{coello2007evolutionary,zhou2011multiobjective}, which efficiently perform the search for Pareto front approximations of optimization problems comprising multiple objectives, without requiring information about their derivatives whatsoever. 

\section{Proposed Multi-criteria Counterfactual Generation Framework}\label{sec:framework}

This section covers the proposed framework, including the intuition behind its conceptual design (Subsection \ref{ssec:design}), a detailed description of its constituent parts and mathematical components underneath (Subsection \ref{ssec:structure}), and an outline of the target audiences that can consume the produced counterfactual explanations, supported by hypothetical use cases illustrating this process (Subsection \ref{ssec:audiences}).

\subsection{Design Rationale} \label{ssec:design}

The explainability framework explores the weaknesses of a target model by means of counterfactual instances generated by a GAN architecture. One of the key aspects of this framework is that it focuses on discovering the reality-bound weaknesses of the target model in the form of examples that, without exiting the realm of plausibility, are able to confound the target model. For instance, for a classifier mapping human faces to their gender (\texttt{male}, \texttt{female}), the framework can generate modifications of a given input face that are still considered to be real, but they make the audited model change their predicted gender. The overarching motivation of the framework comes from the human inability to asses the working boundaries of a given model in highly-dimensional spaces. In such complex areas, such as image classification, the domain in which images are bound is complex to be characterized, thereby requiring complex generative modeling approaches capable of modelling it and drawing new samples therefrom. The generator of a GAN architecture serves for this purpose, whereas the discriminator of the GAN allows verifying whether an output produced by the generator is close to the distribution of the dataset at hand, hence giving an idea of the plausibility of the generated instance.

At this point it is worth pausing at the further insights that the GAN-based framework can provide. Modifications of an input image producing a counterfactual can be edited by changing the value of variables that affect the output of the GAN generator. Such variables can represent attributes of the input image that ease the interpretation of the results of the counterfactual study regarding the existence of miss-representations of the reality captured in the dataset at hand and transferred to the audited models. For instance, in the face-gender classifier exemplified previously, let us consider a GAN model with editable attributes (e.g. an AttGAN \cite{he2019attgan}), including color hair, face color or facial expressions. A counterfactual study of a \texttt{man} face could reveal that for the face to be classified as a \texttt{woman}, the color hair attribute of all produced counterfactuals share the same value (\emph{blonde}). Besides the inherent interpretative value of the counterfactuals themselves, our framework can also identify data biases that may have propagated and influenced the generalization capabilities of the audited model.

\subsection{Structure and Modules} \label{ssec:structure}

\begin{figure}[!ht]
	\vspace{-1mm}
	\centering
	\includegraphics[width=\columnwidth]{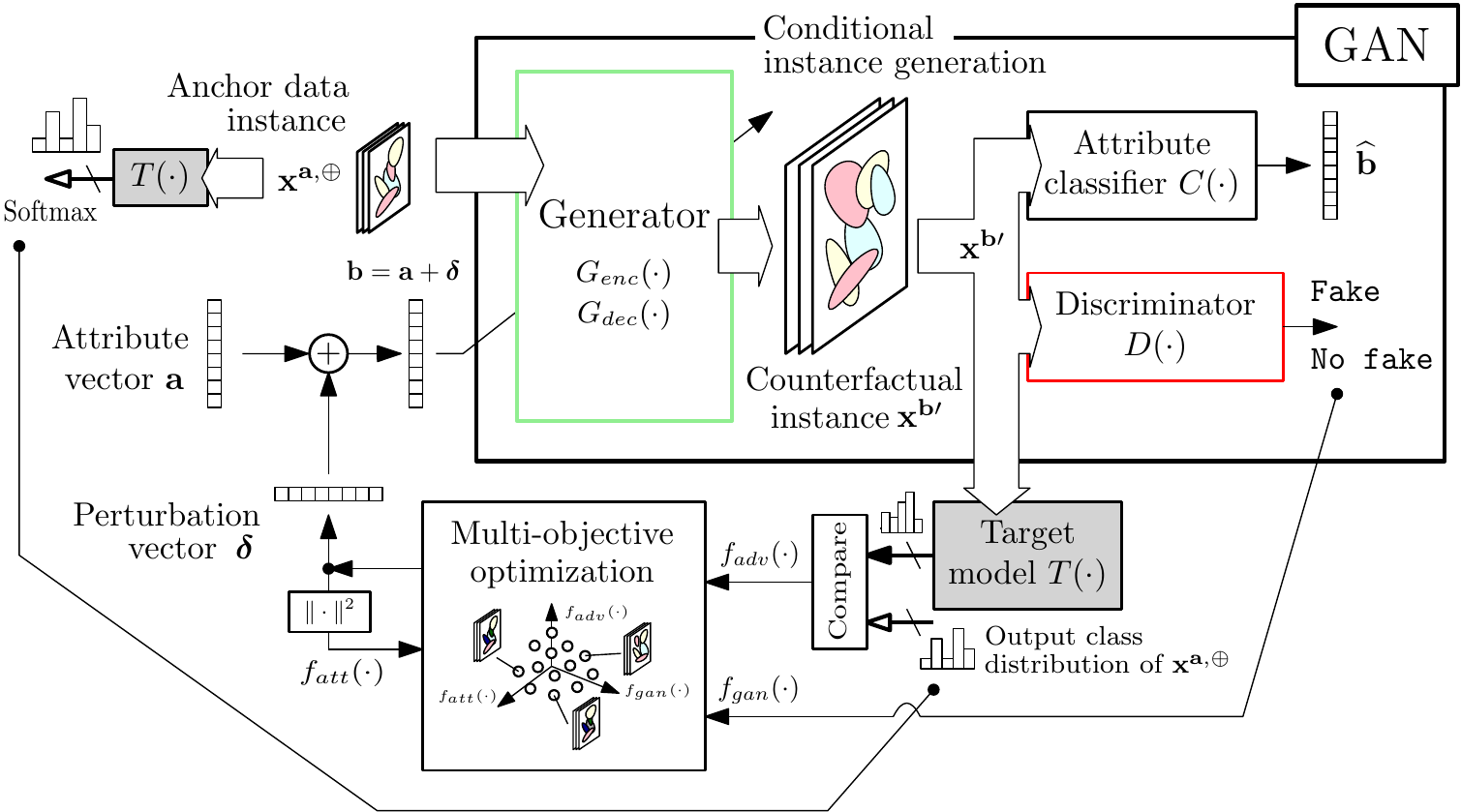}
	\caption{Block diagram of the proposed framework, which is capable of producing counterfactual instances for an audited model $T(\cdot)$ based on three criteria: plausibility, adversarial power and change intensity. {\color{black}Given an anchor data instance and generator, discriminator and an attribute classifier trained over the dataset of the task at hand, the multi-objective solver seeks the set of Pareto-optimal perturbation vectors ${\bm\delta}$ that best balance between the three aforementioned objectives, which are quantified by the already trained blocks of the GAN architecture}.}
	\label{fig:Architectures}
	\vspace{-1mm}
\end{figure}

Following the diagram shown in Figure \ref{fig:Architectures}, the design of the proposed framework can be split in {\color{black}four main blocks: 
\begin{itemize}[leftmargin=*]
	\item Target model $T(\cdot)$ to be audited, i.e., the classification model for which the counterfactual study is performed.
	\item A GAN architecture whose generator module allows inducing conditional perturbations on an input data instance $\mathbf{x}^{\mathbf{a},\oplus}$ (anchor) based on an attribute vector $\mathbf{b}$. Its discriminator module $D(\cdot)$ permits to evaluate the plausibility of a synthetically generated instance.
	\item An attribute classifier $C(\cdot)$ that predicts the present attributes in any image fed at its input. The attribute classifier is needed for training the conditional generator module of the GAN, so that the generation of ne instances allows inserting attribute-based modifications into an original instance without compromising the plausibility of the produced counterfactual.
	\item A multi-objective optimization algorithm that evolves the perturbation vector ${\bm \delta}=\mathbf{b}-\mathbf{a}$ to be imprinted on the anchor image $\mathbf{x}^{\mathbf{a},\oplus}$ to best balance between plausibility, adversarial power and change intensity. 
\end{itemize}}

The audited model is fed with the counterfactual example produced by the generator model of the GAN architecture, hence its only prerequisite is that the input of the audited model and the output of the generator are of the same dimensions. In what follows we will assume that the target model to be audited is a CNN used for image classification. Nevertheless, the framework can be adapted to audit other models and tasks whenever the output of the GAN discriminator and the input of the audited model are equally sized, and the measure of adversarial power accounts for the change induced by the counterfactual in the prediction of the model. 

The GAN is the part of the framework in charge of generating the counterfactuals fed to the audited model. Therefore, two requirements are set in this module: 1) the discriminator must be trained for a similar data distribution to that of the audited model; and 2) the generator model must be able to generate samples of such a distribution as per an \emph{attribute vector} $\mathbf{b}$ that controls specific features of the generated instance (image). This attribute vector is tuned by the multi-objective optimization algorithm seeking to maintain plausibility as per the discriminator, changing the output of the audited model and minimizing the intensity of the changes inserted in the original input image. 

At this point it is important to emphasize that the audited model is left aside the overall training process of the GAN for several reasons. To begin with, for practicality we assume minimum access to the audited model (black-box analysis). Therefore, the logits of the audited model are exploited with no further information on its inner structure. Furthermore, the goal of the discriminator is to decide whether the generated image follows the distribution of the training set, which must be regarded as a plausibility check. The task undertaken by the audited model can be of different types, for instance, to discriminate among \texttt{male} and \texttt{female}, \texttt{old} and \texttt{young} or any other task. 

The above three-objective optimization problem can be formulated mathematically as follows: let us denote an image on which the counterfactual analysis is to be made as $\mathbf{x}^\mathbf{a}\sim P_{\mathbf{X}}(\mathbf{x})$, which follows a distribution $P_{\mathbf{X}}(\mathbf{x})$ and has an attribute vector $\mathbf{a}\in\mathbb{R}^N$. The generator of the GAN model is denoted as $G(\mathbf{x}^\mathbf{a},\mathbf{b})$, whose inputs are the actual image $\mathbf{x}^\mathbf{a}$ and a desired attribute vector $\mathbf{b}$. In conditional generative models the generator is generally composed of an encoder $G_{enc}$ and a decoder $G_{dec}$. However, for some architectures, the model directly departs from a decoder, given the asumption that the latent code is sampled from a known distribution. 

Leaving the special cases aside for the sake of a clearer explanation, the image conditionally output by the generator is given by $\mathbf{x}^{\mathbf{b}\prime}= G_{dec}(G_{enc}(\mathbf{x}^\mathbf{a}),\mathbf{b})$. Ideally, when fed with the original attribute vector as the target, $\mathbf{x}^{\mathbf{a}\prime} \approx \mathbf{x}^\mathbf{a}$, i.e. the reconstructed image $\mathbf{x}^{\mathbf{a}\prime}=G_{dec}(G_{enc}(\mathbf{x}^\mathbf{a}),\mathbf{a})$ should resemble $\mathbf{x}^\mathbf{a}$ itself. For non-conditional generative architectures, the generated image is given by $\mathbf{x}^{\prime}=G_{dec}(G_{enc}(\mathbf{x}))$, where the objective is to have $\mathbf{x^\prime} \approx \mathbf{x}$. A discriminator $D(\mathbf{x}^{\mathbf{b},\prime})$ along with a classifier $C(\mathbf{x}^{\mathbf{b},\prime})$ is placed next along the pipeline to determine 1) whether the synthesized image $\mathbf{x}^{\mathbf{b},\prime}$ is visually realistic; and 2) whether the predicted attributes match the input ones. Again, for non-conditional GAN architectures, only the discriminator $D(\mathbf{x}^\prime)$ is necessary.

The overall loss function that drives the learning algorithm of the generator and discriminator is defined as a linear combination of the reconstruction and Wasserstein GAN losses. Assuming an encoder-decoder based generator architecture with latent vector $z$, the training loss for encoder $G_{enc}(\mathbf{x}^\mathbf{a})$ and decoder $G_{dec}(\mathbf{z},\mathbf{b})$ are given by:
\begin{equation}\label{eq:minGG}
\min_{G_{enc},G_{dec}} \lambda_1 \mathcal{L}_{rec}(\mathbf{x}^{\mathbf{a}},\mathbf{x}^{\mathbf{a}\prime}) + \lambda_2 \mathcal{L}_{att}^G(\mathbf{b},\smash{\widehat{\mathbf{b}}^\prime}) + \mathcal{L}_{adv}^G(\mathbf{x}^{\mathbf{b}\prime}),
\end{equation}
where:
\begin{align}
&\mathcal{L}_{rec}(\mathbf{x}^{\mathbf{a}},\mathbf{x}^{\mathbf{a}\prime}) = \mathbb{E}_{\mathbf{x}^\mathbf{a}\sim P_{\mathbf{X}}(\mathbf{x})} \left[\vert\vert\mathbf{x}^\mathbf{a}-\mathbf{x}^{\mathbf{b}\prime}\vert\vert_1\right]\;\text{(reconstruction loss),}\\
&\mathcal{L}_{att}^G(\mathbf{b},\smash{\widehat{\mathbf{b}}^\prime}) = \mathbb{E}_{\mathbf{x}^\mathbf{a}\sim P_{\mathbf{X}}(\mathbf{x}),\mathbf{b}\sim P_{\mathbf{B}}(\mathbf{b})} \left[\sum_{n=1}^{N=|\mathbf{b}|} H(b_n,\widehat{b_n}^\prime)\right]\;\text{(attribute loss)},\\
&\mathcal{L}_{adv}^G(\mathbf{x}^{\mathbf{b}\prime})= -\mathbb{E}_{\mathbf{x}^\mathbf{a}\sim P_{\mathbf{X}}(\mathbf{x}),P_{\mathbf{B}}(\mathbf{b})} \left[D(\mathbf{x}^{\mathbf{b}\prime})\right]\;\text{(adversarial loss).}
\end{align}

In the above expressions, $\mathbb{E}[\cdot]$ denotes expectation; $P_{\mathbf{B}}(\mathbf{b})$ indicates the distribution of possible attribute vectors $\mathbf{b}=\{b_n\}_{n=1}^N\in\mathbb{R}^N[0,1]$; $H(b_n,\smash{\widehat{b_n}^\prime})$ is the cross-entropy of binary distributions given by $b_n$ and $\smash{\widehat{b_n^\prime}} \in \smash{\widehat{\mathbf{b}}^\prime} = C(\mathbf{x}^{\mathbf{b}^\prime})$; and $D(\mathbf{x}^{\mathbf{b}\prime})=0$ if $\mathbf{x}^{\mathbf{b}\prime}$ is predicted to be fake. 

When it comes to the discriminator $D(\cdot)$ and the classifier $C(\cdot)$, their training loss is given by:
\begin{equation} \label{eq:minDC}
\min_{D, C} \lambda_3 \mathcal{L}_{att}^C(\mathbf{x}^\mathbf{a},\mathbf{a})+\mathcal{L}_{adv}^D(\mathbf{x}^\mathbf{a},\mathbf{b}),
\end{equation}
with:
\begin{align}
&\mathcal{L}_{att}^C(\mathbf{x}^\mathbf{a},\mathbf{a}) = \mathbb{E}_{\mathbf{x}^\mathbf{a}\sim P_{\mathbf{X}}(\mathbf{x})} \left[\sum_{n=1}^{|\mathbf{a}|} H(a_n,\smash{\widehat{a_n}}^\prime)\right],\\
& \mathcal{L}_{adv}^D(\mathbf{x}^\mathbf{a},\mathbf{b})=-\mathbb{E}_{\mathbf{x}^\mathbf{a}\sim P_{\mathbf{X}}(\mathbf{x})} \left[D(\mathbf{x}^{\mathbf{a}})\right]+\mathbb{E}_{\mathbf{x}^\mathbf{a}\sim P_{\mathbf{X}}(\mathbf{x}),P_{\mathbf{B}}(\mathbf{b})} \left[D(\mathbf{x}^{\mathbf{b}\prime})\right],
\end{align}
where $\smash{\widehat{a_n}}^\prime \in C(\mathbf{x}^\mathbf{a})$, and coefficients $\{\lambda_i\}_{i=1}^3$ permit to balance the importance of the above terms during the training process of the GAN architecture. For more general approaches, such as non-conditional GANs, the training loss is given by:
\begin{equation}
\min_{G_{enc},G_{dec}} \lambda_1 \mathcal{L}_{rec}(\mathbf{x},\mathbf{x}^\prime) + \mathcal{L}_{adv}^G(\mathbf{x}^\prime),
\end{equation}
where:
\begin{align}
&\mathcal{L}_{rec}(\mathbf{x},\mathbf{x}^\prime) = \mathbb{E}_{\mathbf{x}\sim P_{\mathbf{X}}(\mathbf{x})} \left[\vert\vert\mathbf{x}-\mathbf{x}^\prime\vert\vert_1\right],\\
&\mathcal{L}_{adv}^G(\mathbf{x}^\prime)= -\mathbb{E}_{\mathbf{x}\sim P_{\mathbf{X}}(\mathbf{x})} \left[D(\mathbf{x}^\prime)\right],
\end{align}
and again, coefficient $\lambda_1\in\mathbb{R}[0,1]$ allows tuning the relative importance of the reconstruction loss when compared to the adversarial loss. Once these losses have been defined, the GAN is trained via back-propagation to minimize the losses in Expressions \eqref{eq:minGG} and \eqref{eq:minDC} when measured over a training dataset.

Once trained, we exploit the GAN architecture to find counterfactual examples for a given test sample $\mathbf{x}^{\mathbf{a},\oplus}$ and an audited model $T(\mathbf{x})$, with classes $\{\texttt{label}_1,\ldots,\texttt{label}_L\}$. Specifically, we model the counterfactual generation process as a perturbation inserted into the attribute vector $\mathbf{a}$ of the test sample, i.e. $\mathbf{b}=\mathbf{a}+\bm{\delta}$, with $\bm{\delta}\in\mathbb{R}^N$. This perturbed attribute vector, through $G_{enc}$ and $G_{dec}$, yields a plausible image $\mathbf{x}^{\mathbf{b},\prime}$ that, when fed to the target model $T(\cdot)$, changes its predicted output. The conflict between adversarial power, plausibility and intensity of the perturbation from which the counterfactual example is produced gives rise to the multi-objective problem formulated as:
\begin{equation}\label{Cost} 
\min_{\bm{\delta}\in\mathbb{R}^N}  f_{gan}(\mathbf{x}^{\mathbf{a},\oplus},\bm{\delta};G,D),f_{adv}(\mathbf{x}^{\mathbf{a},\oplus},\bm{\delta};G,T),f_{att}(\bm{\delta}), 
\end{equation}
where:
\begin{itemize}[leftmargin=*]
\item $f_{gan}(\mathbf{x}^{\mathbf{a},\oplus},\bm{\delta};G,D)$ quantifies the \emph{unlikeliness} (no plausibility) of the generated counterfactual instance through $G(\cdot)$, which is given by the difference between the output of the discriminator $D(\cdot)$ corresponding to $\mathbf{x}^{\mathbf{a},\oplus}$ and $\mathbf{x}^{\mathbf{b},\prime}$ (Wasserstein distance). The more negative this difference is, the more confident the discriminator is about the plausibility of the generated counterfactual $\mathbf{x}^{\mathbf{b},\prime}$; 

\item $f_{adv}(\mathbf{x}^{\mathbf{a},\oplus},\bm{\delta};G,T)$ informs about the probability that the generated counterfactual does not confuse the target model $T(\cdot)$, which is given by the negative value of the cross-entropy of the soft-max output of the target model when queried with counterfactual $\mathbf{x}^{\mathbf{b},\prime}$; and

\item $f_{att}(\bm{\delta})$ measures the intensity of adversarial changes made to the input image $\mathbf{x}^{\mathbf{a},\oplus}$, which is given by $\vert\vert \bm{\delta} \vert\vert_2$. As we will later discuss, this measure can be replaced by other measures of similarity that do not operate over the perturbed attribute vector, but rather over the produced counterfactual image (for instance, structural similarity index measure SSIM between $\mathbf{x}^{\mathbf{a},\oplus}$ and $\mathbf{x}^{\mathbf{b},\prime}$).
\end{itemize}

To efficiently find a set of input parameter perturbations $\{\bm{\delta}\}$ balancing among the above three objectives in a Pareto-optimal fashion, we resort to multi-objective optimization algorithms. Specifically, we opt for derivative-free meta-heuristic solvers, which allow efficiently traversing the search space $\mathbb{R}^N$ of decision variables $\bm{\delta}$ and retaining progressively better non-dominated counterfactual instances without requiring any information of the derivatives of the objectives under consideration. {\color{black} The multi-objective solver makes use of the already trained GAN using the weighted loss functions defined in Expressions \eqref{eq:minGG} and \eqref{eq:minDC} to evaluate $f_{gan}(\cdot)$, $f_{adv}(\cdot)$ and $f_{att}(\cdot)$ for a given anchor instance $\mathbf{x}^{\mathbf{a},\oplus}$ and candidate perturbation $\bm{\delta}$, so that such objective values are used to select, evolve and filter out perturbations that yield counterfactuals dominated in the Pareto space spanned by these three objectives. In other words, the multi-objective solver allows efficiently traversing the space of possible perturbations in search for counterfactuals that dominate the interplay between the aforementioned objectives. Such objectives can be evaluated for every given perturbation $\bm{\delta}$ based on the outputs of the target model $T(\cdot)$ and the $G_{enc}(\cdot)$, $G_{dec}(\cdot)$ and $D(\cdot)$ modules of the already trained GAN architecture.}
\begin{algorithm}[h!]
	\SetAlgoLined
	\DontPrintSemicolon
	\KwIn{Target model to be audited $T(\mathbf{x})$; GAN architecture ($G,D$); attribute classifier $C(\mathbf{x})$; annotated training set $\mathcal{D}_{train}$; test image $\mathbf{x}^{\mathbf{a},\oplus}$ for counterfactual study; weights $\{\lambda_i\}_{i=1}^3$}
	\KwOut{Multi-criteria counterfactuals balancing between $f_{gan}(\cdot)$, $f_{adv}(\cdot)$ and $f_{att}(\cdot)$}
	Train GAN architecture via back-propagation over training dataset and loss functions in Expressions \eqref{eq:minGG} to \eqref{eq:minDC}\;
	Initialize a population of perturbation vectors $\bm{\delta}\in\mathbb{R}^N$\;
	\While{stopping criterion not met}{
		Apply search operators to yield offspring perturbation vectors\;
		Evaluate $f_{gan}(\cdot)$ (\emph{plausibility}), $f_{adv}(\cdot)$ (\emph{adversarial success}) and $f_{att}(\cdot)$ (\emph{change intensity}) of offspring perturbations\;
		Rank perturbations in terms of Pareto optimality\;
		Retain the Pareto-best perturbations in the population\;
	}
	Select non-dominated perturbations from population\;
	Produce counterfactual images by querying the GAN with $\mathbf{x}^{\mathbf{a},\oplus}$ and each selected perturbation vector
	\caption{Generation of multi-criteria counterfactuals}
	\label{alg:counter}
\end{algorithm}

Algorithm \ref{alg:counter} summarizes the process of generating counterfactuals for target model $T(\cdot)$, comprising both the training phase of the GAN architecture and the meta-heuristic search for counterfactuals subject to the three conflicting objectives. The overall framework departs from the training process of a GAN architecture (line \textbf{1}) over a training dataset $\mathcal{D}_{train}$ that collects samples (images) annotated with their attribute vectors $\mathbf{a}$ (only for conditional GANs). Once trained, and similarly to the usual workflow of population-based heuristic solvers, the algorithm initializes uniformly at random a population of perturbation vectors (line \textbf{2}), which are iteratively evolved and refined (lines \textbf{3} to \textbf{8}) as per the Pareto optimality of the counterfactual images each of them produces. To this end, evolutionary search operators (crossover and mutation) are applied over the population (line \textbf{4}) to produce offspring perturbation vectors, which are then evaluated (line \textbf{5}) and ranked depending on their Pareto dominance (line \textbf{6}). By keeping in the population those perturbation vectors that score best in terms of Pareto optimality (line \textbf{7}) and iterating until a stopping criterion is met, the framework ends up with a population of Pareto-superior perturbation vectors (line \textbf{9}){\color{black}, that can be inspected visually to understand which image components affect the most along the direction of each objective (line \textbf{10}). Indeed, since decision variables evolved by the multi-objective solvers can be directly linked to attributes imprinted to the anchor image, the amplitude of any given component of an evolved vector can be interpreted as the intensity by which the corresponding attribute is modified in the generated counterfactual. Consequently, it is possible to determine which attributes are more relevant depending on the region of the objective space under focus.}

\subsection{Target Audiences and Examples of Use Cases} \label{ssec:audiences}

To round up the presentation of the proposed framework, we pause briefly at the target audiences envisioned for its use, and scenarios that could illustrate its use in practical settings. Many examples could be used to exemplify these scopes, among which three specific areas currently under active investigation are chosen: bio-metric authentication, the discovery of new materials and creative industrial applications. These three use cases target two different audiences: developers and final users. 

The use of bio-metric authentication is extensive nowadays in a manifold of sectors managing critical assets. However, auditing machine learning models used for bio-metric authentication is not straightforward. They can be audited by adversarial attack testing, but this analysis focuses on subtle (namely, not noticeable) adversarial perturbations made to an input to the audited model. Therefore, they aim at analyzing the robustness of the model against malicious attacks designed not to be easily detectable (e.g. one-pixel attacks), rather than at discerning which plausible inputs can lead to a failure of the authentication system even if not deliberately designed for this purpose. The framework presented in this work can be of great value for developers to explore the reality-bound limitations of their methods. {\color{black}This can help them determine complementary information requested during the process to increase the robustness of the model against plausible authentication breakpoints, which can be uncovered by examining biometric features that are specially sensitive for the adversarial power objective.} 

New material discovery is also a field in which high-dimensional datasets are utilized. The addition of our proposed framework might help experts reduce the amount of non-plausible composites to be synthesized, or to discover diverse alternative materials with differing properties in terms of elasticity, conductivity or thermal expansion, to cite a few. This would in turn ease the practice of material experts by considerably reducing the space of possible materials to be explored, and opening new possibilities in their laboratory processes without requiring any technical knowledge in Artificial Intelligence.

Finally, we highlight the possibilities brought by the proposed framework for the creative industry. Such a framework could be coupled with a design software so that it would help in the generation of creative content by proposing new alternatives of already produced products (e.g. new designs of mechanical components, new audiovisual pieces, novel architectural proposals) with varying levels of compliance with respect to plausibility, amount of the change and properties that are specific to the use case at hand. In essence, the framework could be of great use for aiding the creative process hand in hand with the expert. 

\section{Experimental Setup}\label{sec:experiments}

This section introduces the actual architectures and  models that were used to prove the framework. Six GAN architectures are presented, followed by six third-party classifiers that were audited in the experiments. All GAN architectures are extracted from the literature as pre-trained cases from the original authors themselves. The classifiers are trained with the test sets of each of the GAN dataset to assure the same data domain is maintained and no knowledge leakage is produced. All the source code for reproducing these experiments will be released at \url{https://github.com/alejandrobarredo/COUNTGEN-Framework}, together with a Python library that can be used for applying this framework over custom datasets.

\subsection{Considered GAN Architectures}

The architectures utilized fall under three main GAN categories. Although each of them consists of a particular implementation containing its particular caveats. The different GAN approaches are: Conditional GAN, Unconditional GAN and a combination of both. 

\begin{figure}[ht]
	\centering
	\subfigure[BicycleGAN]{%
		\includegraphics[width=0.47\columnwidth]{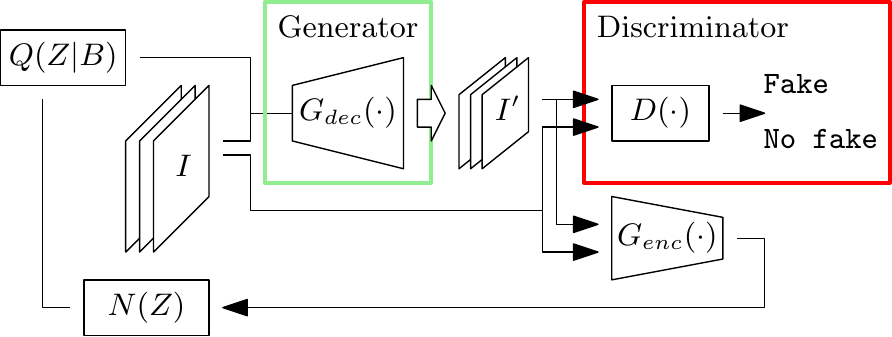}
		\label{fig:BicycleGAN_Architecture}}
	\quad
	\subfigure[AttGAN]{%
		\includegraphics[width=0.47\columnwidth]{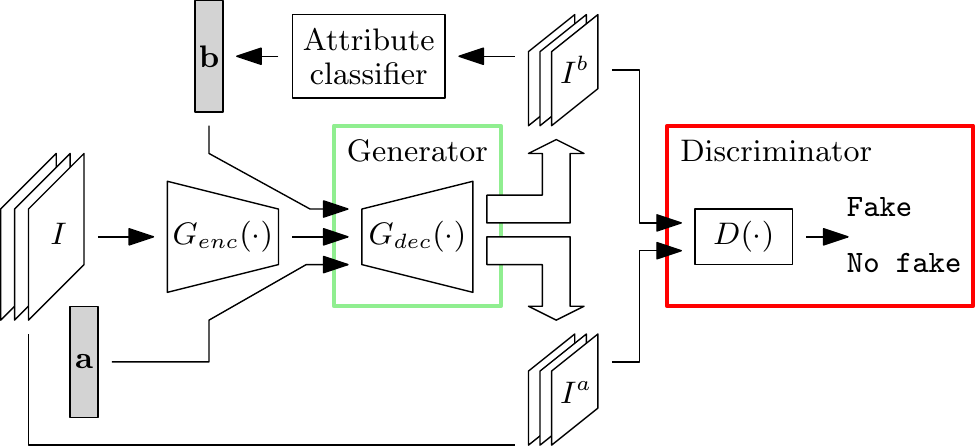}
		\label{fig:AttGAN_Architecture}}
	\quad
	\subfigure[BigGAN]{%
		\includegraphics[width=0.47\columnwidth]{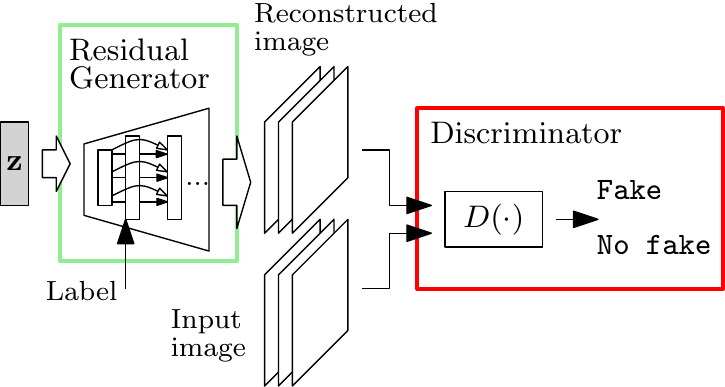}
		\label{fig:BigGAN_Architecture}}
	\quad
	\subfigure[ShapeHDGAN]{%
		\includegraphics[width=0.47\columnwidth]{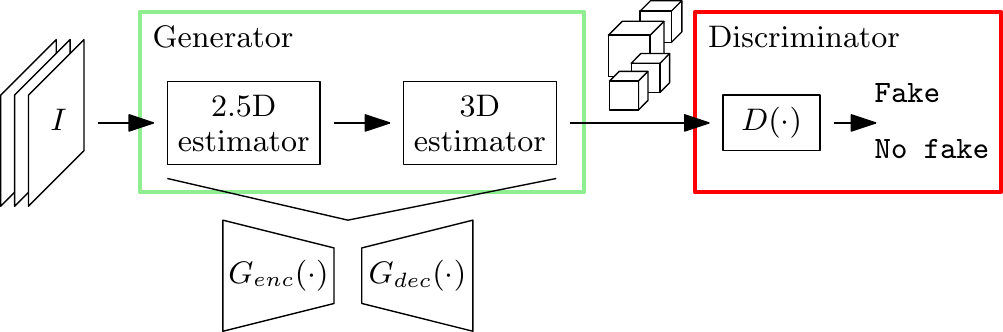}
		\label{fig:ShapeHD_Architecture}}
	\quad
	\subfigure[StyleGAN2]{%
		\includegraphics[width=0.47\columnwidth]{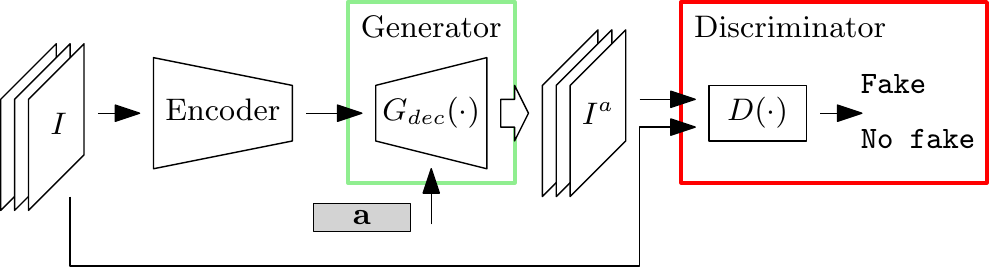}
		\label{fig:StyleGAN2_Architecture}}
	\quad
	\subfigure[Conditional GAN]{%
		\includegraphics[width=0.47\columnwidth]{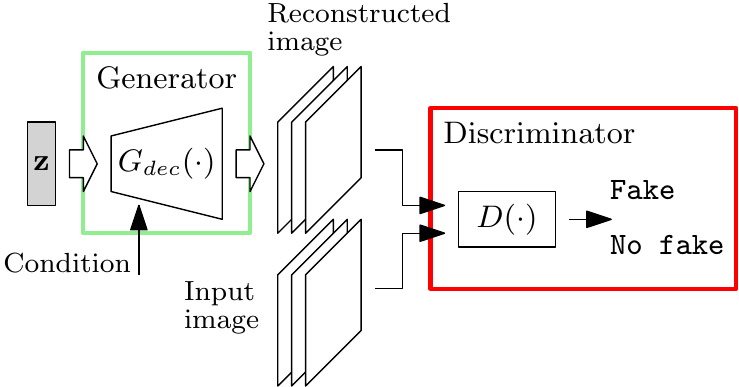}
		\label{fig:MnistGAN_Architecture_GAN}}
	\quad
	\caption{Block diagram of the proposed system comprising the (a) BicycleGAN; (b) AttGAN; (c) BigGAN; (d) ShapeHDGAN; (e) StyleGAN2; and (f) Conditional GAN (CGAN).}
	\label{fig:figure}
\end{figure}

\subsubsection{BicycleGAN}

This first BicycleGAN architecture combines conditional and unconditional GAN architectures for the task of image-to-image translation \cite{zhu2017toward}. To this end, BicycleGAN generates the output as a distribution of solutions in a conditional generative setting. The mapping is disambiguated through a latent vector which can be sampled at test time. The authors present their solution as an improSvement for the known \emph{mode collapse} problem, since it reduces the pitfall of having one-to-many solutions as a result of utilizing a low-dimensional latent vector. 

As shown in the diagram of Figure \ref{fig:BicycleGAN_Architecture}, BicycleGAN combines conditional and unconditional GAN architectures to generate their own. The first part, highlighted in green, is that of a cVAE-GAN \cite{kingma2013auto,larsen2016autoencoding}: the model first encodes the ground truth into a latent space, and then it is reconstructed by means of a generator trained with a Kullback–Leibler divergence loss. The second combined model is a cLR-GAN \cite{donahue2016adversarial,dumoulin2016adversarially,chen2016infogan}: contrarily to the first part, the cLR-GAN departs from a randomly generated latent vector, while the encoder is trained from recovering it from the output image created in the generator. Finally, the combination of these two different constraints form the BicycleGAN architecture, which enforces the connection between the output and latent code simultaneously for both directions. This resulting architecture is able to generate more diverse and appealing images for every image-to-image translation problem.

The implementation of the network was retrieved from \cite{BicycleGAN-github} with the pretrained models utilized for the experiments covered in the next section. The architecture consists of a U-Net \cite{ronneberger2015u} generator $G(\cdot)$, which in turn contains an encoder-decoder architecture with symmetric skip connections. The discriminator $D(\cdot)$ is composed as a combination of two PatchGAN \cite{isola2017image} of different scales which resolve the fake/real prediction for $70\times 70$ and $140\times 140$ image patches. Finally, for the standalone encoder $G_{enc}(\cdot)$, a ResNet \cite{he2016deep} is utilized. Further information about the structure and training process of the BicycleGAN architecture can be found in \cite{zhu2017toward}.

\subsubsection{AttGAN}

This second architecture presents a conditional GAN capable of editing facial attributes of human faces while preserving the overall detail of the image \cite{he2019attgan}. In the seminal work presenting this architecture, the training process is performed by conditioning the latent vector to match the vector representing the given facial attributes for the image at hand. The network is devised such that this vector is real-valued, which allows for the inference of facial attributes for a given intensity. During inference, attributes can be changed by modifying the values of the variables in the latent vector.

Figure \ref{fig:AttGAN_Architecture} depicts a diagram of the AttGAN model, which is trained by means of two constraining conditions. For one, the model attempts to match the input attributes with the predicted attributes at the end of the architecture. For the other, the model is constrained to match the generated image to that at its input. The latter is governed by a reconstruction loss. The former, forcing the latent vector to match the attributes of the images, is governed by a standard cross-entropy loss. The combination of these two constraints result in a model capable of generating faces with varying attributes and remarkable realism.

The implementation of the network is that available at \cite{AttGAN-github}. The discriminator $D(\cdot)$ is composed of a stack of convolutional layers followed by fully-connected layers. The classifier $C(\cdot)$ shares all the convolutional layers from $D(\cdot)$, and follows the same structure ended in fully-connected layers. The encoder $G_{enc}(\cdot)$ is composed of several convolutional layers, while the decoder $G_{dec}(\cdot)$ is composed of a stack of transposed convolutional layers. As in BicycleGAN, a symmetrical skip connection is set between the encoder and decoder. Further information about the architecture and training process can be accessed in \cite{zhu2017toward}.

\subsubsection{BigGAN}

Despite increasing achievements in GAN modeling, the scale and diversity of datasets such as \texttt{ImageNet} has remained a hard task over the years. In \cite{brock2018large}, the authors train a generative network at the largest scale yet possible. In such a research they study the instabilities specific to the scale. They discovered the so-called \emph{truncation trick} as a result of using orthogonal regularization to control the trade-off between the sample fidelity and variety by reducing the variance of the generators input. This improvement allowed for a new state of the art for class conditional image synthesis. When trained on ImageNet at a resolution of $128\times 128$ pixels, the newly proposed BigGAN architecture depicted in Figure \ref{fig:BigGAN_Architecture} achieves an inception score almost three times as large as that of previous image synthesis models. We use the pretrained BigGAN implementation\footnote{BigGAN Pytorch implementation, \url{https://github.com/ajbrock/BigGAN-PyTorch}, accessed on November 12th, 2021.} made available by the authors of \cite{brock2018large}.

\subsubsection{ShapeHDGAN}

This fourth ShapeHDGAN architecture is capable of rendering 3D meshes of objects from single 2D views. This particular task is of great complexity given that the solution landscape is composed of countless shapes that do not pertain to an object and renders them implausible. Most existing approaches fail at generating detailed objects. ShapeHDGAN gives a solution to this problem by virtue of an generative environment with adversarially learned shape priors that serve the purpose of penalizing if the model renders unrealistic meshes. 

As shown in Figure \ref{fig:ShapeHD_Architecture} the model consists of two main components. A 2.5D sketch estimator and a 3D shape estimator that predicts a 3D object from an image. It consists of three stages. In the first stage, the 2.5D estimator -- a encoder-decoder structure -- predicts the object depth, normals and silhouette from a RGB image. Then, the second stage generates a 3D shape from the previous 2.5D sketch. The last stage is composed by an adversarially trained CNN that tunes the generated shape into a real object.

The implementation was retrieved from \cite{ShapeHD-github}. The 2.5D sketch estimator is composed by a ResNet-18 encoder $G_{enc}(\cdot)$ mapping a $256\times 256$ image into $512$ feature maps of size $8\times 8$. The $G_{dec}(\cdot)$ model has four stacked transposed convolutional layers. The predicted silhouette permits to mask the depth and normal estimations to be then used as the input of the 3D generator. The 3D shape estimator is also composed of an encoder-decoder architecture. The encoder is an adapted from a ResNet-18 to handle 4 channels and encode them into a $200$-dimensional latent vector. The decoder comprises five stacked transposed convolutional layers, which generate a $128\times 128\times 128$ voxel at its output. Further details are available at \cite{shapehd}.

\subsubsection{StyleGAN2}

StyleGAN \cite{karras2019style} is a unconditional GAN architecture with one of the most realistic results for unconditional generative image modeling. For this study, we choose the StyleGAN2 implementation, which is a revised variant that improves upon the artifacts of the original StyleGAN model \cite{WhichFaceIsReal} by virtue of small albeit intelligently devised modifications to the generator model of the original StyleGAN model. The implementation was retrieved from \cite{StyleGAN2-github} with the pre-trained models for the experiments carried out in the following sections.

\subsubsection{Conditional GAN}

Finally we decided to add a last model that allows us to explore some variations within. This time, we selected a well-known conditional GAN architecture \cite{mirza2014conditional} trained over the MNIST image classification dataset. The conditional GAN departs from a random noise vector and a single variable that acts as a condition for the generation process. In this way, the generative network learns to switch between the learned distributions for each label by means of a input condition. This feature resembles to that of AttGAN, with the difference that in this one, the models does not start from an encoding.

Figure \ref{fig:MnistGAN_Architecture_GAN} shows the structure of the conditional GAN model. The implementation was retrieved from the public python library GANS2 \cite{MnistGAN-github} which includes a set of ready-to-build, plug-and-play GAN architectures. 

\subsection{Audited Classification Models}

After introducing the GAN models under consideration, we now introduce the models that will be audited by means of our GAN-based counterfactual generation framework. For the experiment utilizing BicycleGAN, a classifier is trained to predict the type of footwear corresponding to the image fed at its input (\texttt{Shoe} versus \texttt{NoShoe}). For the case considering AttGAN, the classifier to be audited predicts whether the human face input to the model corresponds to a \texttt{male} or to a \texttt{female}. The case using BigGAN considers a ResNet-50 classifier that discriminates among the 1000 classes represented in the \texttt{ImageNet} dataset. When the framework considers ShapeHDGAN, the classifier is trained to distinguish between a \texttt{chair} and a \texttt{Xbox}. For StyleGAN2, the classifier discriminates whether the input image is a \texttt{cathedral} or an \texttt{office}. Finally, the classifier audited by our framework configured with the cGAN aims to address a multi-class classification problem over the same MNIST dataset, yet ensuring that different data partitions are used for training the cGAN model and the classifier itself.

These third-party models consist of several convolutional and residual layers, ending in a series of fully-connected layers that connect the visual features extracted by the former with the categories defined in the dataset under consideration. Every classifier model was trained with the test data that was not used for training the corresponding GAN architecture, thereby ensuring no information leakage between the generators and the third-party models to be audited. Table \ref{tab:Classifiers} summarizes the topological configuration of the models for which counterfactuals are generated by our framework, as well as the training parameters set for every case.
\begin{table}[h!]
	\caption{Structure and training parameters of the models audited by the proposed framework.}
	\resizebox{\columnwidth}{!}{%
		\begin{tabular}{clc}
			\toprule 
			\multirow{2}{*}{GAN} & \multicolumn{2}{c}{Audited classifier $T(\mathbf{x})$} \\
			\cmidrule{2-3}
			& \multicolumn{1}{c}{Network architecture} & \multicolumn{1}{c}{Training parameters}\\
			\midrule
			BicycleGAN & \makecell[l]{Conv2d($64$, $3\times 3$, ReLu) + Conv2d(32, $3\times 3$, ReLu) \\ + Dense(1, Sigmoid)} & \makecell[l]{Adam, binary \\cross-entropy loss} \\			
			\midrule
			AttGAN & \makecell[l]{Conv2d($16$, $3\times 3$, ReLu) + Dropout(0.1) \\
				+ Conv2d(4, $3\times 3$, ReLu) + Dense(1,Sigmoid)} & \makecell[l]{Adam, binary \\cross-entropy loss} \\
			\midrule
			BigGAN & \makecell[l]{Conv2d($64$, $7\times 7$, ReLu) + MaxPooling($3\times 3$) \\
					+ Conv2d(64, $3\times 3$, ReLu) + Conv2d(128, $3\times 3$, ReLu)\\
					+ Conv2d(256, $3\times 3$, ReLu) + Conv2d(512, $3\times 3$, ReLu) \\
					+ AvgPooling($7\times 7$) + Dense(500,1000) + Dense(1000,Softmax)} & \makecell[l]{SGD(0.01, 0.9), categorical \\cross-entropy loss}\\
			\midrule
			ShapeHDGAN & \makecell[l]{Conv2d($32$, $3\times 3$, ReLu) + BatchNorm \\ + MaxPooling($2\times 2$) + Conv2d($8$, $3\times 3$, ReLu) + BatchNorm \\
			+ MaxPooling($2\times 2$) + Dense(100,ReLu) + Dense(1, Sigmoid)} & \makecell[l]{SGD(0.01, 0.9), binary \\cross-entropy loss} \\
			\midrule
			StyleGAN2 & \makecell[l]{Conv2d($16$, $3\times 3$, ReLu) + Dropout(0.1) \\
				+ Conv2d(4, $3\times 3$, ReLu) + Dense(1,Sigmoid)} & \makecell[l]{Adam, binary \\cross-entropy loss} \\
			\midrule
			cGAN & \makecell[l]{Conv2d($32$, $3\times 3$, ReLu) + BatchNorm \\ + MaxPooling($2\times 2$) + Dense(100,ReLu) \\ + Dense(10,SoftMax)} & \makecell[l]{SGD(0.01, 0.9), categorical \\cross-entropy loss} \\
			\bottomrule
			
			\multicolumn{3}{l}{{\small Conv2d(A,B,C): convolutional layer with A filters of size B and activation C.}}\\
			\multicolumn{3}{l}{{\small SGD($l$,$m$): Stochastic Gradient Descent with learning rate $l$ and momentum $m$.}} \\ 
			\multicolumn{3}{l}{{\small In all cases the batch size is set to 16 instances, and the number of epochs is 10.}} \\
			\multicolumn{3}{l}{{\small Flattening operations are not displayed for clarity.}}
		\end{tabular}%
	}
	\label{tab:Classifiers}
\end{table}

The accuracy achieved by the trained classifiers over a 20\% holdout of their dataset are reported in Table \ref{tab:classifier_scores}, together with the number of classes, total examples to train and validate the audited model, and the class balance ratio. As can be observed in this table, the audited models reach a very high accuracy (over 94\% in most cases, except for the \texttt{ImageNet} classifier due to the notably larger number of classes of the dataset), so that the adversarial success of the produced counterfactual examples can be rather attributed to the explanatory capabilities of the devised framework than to a bad performance of the audited classifier.{\color{black} Furthermore, we verified the capability of the trained GANs to conditionally generate new instances based on a vector of attributes, by 1) verifying the convergence of the discriminator and classifier loss function over the training epochs, and 2) by visually inspecting the quality of several test instances and arbitrary perturbations.}
\begin{table}[ht]
	\centering
	\caption{Dataset and accuracy of the different black-box classifiers under target}
	\resizebox{\columnwidth}{!}{%
		\begin{tabular}{ccccccc}
			\toprule 
			GAN & Dataset & \# Examples & Classes & Class Balance & Accuracy & Source \\
			\midrule
			BicycleGAN & \texttt{Edges2Shoes} & 300 & 2 & 45\%/55\% & 94\% & \cite{isola2017image}\\
			AttGAN & \texttt{CelebA} & 900 & 2 & 49\%/51\% & 98\% & \cite{liu2015faceattributes}\\
			BigGAN & \texttt{ImageNet} & 1,281,167 (training) & 1000 & Varying & 75\% & \cite{he2016deep} \\
			ShapeHDGAN & \texttt{ShapeNet} & 600 & 2 & 49\%/51\% & 96\% & \cite{DBLP:journals/corr/ChangFGHHLSSSSX15}\\
			StyleGAN2 & \texttt{Style} & 540 & 2 & 49\%/51\% & 98\% & \cite{yu15lsun}\\
			cGAN & \texttt{MNIST} & 9000 & 10 & 10\% each & 96\% & \cite{lecun-mnisthandwrittendigit-2010}\\
			\bottomrule
		\end{tabular}%
	}
	\label{tab:classifier_scores}
\end{table}

\subsection{Multi-objective Optimization Algorithm}

We recall that the optimizer is in charge for tuning the output of the GAN generator to 1) maximize the difference in the result of the audited classifier (adversarial power); 2) minimize the amount of changes induced in the produced counterfactual parameters (change intensity); and 3) maximize the Wasserstein distance between the real and fake examples (plausibility). 

The search for counterfactual instances optimally balancing among these objectives can be efficiently performed by using a multi-objective evolutionary algorithm. Among the multitude of approaches falling within this family of meta-heuristic solvers, we select NSGA2 \cite{deb2002fast} with a population size of 100 individuals, 100 offspring produced at every generation, polynomial mutation with probability $1/N$ (with $N$ denoting the number of decision variables, which vary depending on the experiment and GAN under consideration) and distribution index equal to 20, SBX crossover with probability $0.9$ and distribution index 20, and 50 generations (equivalent to 5000 evaluated individuals per run). The use of this optimizer allows for a genetic search guided by non-dominated sorting in the selection phase, yielding a Pareto-dominant set of counterfactual examples that constitute the output of the framework. For its implementation we rely on the jMetalPy library for multi-objective optimization \cite{benitez2019jmetalpy}.

\section{Results and Discussion}\label{sec:discussion}

We now discuss on the results obtained from the experiments described above, articulating the discussion around the provision of an informed response to three main research questions:
\begin{itemize}
    \item[Q1.] Is counterfactual generation an optimization problem driven by several objectives?
    \item[Q2.] Do the properties of the generated counterfactual examples conform to general logic for the tasks and datasets at hand?
    \item[Q3.] Do multi-criteria counterfactual explanations serve for broader purposes than model explainability?
\end{itemize}

Answers to each of these research questions will be summarized after an analysis and discussion held over the produced counterfactual examples for each of the audited models detailed in Table \ref{tab:Classifiers}. For every experiment, we draw at random one anchor image $\mathbf{x}^{\mathbf{a},\oplus}$ from the test partition of the audited model and inspect the produced set of counterfactual variants both visually and quantitatively as per the three objectives under consideration. This examination of the results will be arranged similarly across experiments, portraying the output of the framework in a three-dimensional plot comprising the Pareto front approximated by the multi-objective solver. Each of the axes of this plot is driven by one of such objectives: change intensity $f_{att}(\cdot)$, adversarial power $f_{adv}(\cdot)$ and plausibility $f_{gan}(\cdot)$, all defined in Subsection \ref{ssec:structure}. It is important to note that for easing the visualization of the fronts, plausibility and adversarial power are inverted by displaying $1-f_{gan}(\cdot)$ and $1-f_{adv}(\cdot)$, so that $1-f_{gan}(\cdot)\geq 0.5$ denotes the region over which the counterfactual can be considered to be plausible. Similarly, the higher $1-f_{adv}(\cdot)$ is, the larger the difference between the outputs of the audited model when fed with the anchor image $\mathbf{x}^{\mathbf{a},\oplus}$ and its counterfactual variant will be (larger \emph{adversarial power}).

In the depicted Pareto front approximations for every experiment (in the form of three-dimensional scatter plots, parallel lines visualizations and chord diagrams), several specific counterfactual examples scattered over the front are highlighted with colored markers. These markers refer to the images plotted on the last subplot of each figure, so that it is possible to assess the counterfactual image/voxel corresponding to each of such points. The first image shown in the top row of images shown on the right of the figure represents the reference (anchor) image $\mathbf{x}^{\mathbf{a},\oplus}$, which is the departing point for the counterfactual generation. The first image shown in the bottom row of images is always the image belonging to the opposite class (or a targeted class in the case of the MNIST dataset) whose soft-max output corresponding to its class is lowest (worst predicted example of the other class existing in the dataset). Below every image, a bar diagram can be observed representing the value of the objectives corresponding to the image at hand. 

We now proceed with a detailed discussion for every experiment.

\subsection{Experiment \#1: BicycleGAN-based counterfactual generation for auditing a \texttt{Shoe} versus \texttt{NoShoe} footwear classifier}\vspace{3mm} 

\begin{figure}[!h]
	\centering
	\subfigure[]{\includegraphics[width=0.34\columnwidth]{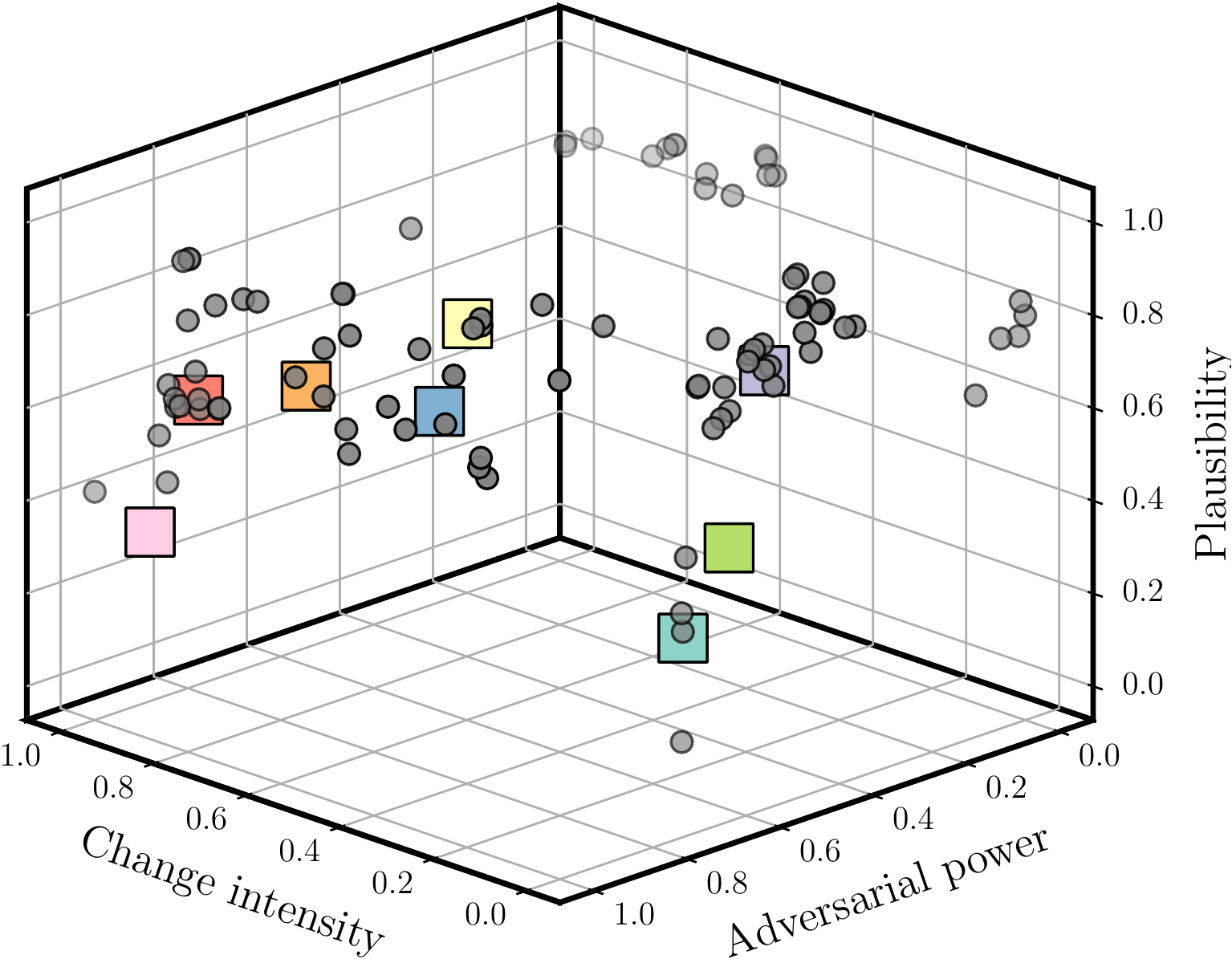}}
	\subfigure[]{\includegraphics[width=0.34\columnwidth]{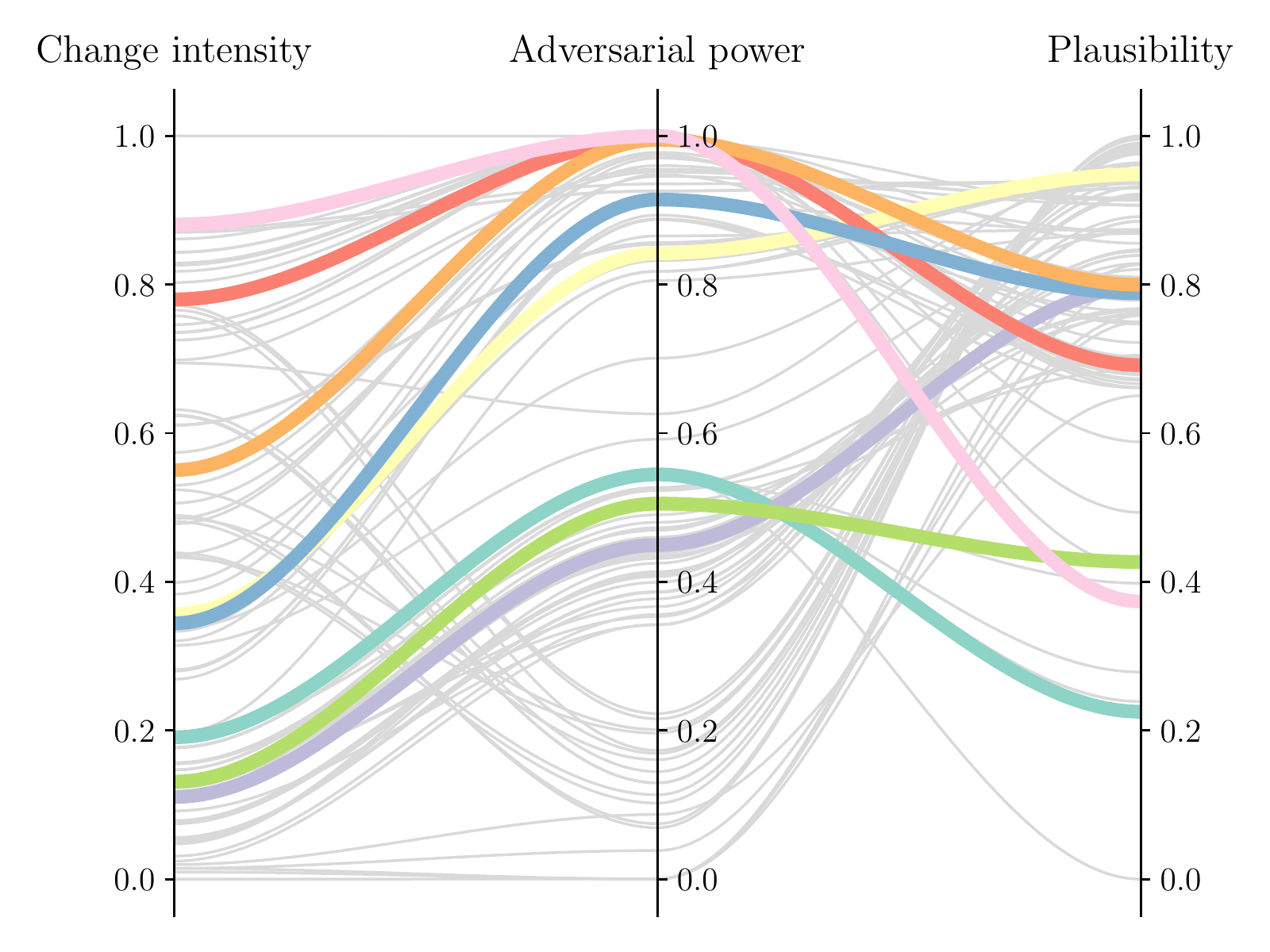}}
	\subfigure[]{\includegraphics[width=0.3\columnwidth]{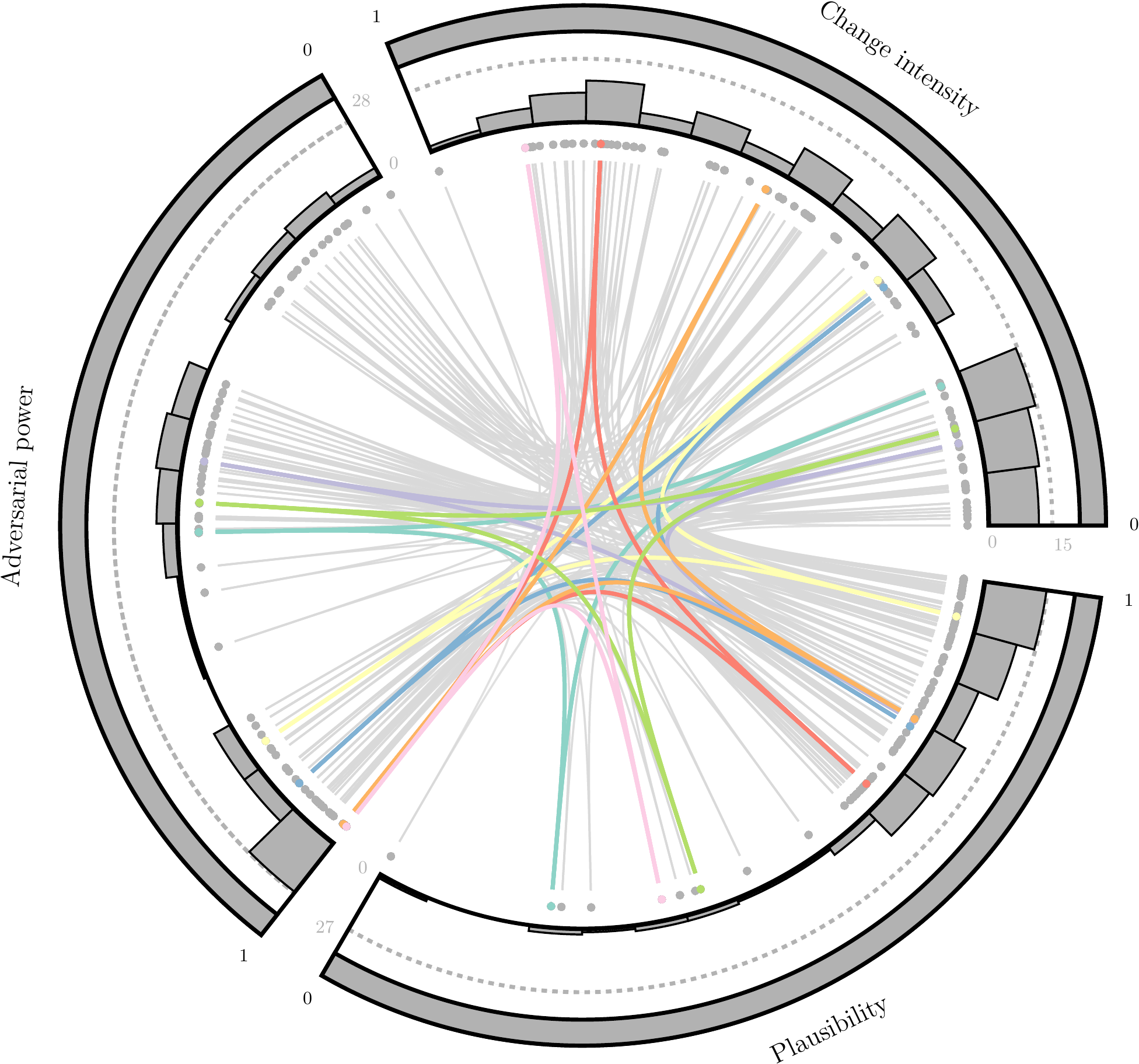}}
	\subfigure[]{\includegraphics[width=0.6\columnwidth]{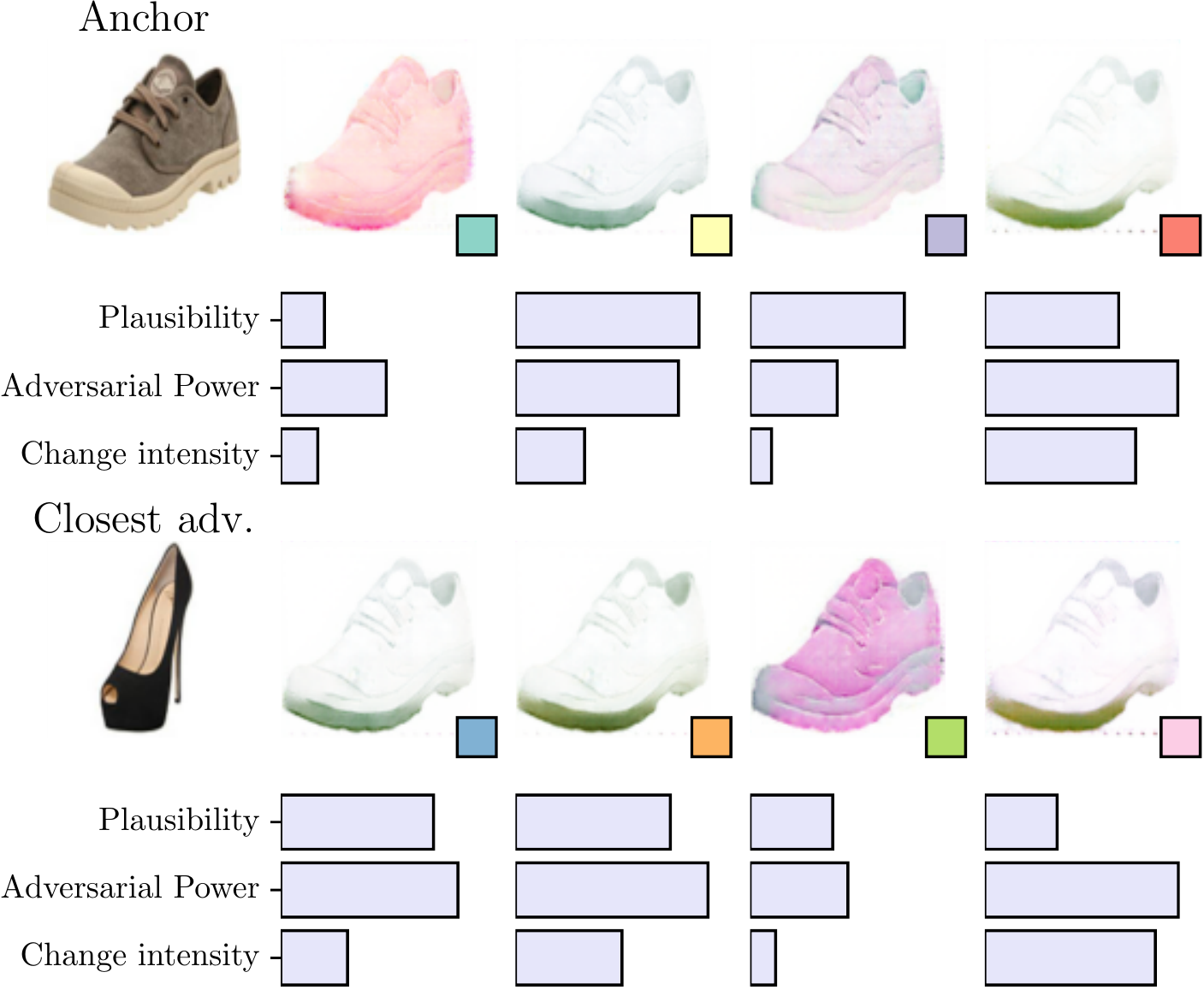}}
	\caption{(a) Pareto front of the counterfactual examples generated for a \texttt{Shoe} example by the proposed framework configured with a BicycleGAN model; (b) parallel lines visualization of the Pareto front; (c) chord plot of the Pareto front; (d) produced counterfactuals, together with the anchor image and the closest image of the opposite class for which the counterfactual is made. We note the color correspondence across all these subplots, by which the reader can follow the position of any of the highlighted counterfactual instances in the front, its objective values and image information.}
	\label{fig:bicyclegan_results}
\end{figure}

The outcomes of this first experiment are shown in Figure \ref{fig:bicyclegan_results}.a to Figure \ref{fig:bicyclegan_results}.d. The first one depicts a three-dimensional scatter plot of the Pareto front approximation generated by our framework for the man shoe selected for the anchor image $\mathbf{x}^{\mathbf{a},\oplus}$. Figures \ref{fig:bicyclegan_results}.b and \ref{fig:bicyclegan_results}.c correspondingly depict the parallel lines plot and chord plot of the counterfactual examples, which not only ease the visual inspection of the objective ranges spanned by the Pareto front, but also show the diversity sought for the counterfactual examples selected from the front. A color correspondence is fixed across subplots for the reader to track each counterexample through them. In Figure \ref{fig:bicyclegan_results}.d a coloring pattern is distinguishable over all the counterfactual examples highlighted in the approximated front. The image of the man shoe that serves as the anchor image $\mathbf{x}^{\mathbf{a},\oplus}$ appears to be complete. However, original colors are removed and uniformized all over the image. This fact informs about the influence of the color on the predicted label of the model. 

Returning to our intuition exposed in Section \ref{sec:framework}, two concerns must be kept in mind when analyzing these results. First, understanding the constraints of the dataset in use is of paramount importance. The data with which the classifier was trained is composed by different footwear instances. However, this dataset accounts just for a limited subset of the different possible footwear instances available in reality for both classes. This fact will make the predictions of the classifier change sharply between one class and the other when the instance for which it is asked does not conform to the class-dependent distribution of the training dataset. The second concern refers to the spread in the prediction scores. The solution front depicted in Figures \ref{fig:bicyclegan_results}.a to \ref{fig:bicyclegan_results}.c shows a nice spread in the prediction scores at first glance. However, this spread of solutions in the objective space does not entail that the corresponding counterfactual instances are visually diverse. We depict just 8 out of the 100 solutions in the approximated Pareto front, but they suffice to showcase that every generated counterfactual is very similar to each other with the exception of color. This suggests that the classifier is very susceptible to the color feature, and that the shape of the footwear is so relevant for the task that the counterfactual generation process needs to retain this feature to ensure plausibility. This bias is one of the insights provided by the proposed framework in this first experiment. 
\begin{figure}[!ht]
	\vspace{-1mm}
	\centering
	\includegraphics[width=0.8\columnwidth]{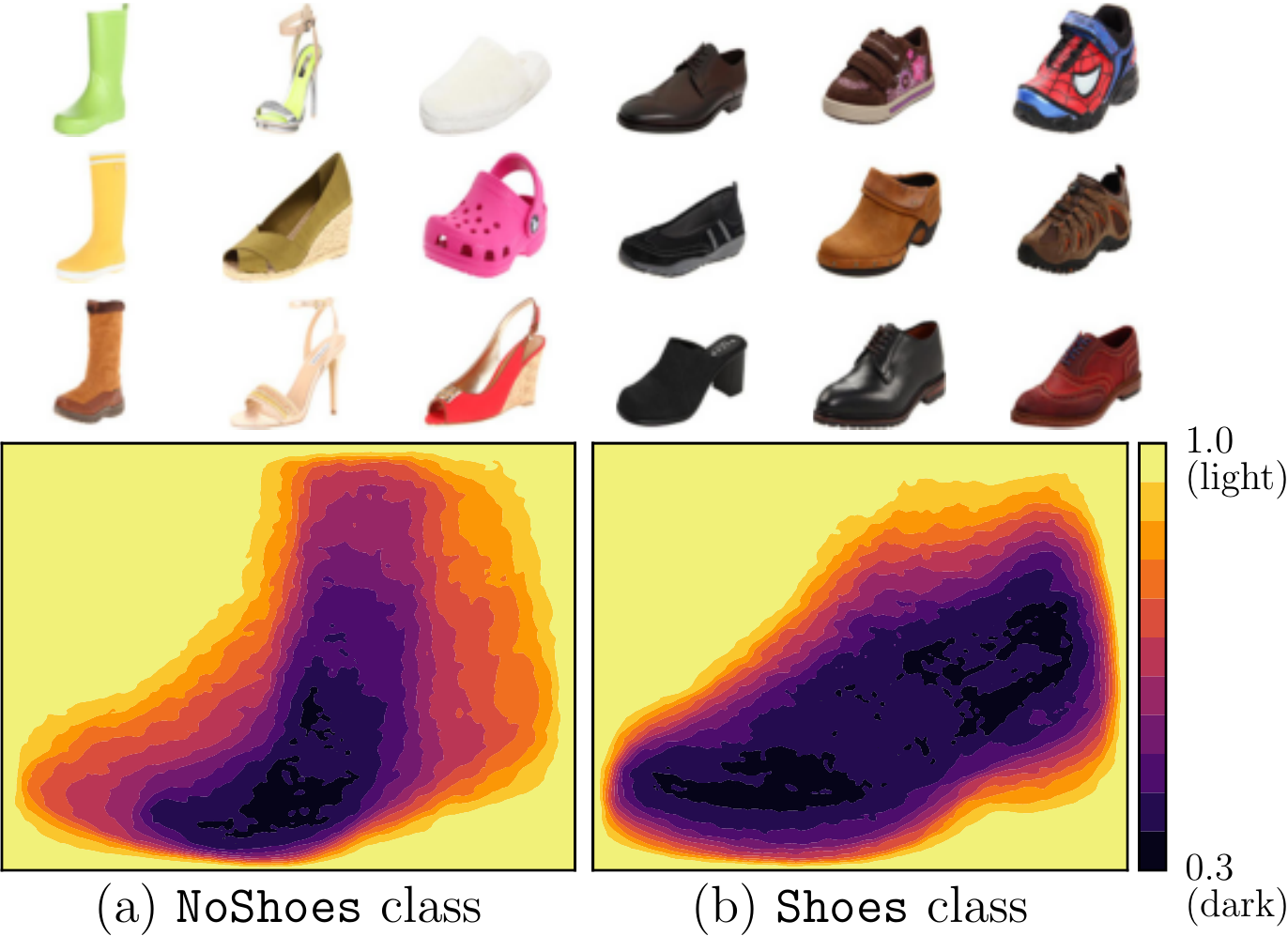}
	\caption{Analysis of the average RGB luminance $\ell(RGB)$ of the \texttt{Shoe} vs \texttt{NoShoe} dataset used to train the target classifier for the BicycleGAN experiment, together with some few examples of every class.}
	\label{fig:BicycleGAN_ana}
	\vspace{-1mm}
\end{figure}

The aforementioned statement is supported by Figure \ref{fig:BicycleGAN_ana}, which depicts the mean luminance of RGB pixels averaged over all the training examples of every class used for the audited model. Luminance has been computed as:
\begin{equation}
\ell(RGB) = \left(0.2126\cdot R + 0.7152\cdot G + 0.0722\cdot B\right)/255,
\end{equation}
where $\ell(RGB)\in\mathbb{R}[0,1]$ denotes a measure of luminance ($0$: dark, $1$: light) of a pixel with $R$ (red), $G$ (green) and $B$ (blue) channel values. As it can be observed in the bottom left plot of this figure, \texttt{shoe} instances have a clear bias in terms of footwear shape and image orientation, whereas the central part of the footwear for both classes is darker than the background. This is the reason why our proposed framework operates exclusively on the color feature and maintains the shape of the footwear when attempting at producing a counterfactual example for a \texttt{shoe}, yielding differently (brighter) colored yet identically shaped variants of the anchor.

\subsection{Experiment \#2: AttGAN-based counterfactual generation for auditing a \texttt{Man} versus \texttt{Woman} gender classifier}

\begin{figure}[!h]
	\centering
	\subfigure[]{\includegraphics[width=0.34\columnwidth]{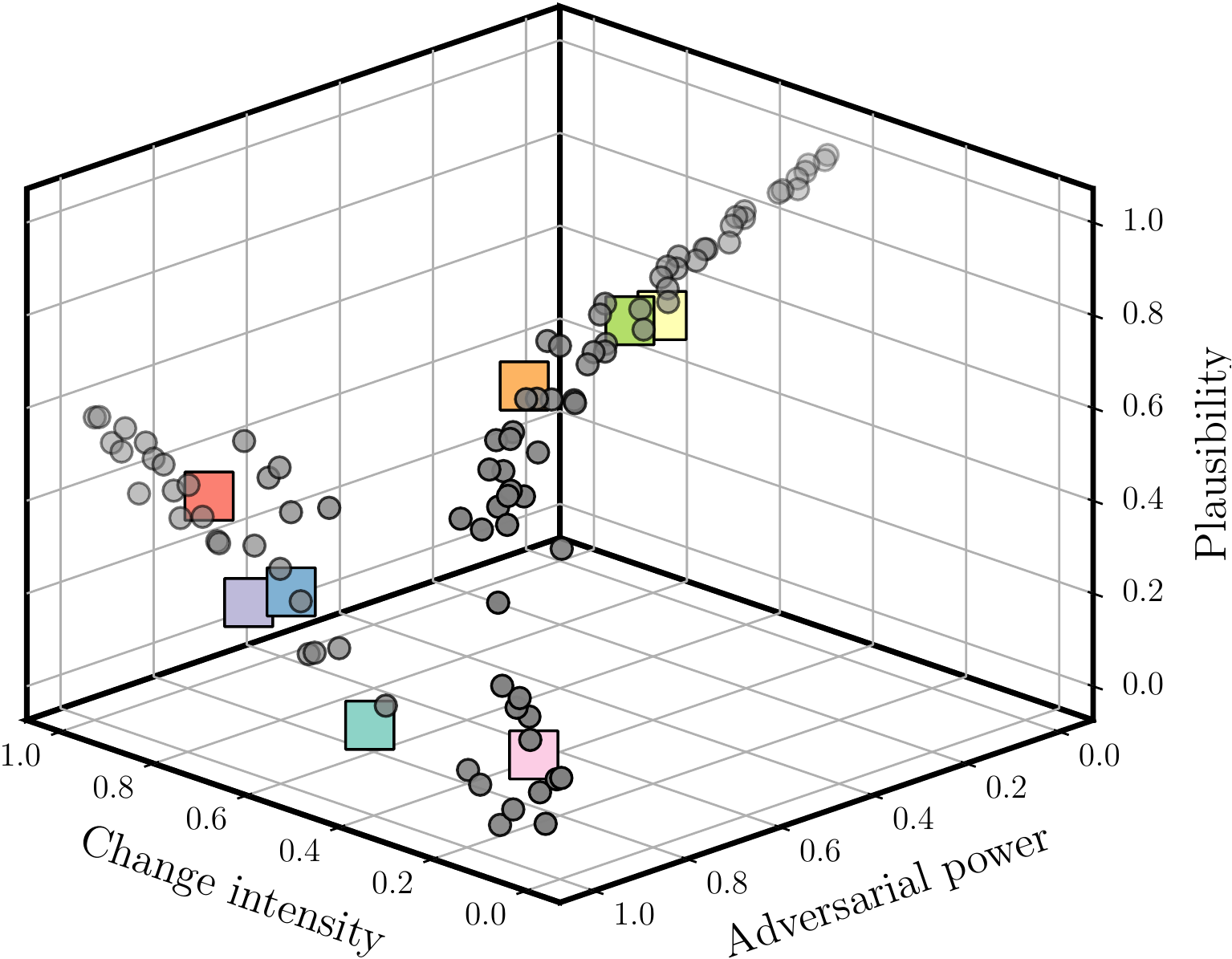}}
	\subfigure[]{\includegraphics[width=0.34\columnwidth]{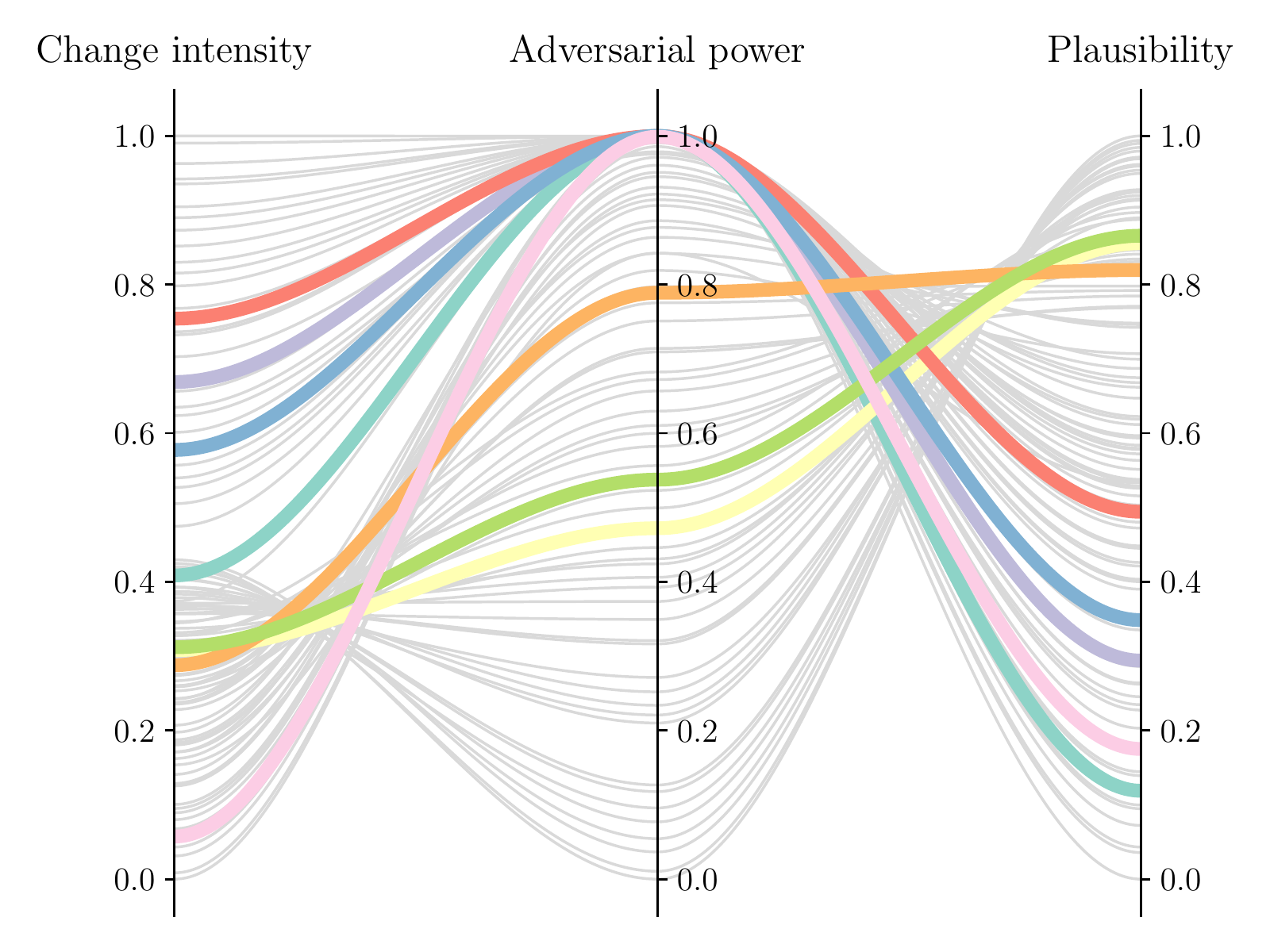}}
	\subfigure[]{\includegraphics[width=0.3\columnwidth]{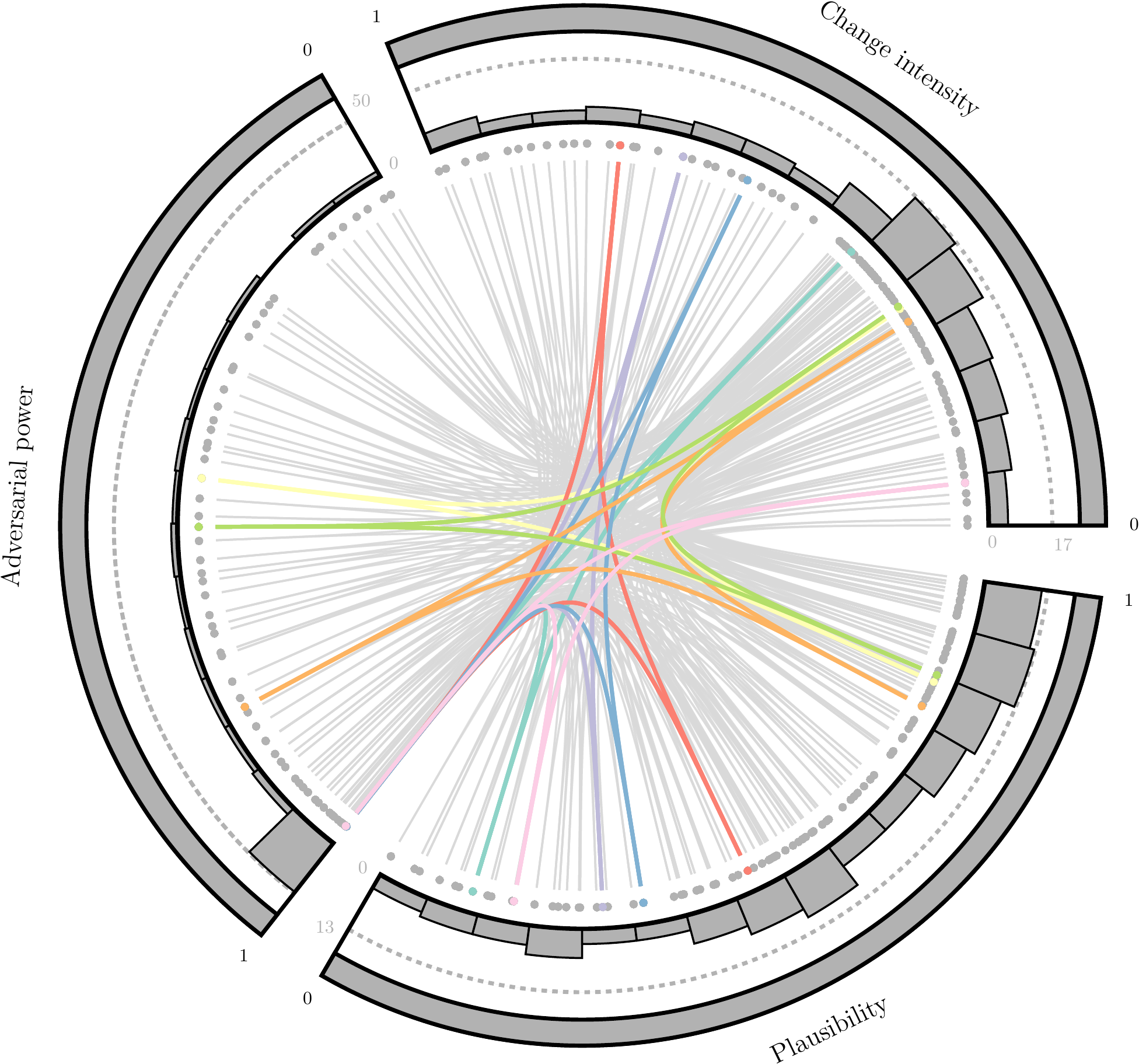}}
	\subfigure[]{\includegraphics[width=0.6\columnwidth]{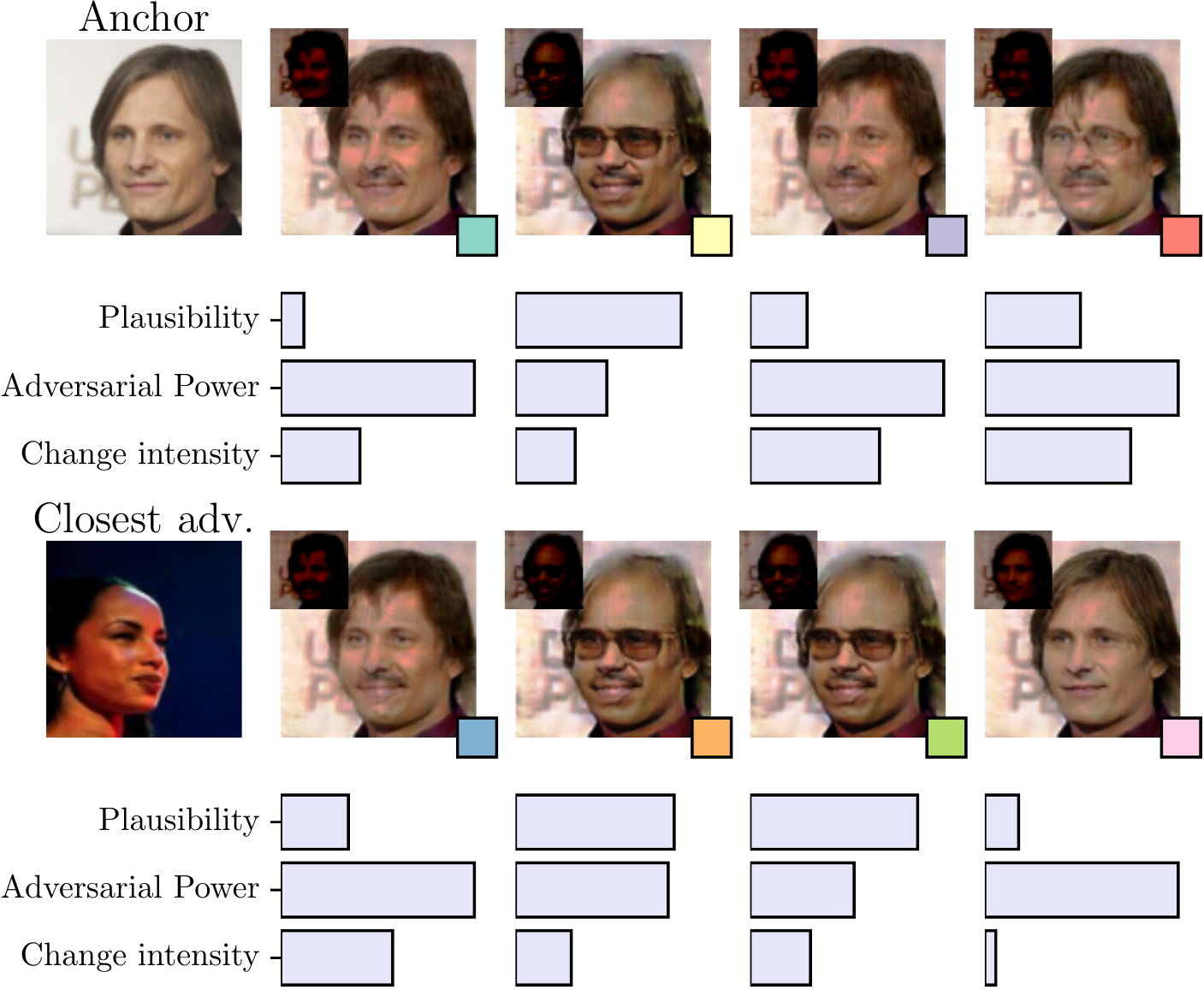}}
	\caption{(a) Pareto front; (b) parallel lines visualization; (c) chord plot; and (d) images of the counterfactual examples generated for a \texttt{male} example by the proposed framework configured with an AttGAN model. Every nested couple of images in subplot (d) represents the original produced counterfactual (small image located in the upper left corner of every plot) and its contrast adjusted version using histogram equalization \cite{gonzalez2002digital}.}
	\label{fig:attgan_results}
\end{figure}

The outcome of the devised framework corresponding to this second experiment is shown in Figures \ref{fig:attgan_results}.a to \ref{fig:attgan_results}.d. In this case, the reference image $\mathbf{x}^{\mathbf{a},\oplus}$ is an instance of the \texttt{Man} class from the \texttt{CelebA} dataset.

We begin by inspecting the shape of the produced Pareto front approximation in Figures \ref{fig:attgan_results}.a to \ref{fig:attgan_results}.c. It can be observed that diverse counterfactual explanations are found in the trade-off between plausibility and adversarial power, as well as between the intensity of the change and plausibility. Interestingly, adversarial power and change intensity seem to be less conflicting with each other, as no counterfactuals with high change intensity ($>0.5$) and low adversarial power were produced by the framework. The reason for the behavior of these two objectives may reside in the characteristics of the dataset and GAN in use: a high perturbation in the attribute vector imprints already enough changes in the generated counterfactual image to mislead the audit classifier, at the cost of a degradation of their plausibility. Similarly, counterfactuals with small change intensity can achieve a wide diversity of adversarial power values, which reveals that some perturbations are more effective than others when inducing a change in the output of the target model. However, it is clear (specially from the parallel lines visualization in Figure \ref{fig:attgan_results}.b) that those small changes that have high adversarial power, in general, result not to be plausible.

When qualitatively examining the generated counterfactuals, the plots nested in Figure \ref{fig:attgan_results}.c reveal that once again, the luminance is a deciding factor for adversarially modifying the anchor image. Leaving aside modifications over the color space, it is important to note that the plausibility of counterfactuals seems to be tightly linked to the insertion of glasses or a smiling pose. On the contrary, counterfactuals that produce an intense drift towards the \texttt{Female} class in the audited classifier insert long blonde hair into the anchor image. In this experiment, these patterns are related with the constraints imposed by the dataset. However, differently from the previous experiment, the produced counterfactuals are not exiting the data domain over which the model was trained, but are rather exploiting biases existing in the data. Most counterfactuals seen in the front have blonde hair, glasses or a smile pose, whether alone or combined. 
\begin{figure}[!ht]
	\centering
	\includegraphics[width=0.8\columnwidth]{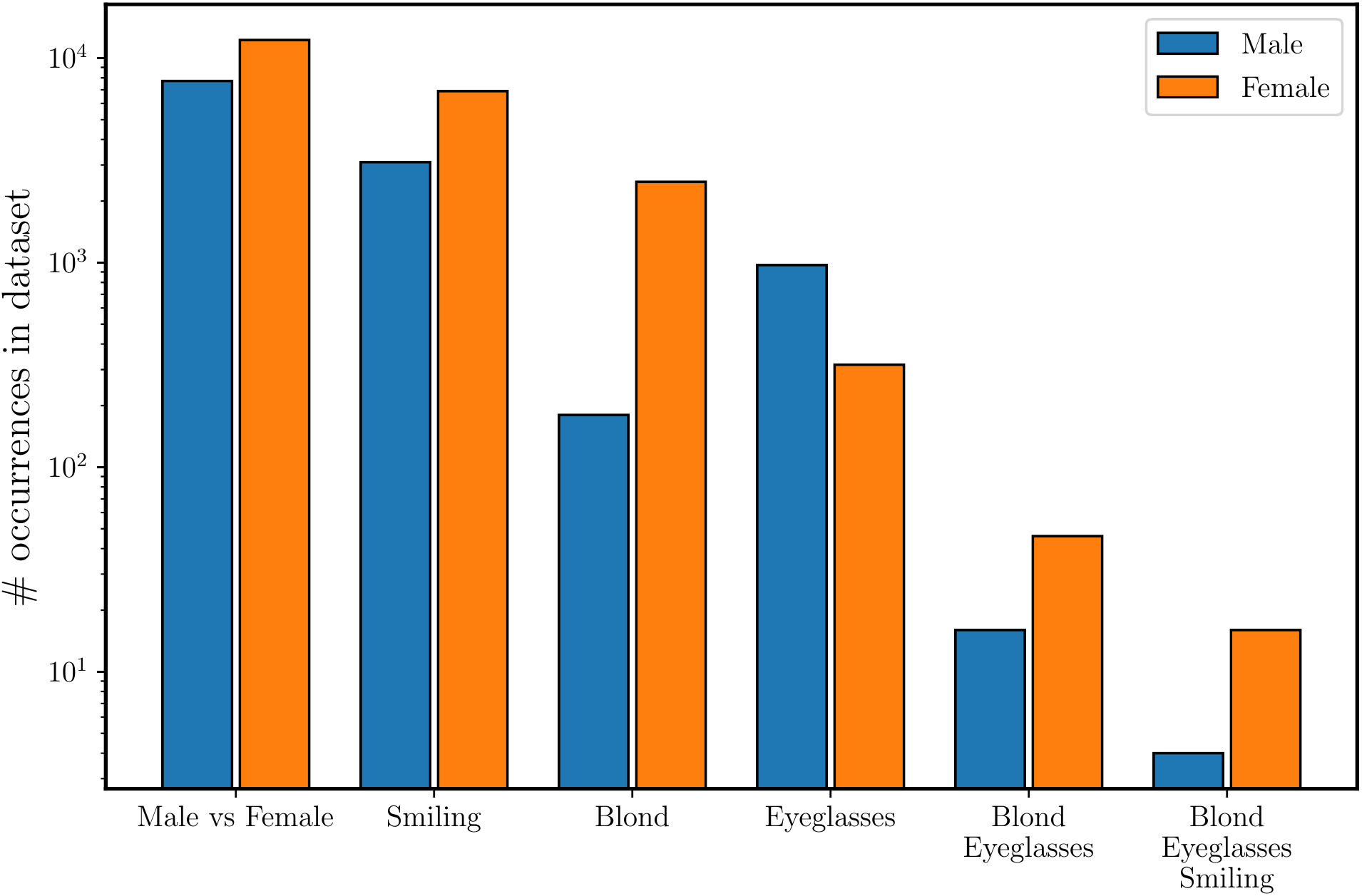}
	\caption{Diagram showing the occurrence within the training dataset of the audited model of different feature combinations, split between \texttt{male} and \texttt{female} examples.}
	\label{fig:AttGAN_ana}
\end{figure}

In order to explore the reason for such a recurring set of counterfactual features, Figure \ref{fig:AttGAN_ana} depicts bar diagrams showing the differences in terms of occurrence over the training examples of different combinations of the three attributes, differentiating between counts corresponding to the \texttt{male} and \texttt{female} classes. It is straightforward to note that the majority of examples featuring any of the combinations of these three attributes belong to the \texttt{female} class. Given a face, if it contains those three attributes, it is quite probably a female. This conclusion is also supported by the fact that the proportion of \texttt{male} instances wearing eyeglasses is notably higher than that of \texttt{female} examples; however, when considering eyeglasses together with the other two attributes, the proportion changes in favor of \texttt{female} examples. This is the reason why, among the counterfactual instances shown in Figure \ref{fig:attgan_results}.c, those using \emph{Eyeglasses} to turn actor Viggo Mortensen into a woman imply changing his hair color to blonde and modifying his expression to include an open smile. In summary: counterfactual instances can help unveil biases in the training data that otherwise could pass unnoticed and could affect the generalization of the target model.

\subsection{Experiment \#3: BigGAN-based counterfactual generation for auditing an \texttt{ImageNet} classifier}

This third experiment is devised to exemplify that the proposed framework can be used to produce counterfactuals in complex tasks comprising a higher number of classes. To this end, as has been mentioned in Section \ref{sec:experiments} we resort to a ResNet18 classifier trained over \texttt{ImageNet}. For the sake of brevity, we will discuss on the set of counterfactual examples generated for a anchor image belonging to class \texttt{Fiddler Crab}, using class \texttt{pajama, pyjama, pj's, jammies} (hereafter, \texttt{Pyjama}) as the target label driving the adversarial modifications imprinted to the anchor image.
\begin{figure}[!h]
	\centering
	\subfigure[]{\includegraphics[width=0.34\columnwidth]{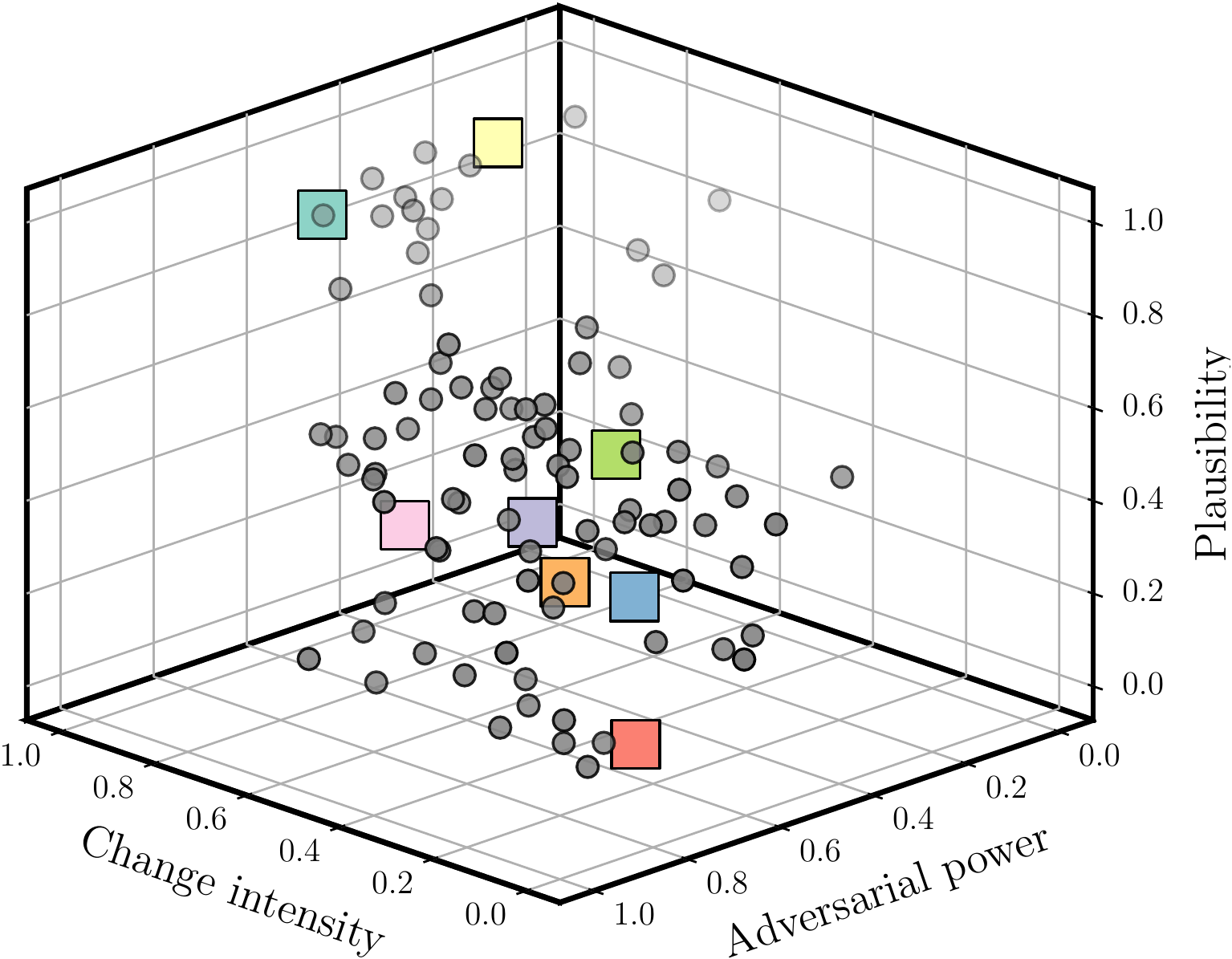}}
	\subfigure[]{\includegraphics[width=0.34\columnwidth]{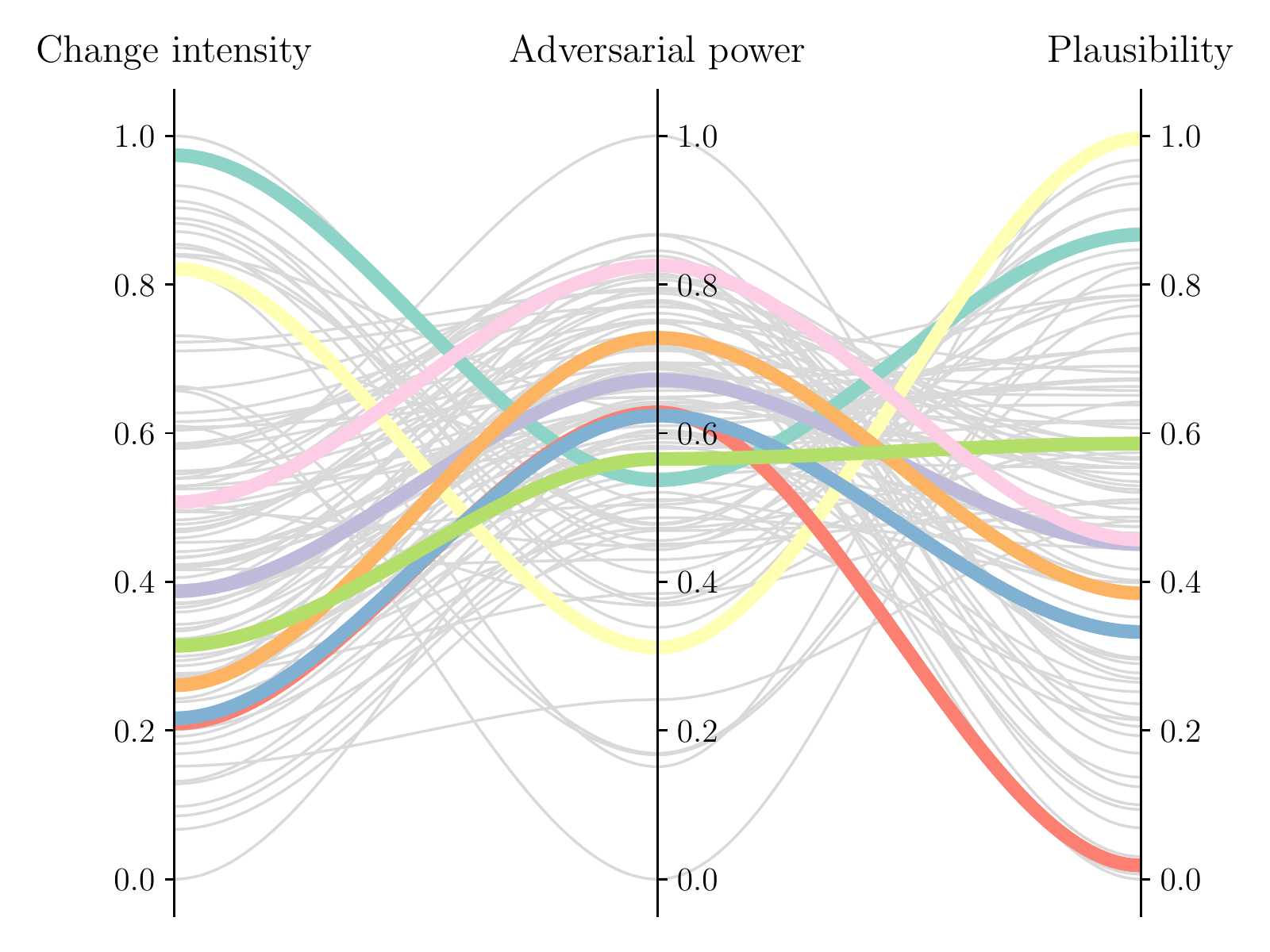}}
	\subfigure[]{\includegraphics[width=0.3\columnwidth]{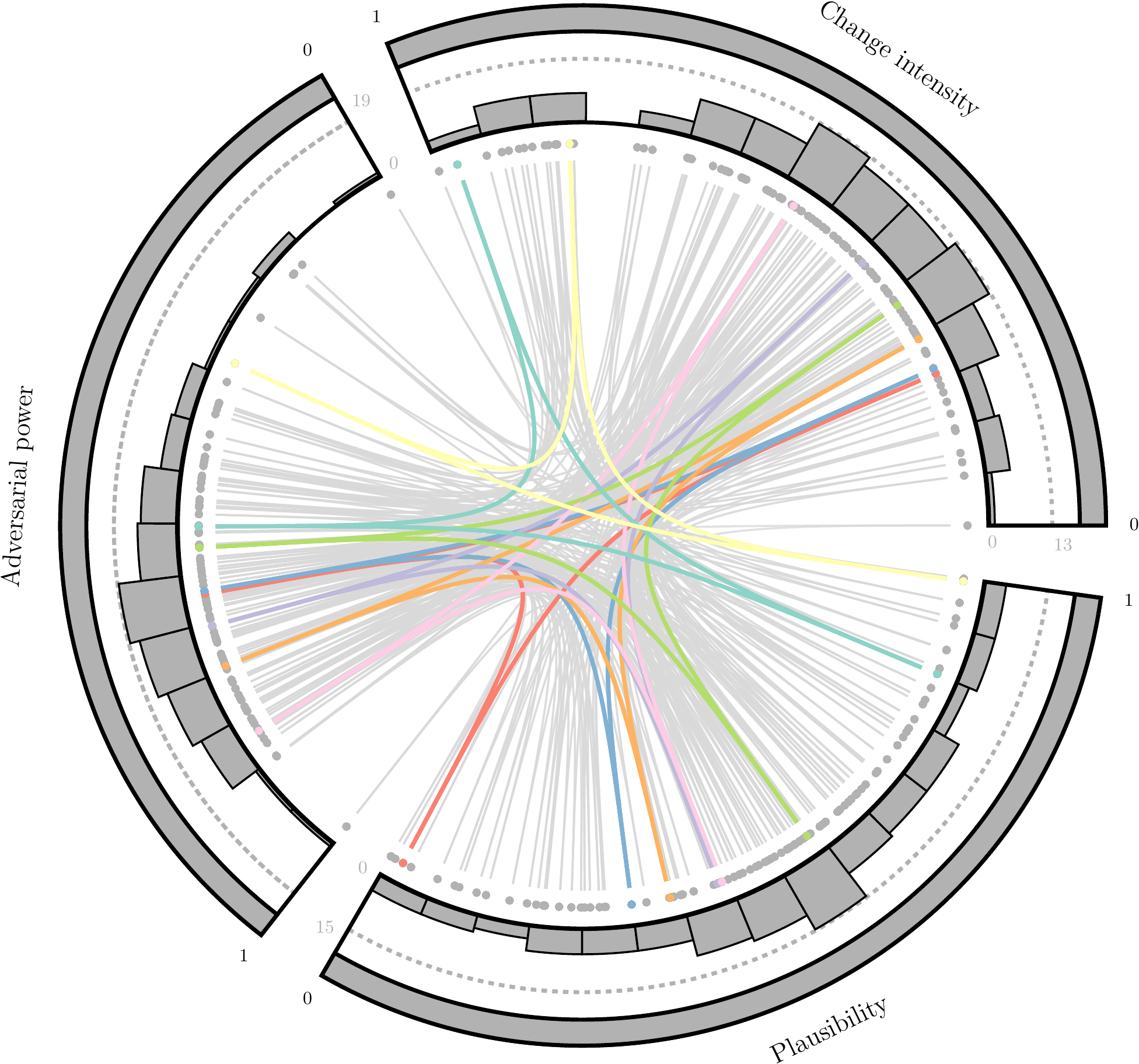}}
	\subfigure[]{\includegraphics[width=0.6\columnwidth]{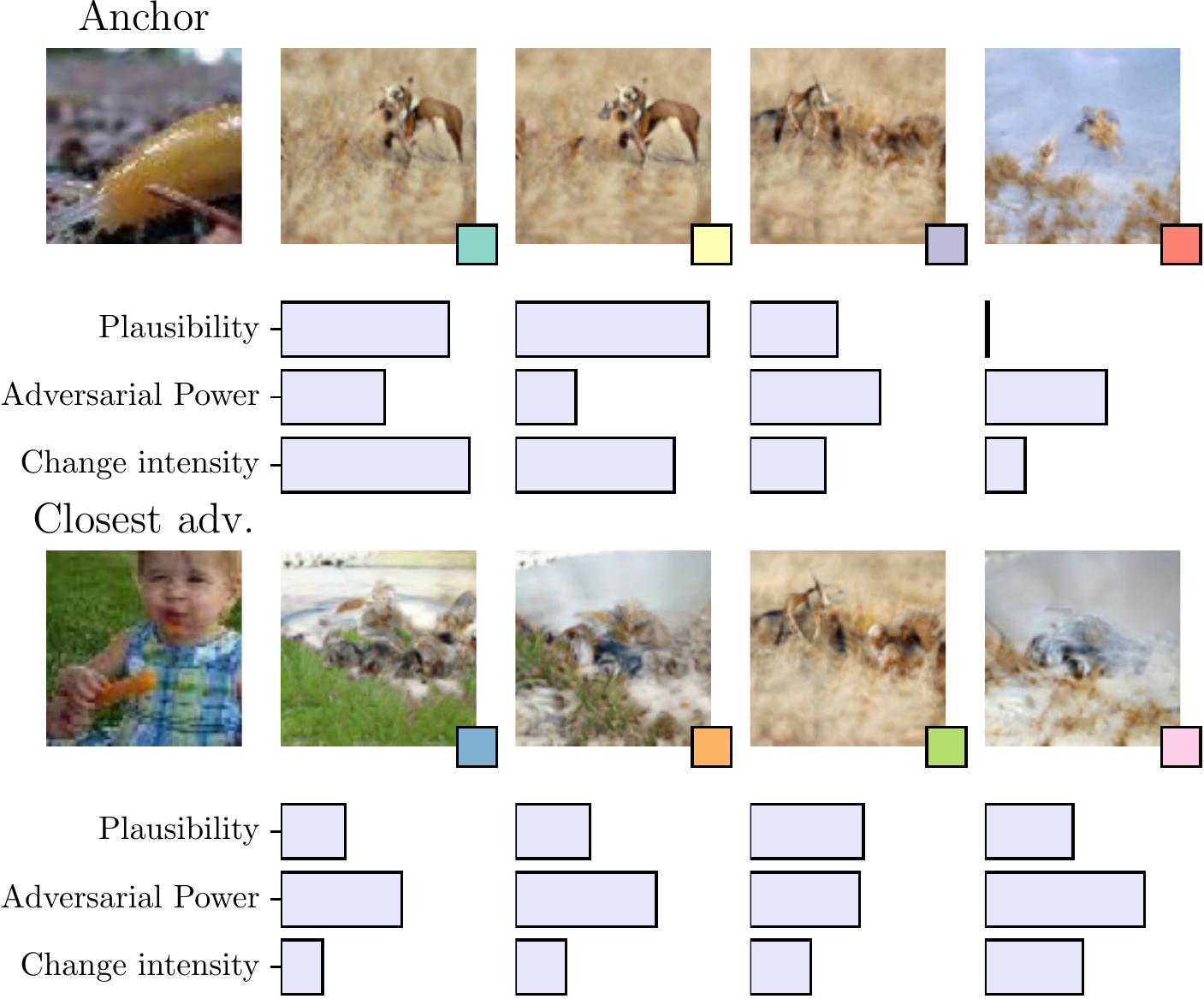}}
	\caption{(a) Pareto front; (b) parallel lines visualization; (c) chord plot; and (d) images of the counterfactual examples generated for a \texttt{Fiddler Crab} example by the proposed framework configured with a BigGAN model.}
	\label{fig:biggan_results}
\end{figure}

Results obtained for this third experiment are collected and shown in Figures \ref{fig:biggan_results}.a to \ref{fig:biggan_results}.d. It must be first noted that one could intuitively expect that the plausible transformation of an image belonging to class \texttt{Fiddler Crab} into a counterfactual belonging to class \texttt{Pyjama} should be difficult to achieve, due to the visual differences that naturally arise between both classes. Thereby, if plausibility is kept as one of the objectives driving the evolution of counterfactuals, \emph{color information} should play an important role in adversarial power, whereas \emph{shape information} would conversely act as a guarantee of plausibility. This is indeed what can be observed in the results. On one hand, the Pareto front does not contain counterfactuals that achieve a large adversarial power and are clearly plausible, evincing the large separation between the two classes. On the other hand, the visualization of the counterfactual images in Figure \ref{fig:biggan_results}.d aligns with the aforementioned intuition: highly plausible counterfactuals retain some image artifacts (e.g., crab legs) that are typical of the class of the anchor image, whereas color information matches that of the closest adversarial of the target label \texttt{Pyjama}. This is also supported by the fact that in unlikely counterfactuals (e.g., \tikzrectangle[fill=myred]{0pt}, \tikzrectangle[fill=myblu]{0pt}, \tikzrectangle[fill=myora]{0pt} and \tikzrectangle[fill=mypink]{0pt}) the only visual aspect that ties the image content to the target label is color information (green, blue and light brown matching the colors present in the closest adversarial), without any discernible image information that could improve the plausibility of the counterfactual.
\vspace{4mm}

\subsection{Experiment \#4: ShapeHDGAN-based counterfactual generation for auditing a \texttt{Chair} versus \texttt{Xbox} voxel classifier}

This fourth case of the devised set of experiments comprises an audited classifier that discriminates whether the voxel at its input is a \texttt{chair} or a \texttt{Xbox}. Therefore, it operates over three-dimensional data, increasing the complexity to qualitatively evaluate the produced counterfactuals with respect to previous experiments. 
\begin{figure}[!h]
	\centering
	\subfigure[]{\includegraphics[width=0.34\columnwidth]{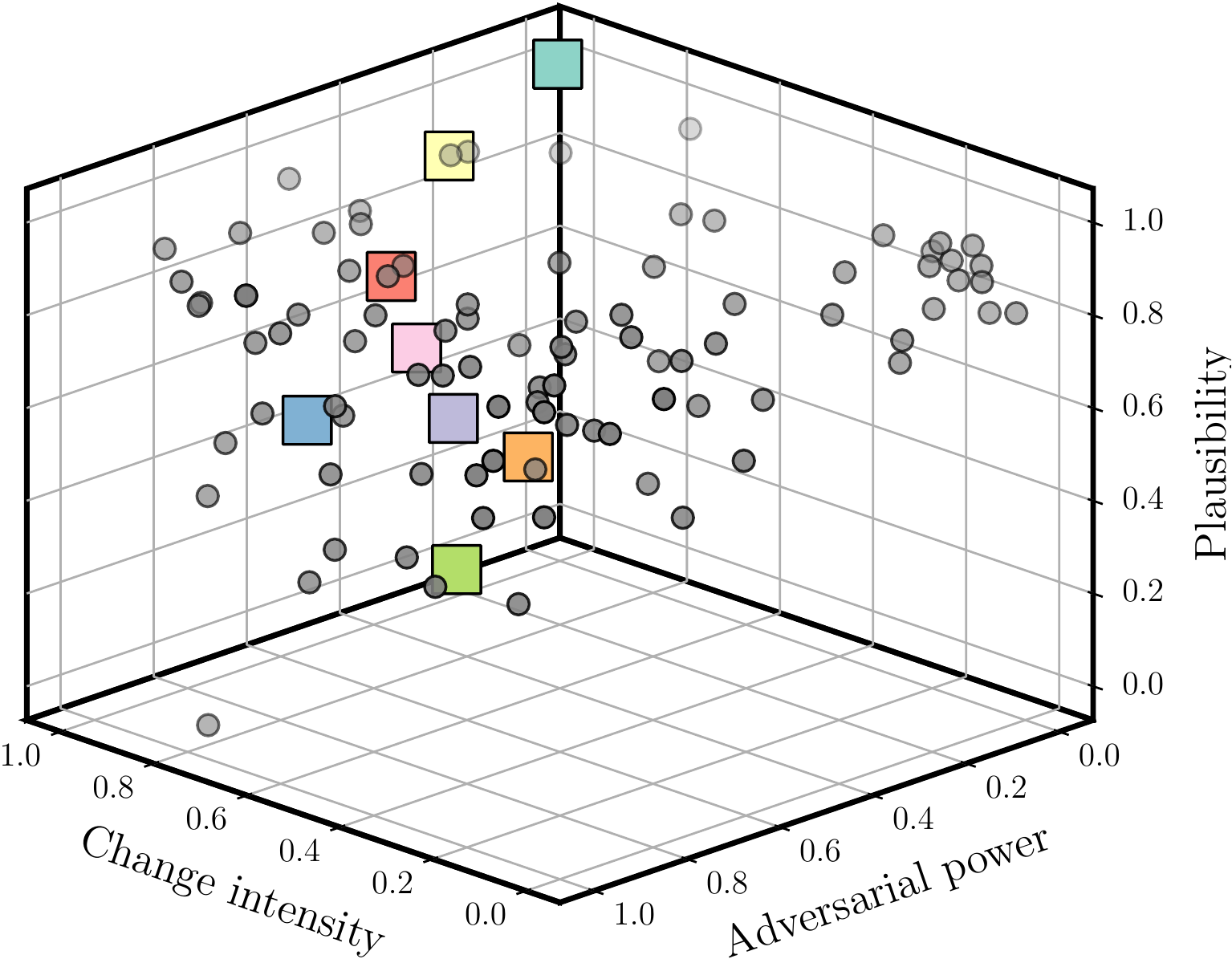}}
	\subfigure[]{\includegraphics[width=0.34\columnwidth]{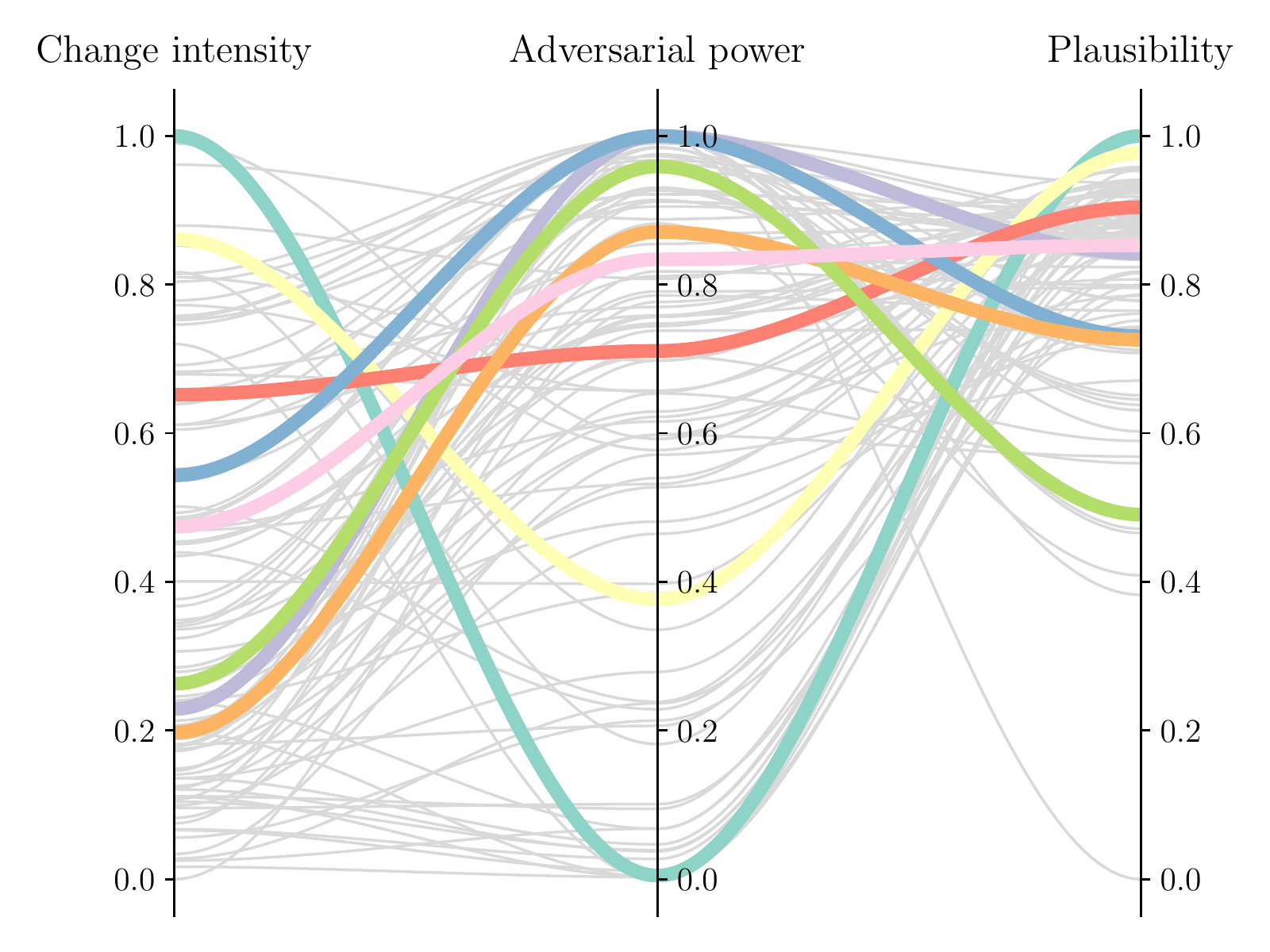}}
	\subfigure[]{\includegraphics[width=0.3\columnwidth]{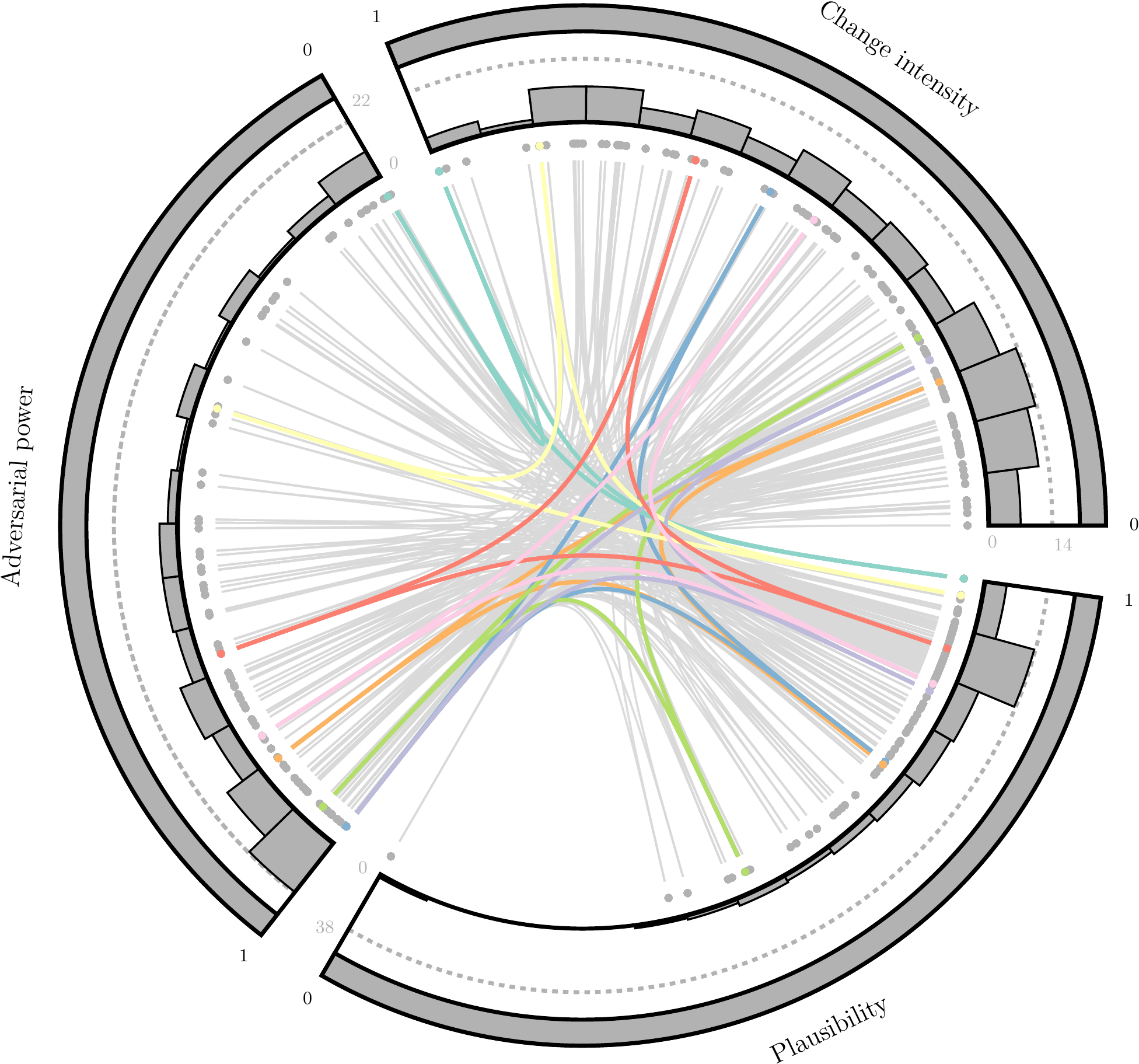}}
	\subfigure[]{\includegraphics[width=0.6\columnwidth]{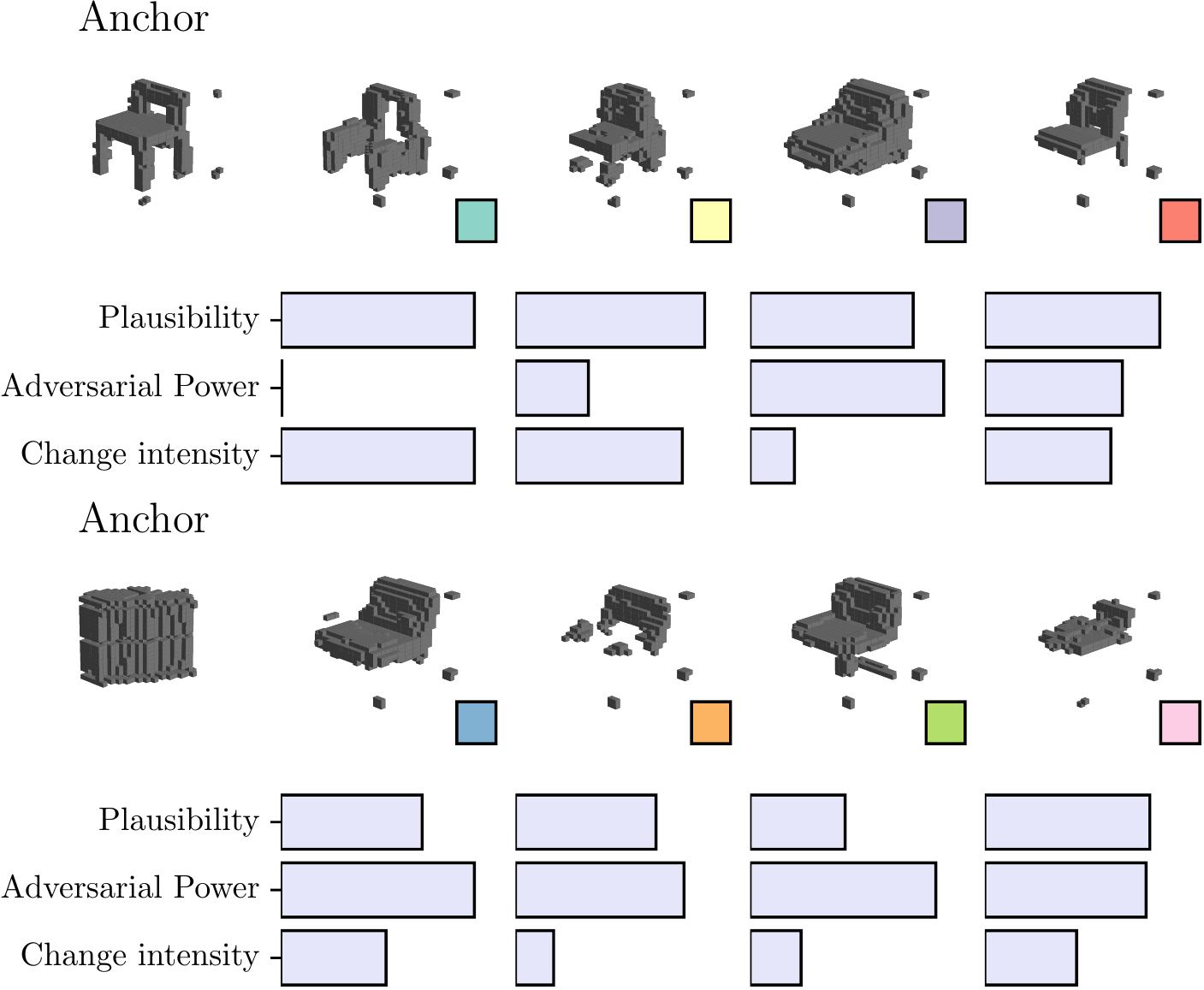}}
	\caption{(a) Pareto front; (b) parallel lines visualization; (c) chord plot; and (d) images of the counterfactual examples generated for a \texttt{chair} example by the proposed framework configured with a Shape3DGAN model.}
	\label{fig:3dgan_results}
\end{figure}

The results elicited for a \texttt{chair} target instance $\mathbf{x}^{\mathbf{a},\oplus}$ are shown in Figures \ref{fig:3dgan_results}.a to \ref{fig:3dgan_results}.d. A first inspection of the counterfactual voxels highlighted in the approximated Pareto front suggests that it is hard to analyze what the audited classifier observes in these inputs to get fooled and predict a \texttt{Xbox}. It appears that a more dense middle part is capable of misleading the classifier. Voxels being generated by the framework resemble a chair, but possess a clearly more dense middle part. It is quite revealing to see how a chair and an xbox can be of any resemblance. Interestingly, a concern to bring up here is that of scale. These voxels (both the generated ones and the dataset over which the ShapeHDGAN model was trained) are normalized, which in turn means that the size of the object has been lost. Scaling can be an interesting option to improve the resolution of small objects. In this case, however, it can be the reason to make this classifier prone to error. 
\begin{figure}[!ht]
	\centering
	\includegraphics[width=0.9\columnwidth]{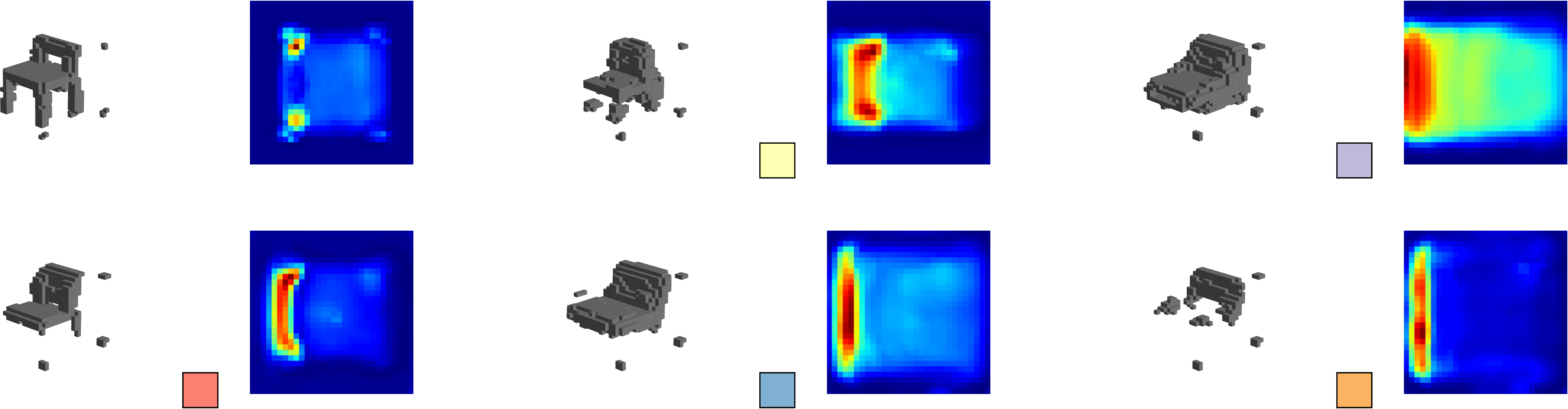}
	\caption{Local explanations (heatmaps via Grad-CAM++) corresponding to the anchor voxel (leftmost pair of images) and two of the counterfactuals depicted in Figure \ref{fig:3dgan_results}.}
	\label{fig:3DGAN_ana}
\end{figure}

This last statement can be buttressed by analyzing which structural parts of the counterfactual voxel are of highest importance for the audited classifier to produce its prediction. This can be done by resorting to gradient-based local post-hoc explanation methods such as Grad-CAM++ \cite{chattopadhay2018grad}. As can be seen in the examples depicted in this figure, most of the observational focus of the model is placed on the vertical rectangular part of the \texttt{chair}, which conforms to intuition given that the actual shape of a \texttt{Xbox} is rectangular. Therefore, counterfactuals for a \texttt{chair} instance wherein the vertical part (\emph{backrest}) is reinforced can bias the audited model without jeopardizing their plausibility.

\subsection{Experiment \#5: StyleGAN2-based counterfactual generation for auditing a \texttt{Cathedral} versus \texttt{Office} classifier}

\begin{figure}[!h]
	\centering
	\subfigure[]{\includegraphics[width=0.34\columnwidth]{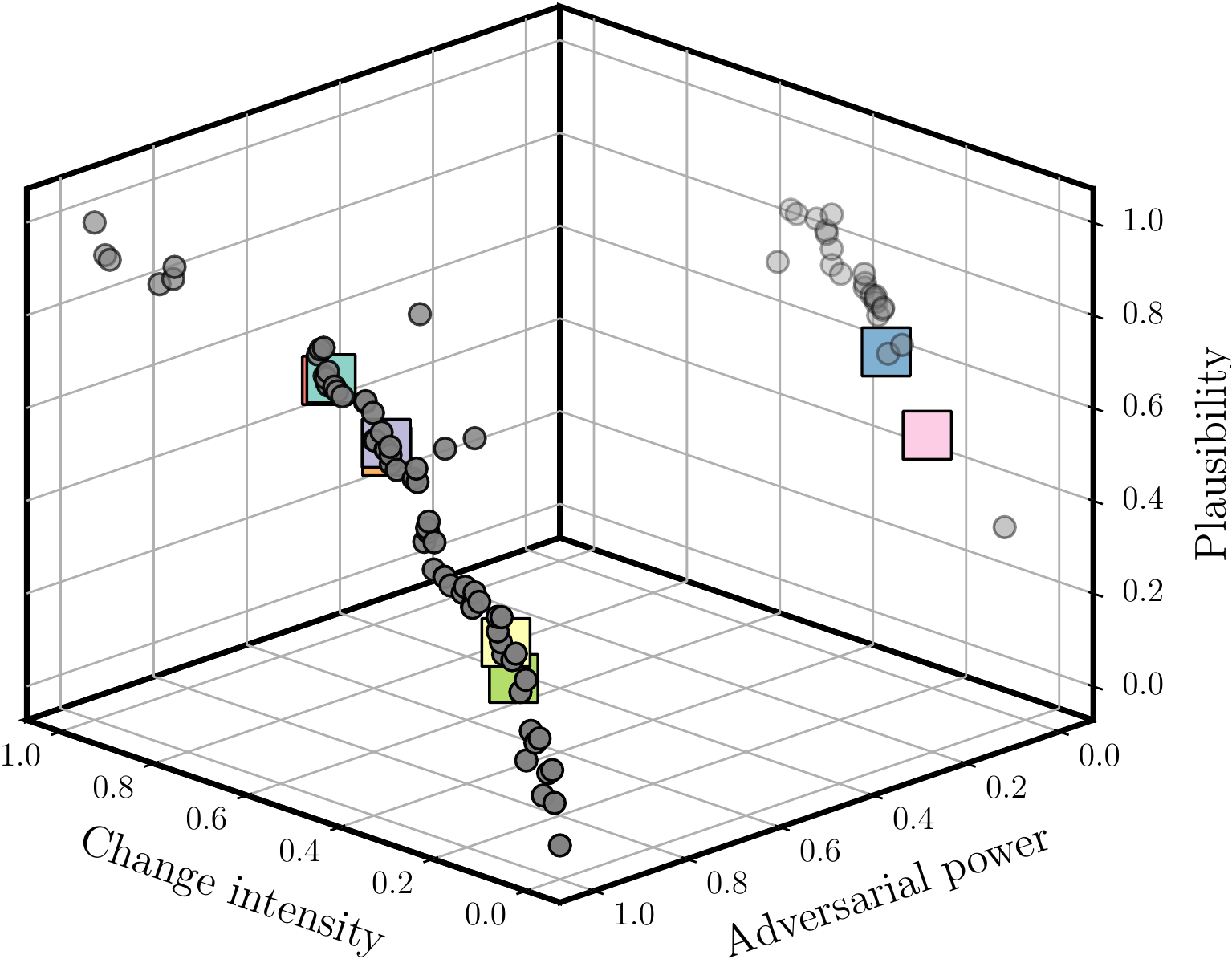}}
	\subfigure[]{\includegraphics[width=0.34\columnwidth]{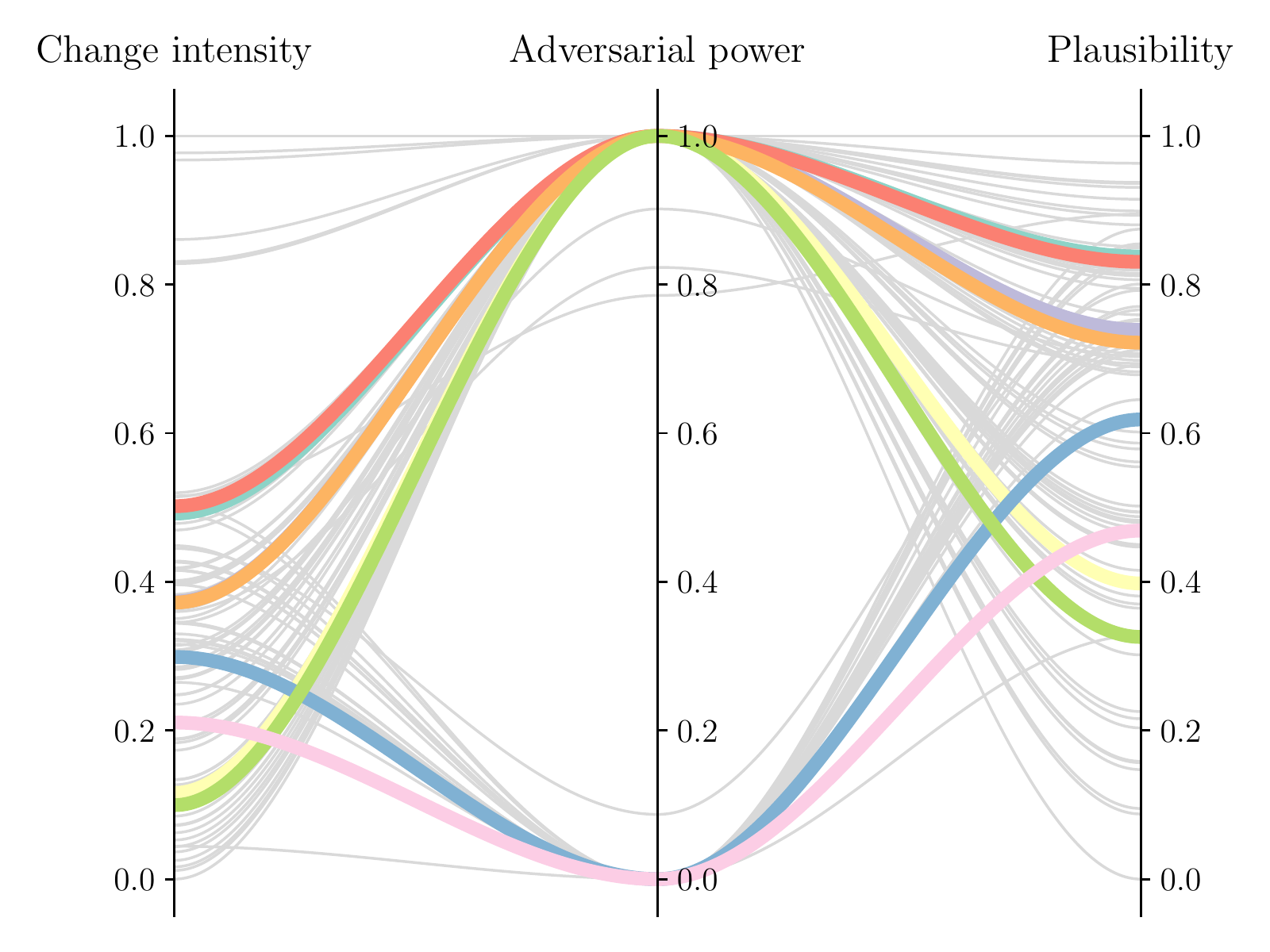}}
	\subfigure[]{\includegraphics[width=0.3\columnwidth]{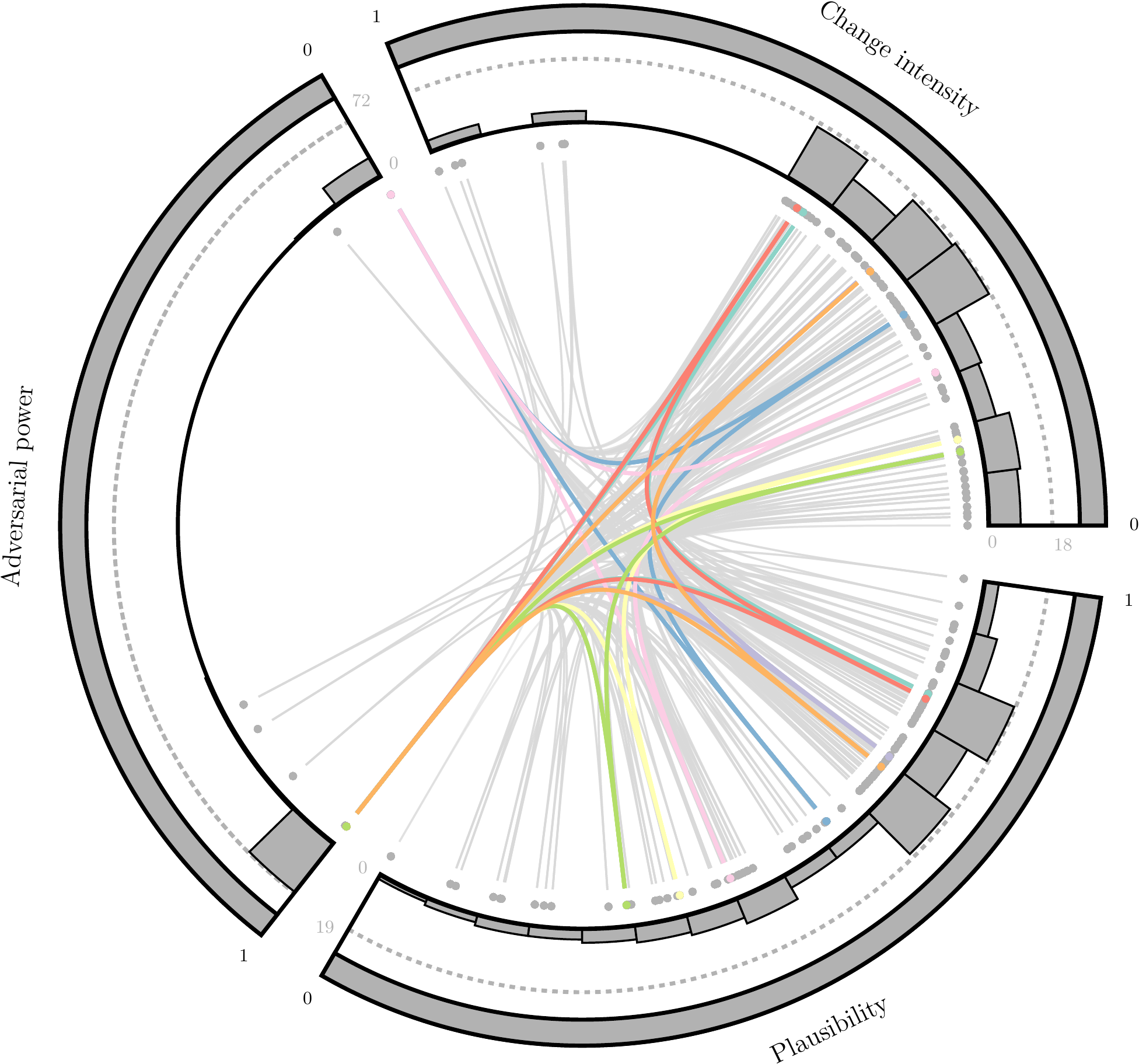}}
	\subfigure[]{\includegraphics[width=0.6\columnwidth]{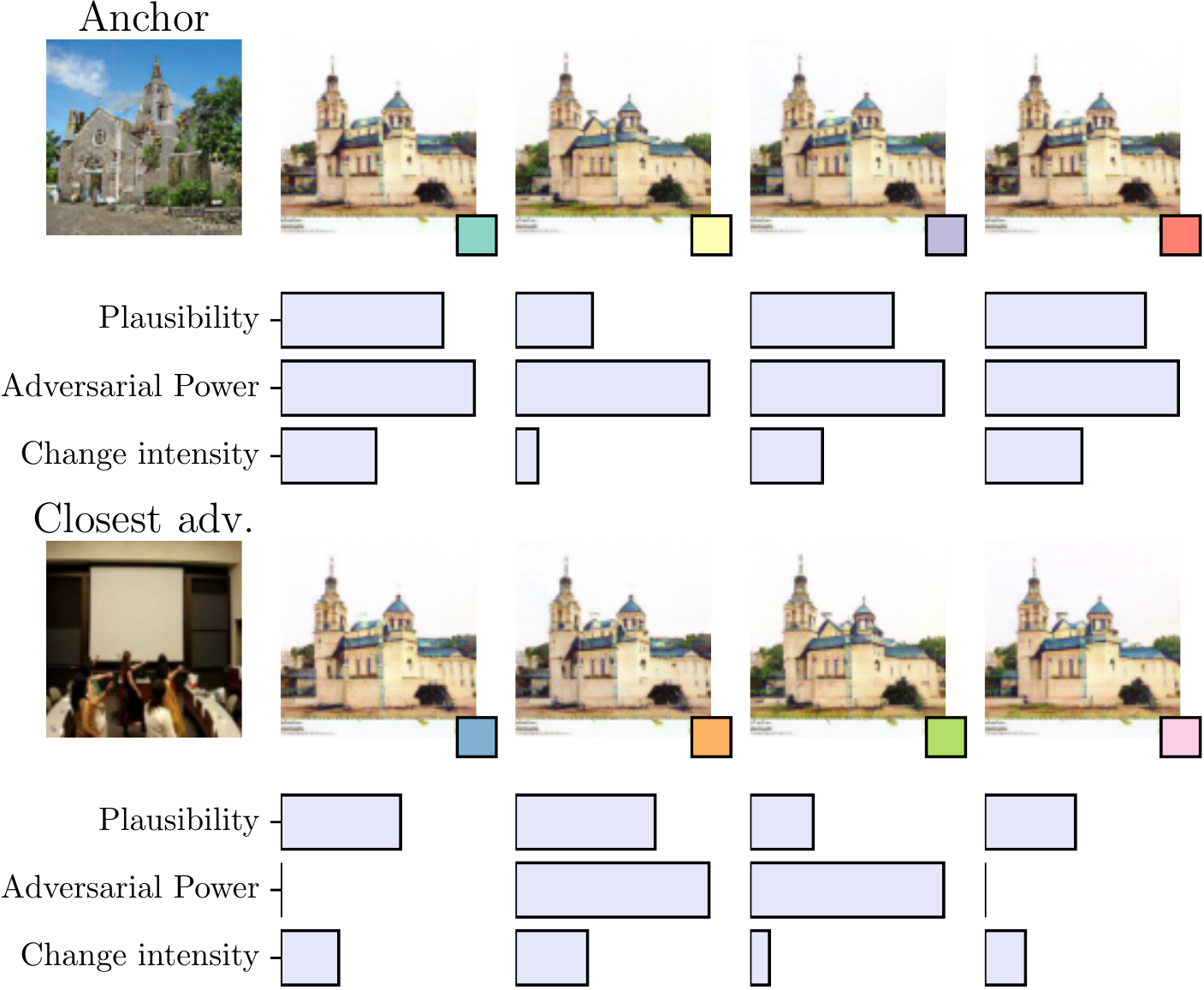}}
	\caption{(a) Pareto front; (b) parallel lines visualization; (c) chord plot; and (d) images of the counterfactual examples generated for a \texttt{Cathedral} example by the proposed framework configured with a StyleGAN2 model.}
	\label{fig:stylegan_results}
\end{figure}

The results of this experiment (Figures \ref{fig:stylegan_results}.a to \ref{fig:stylegan_results}.d) unveil a link between the luminance of the overall counterfactual image and its ability to mislead the model. However, in this time the spread of counterfactuals over the adversarial power dimension of the Pareto front is notably lower than in the previous experiments. If the results are compared with those of Figure \ref{fig:bicyclegan_results}, in this case the missing spread observed in the front is validated with what can be visually discerned in the counterfactuals given by the framework. 

\begin{figure}[!ht]
	\centering
	\includegraphics[width=\columnwidth]{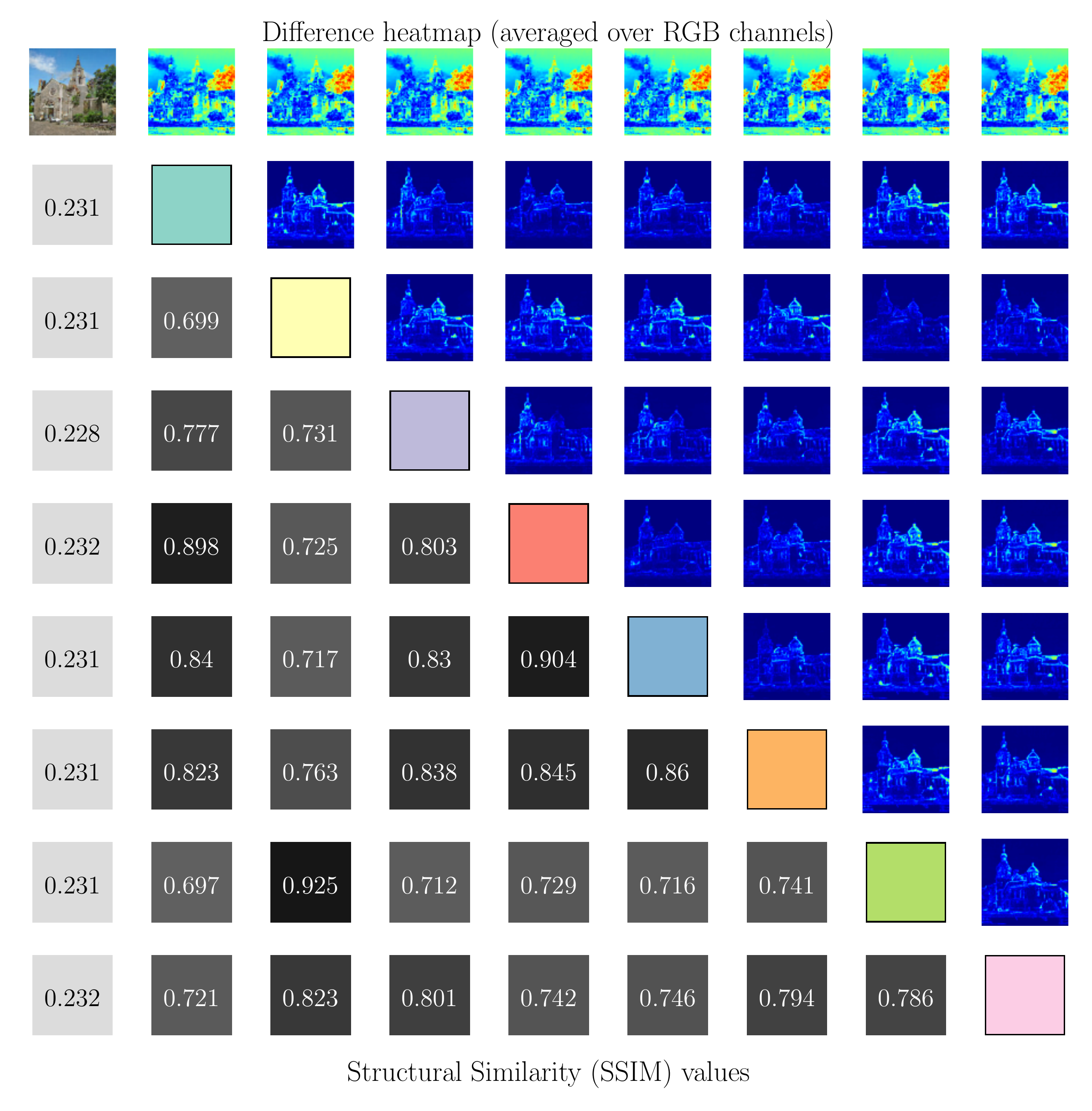}
	\caption{Comparison between the original counterfactuals and the anchor image following the colored markers shown previously in Figure \ref{fig:attgan_results}: the upper triangular part of the matrix is composed by heatmaps depicting the mean absolute difference of the RGB pixels of every pair of images in comparison, whereas the lower triangular part denotes the SSIM value quantitatively reflecting the similarity between the images.}
	\label{fig:stylegan_ana}
\end{figure}

Following this last observation, we focus on the analysis of the visual differences between the anchor image and the produced counterfactuals shown in Figure \ref{fig:stylegan_ana}. Specifically, the plot shows the heatmap of mean absolute differences (averaged over the RGB channels) and the SSIM (Structural Similarity Index Measure \cite{wang2004image}) among the original anchor image and its counterfactual version. As can be inferred from the visuals in the first row of the figure, the framework is exploiting a burned-out background with minor structural changes in the image. This seems reasonable with the spread found in the front: it shows that these changes in the background of the image can completely fool the model, but there are not changes that would account for a well spread front since the structural differences among both classes are large. Furthermore, \texttt{cathedral} instances undergo a misrepresentation bias in the dataset, in the sense that none of the \texttt{cathedral} training examples has a totally overcast sky. This suggests that whitening the background of the image may grant a chance for the counterfactual to mislead the audited classifier, yet without any guarantee for success given the scarce similarity between images belonging to both classes.

\subsection{Experiment \#6: CGAN-based counterfactual generation for auditing a \texttt{MNIST} classifier}

In correspondence to Q3, this last experiment is devised to elucidate whether the output of the proposed framework can be used for any other purpose than the explainability of the target model. To this end, we run and assess visually the counterfactuals generated for the digit classification task defined over the well-known \texttt{MNIST} dataset. The characterization of every class defined in this dataset is done by a na\"ive conditional GAN. 
\begin{figure}[!h]
	\centering
	\subfigure[]{\includegraphics[width=0.34\columnwidth]{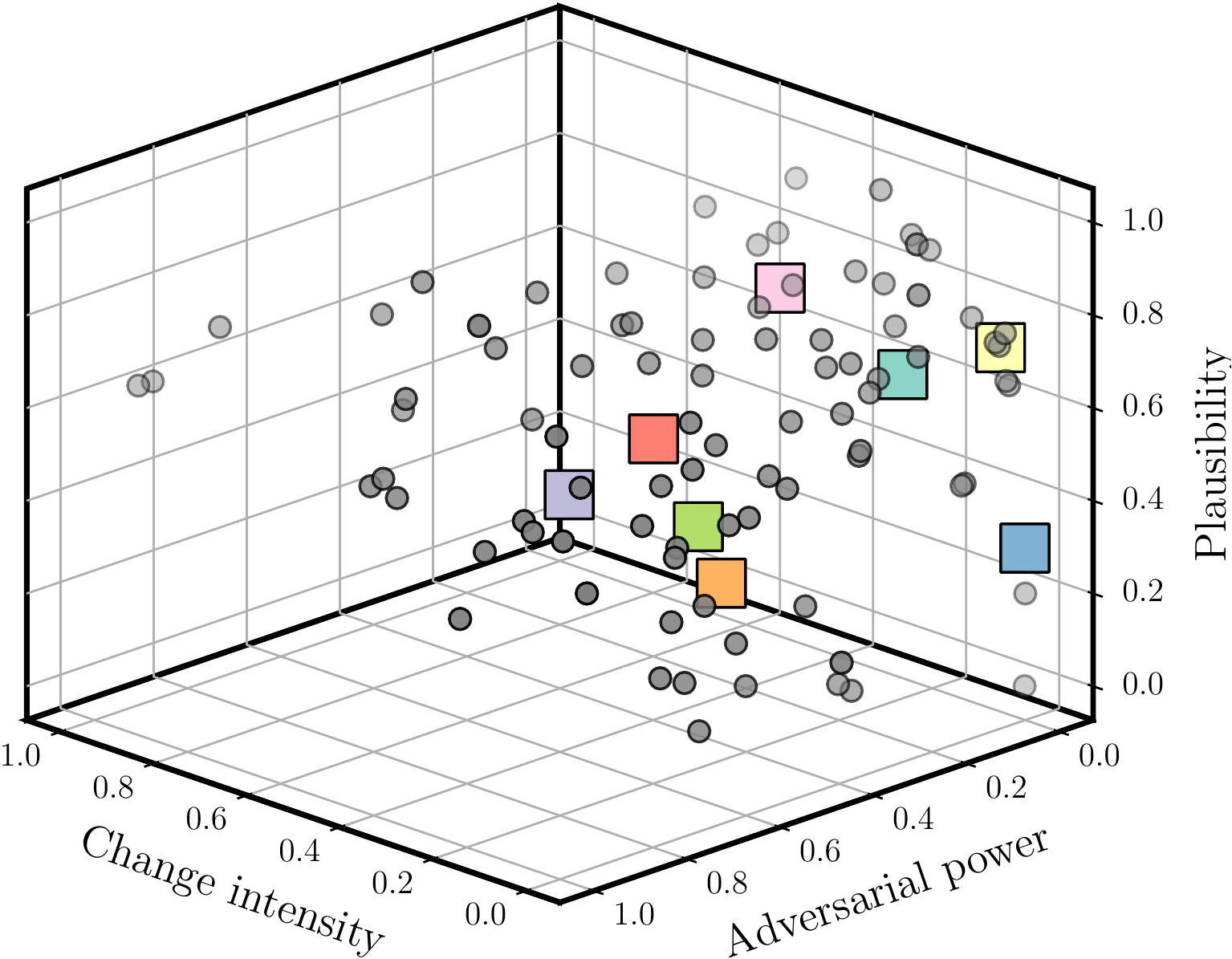}}
	\subfigure[]{\includegraphics[width=0.34\columnwidth]{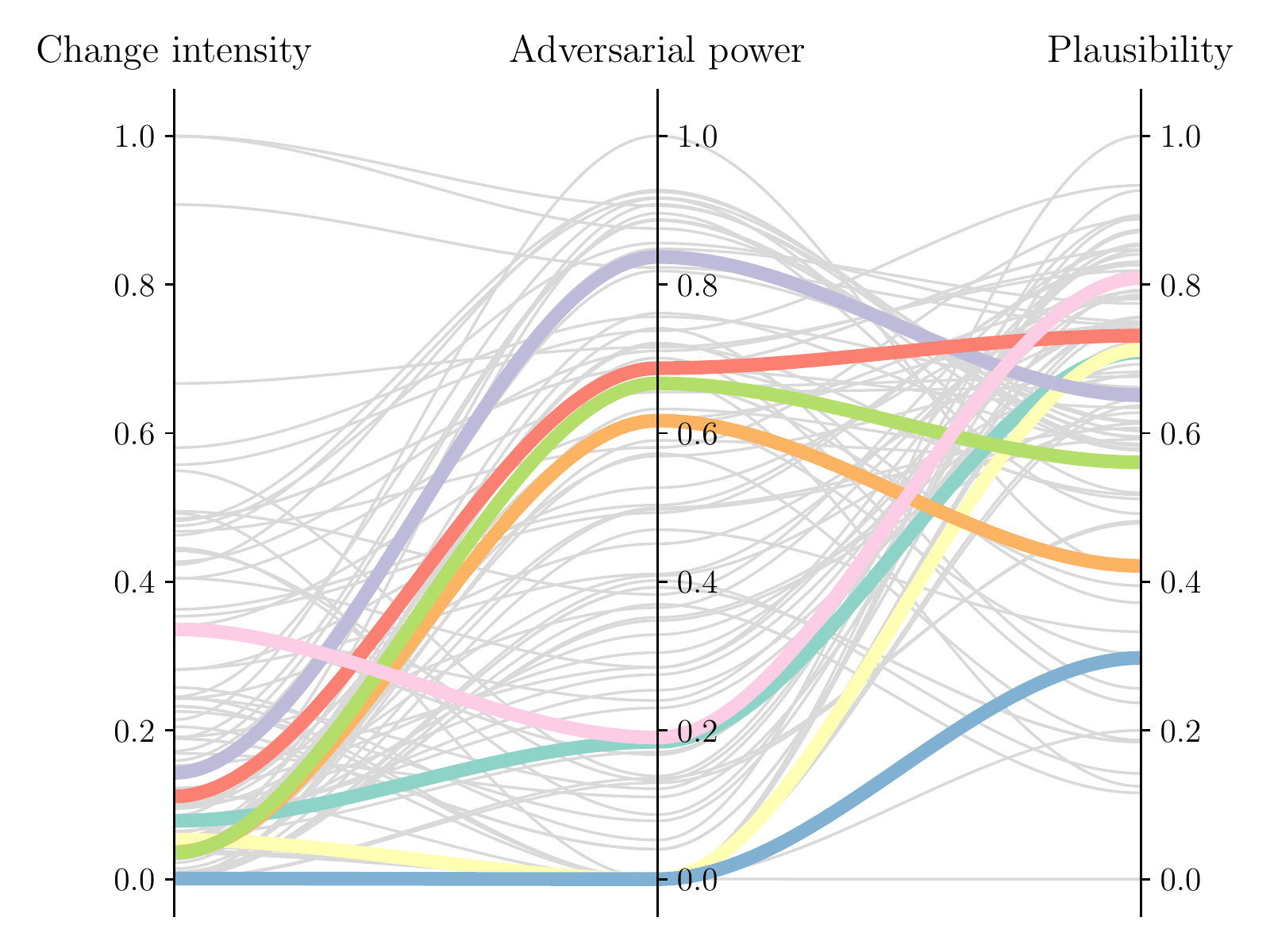}}
	\subfigure[]{\includegraphics[width=0.3\columnwidth]{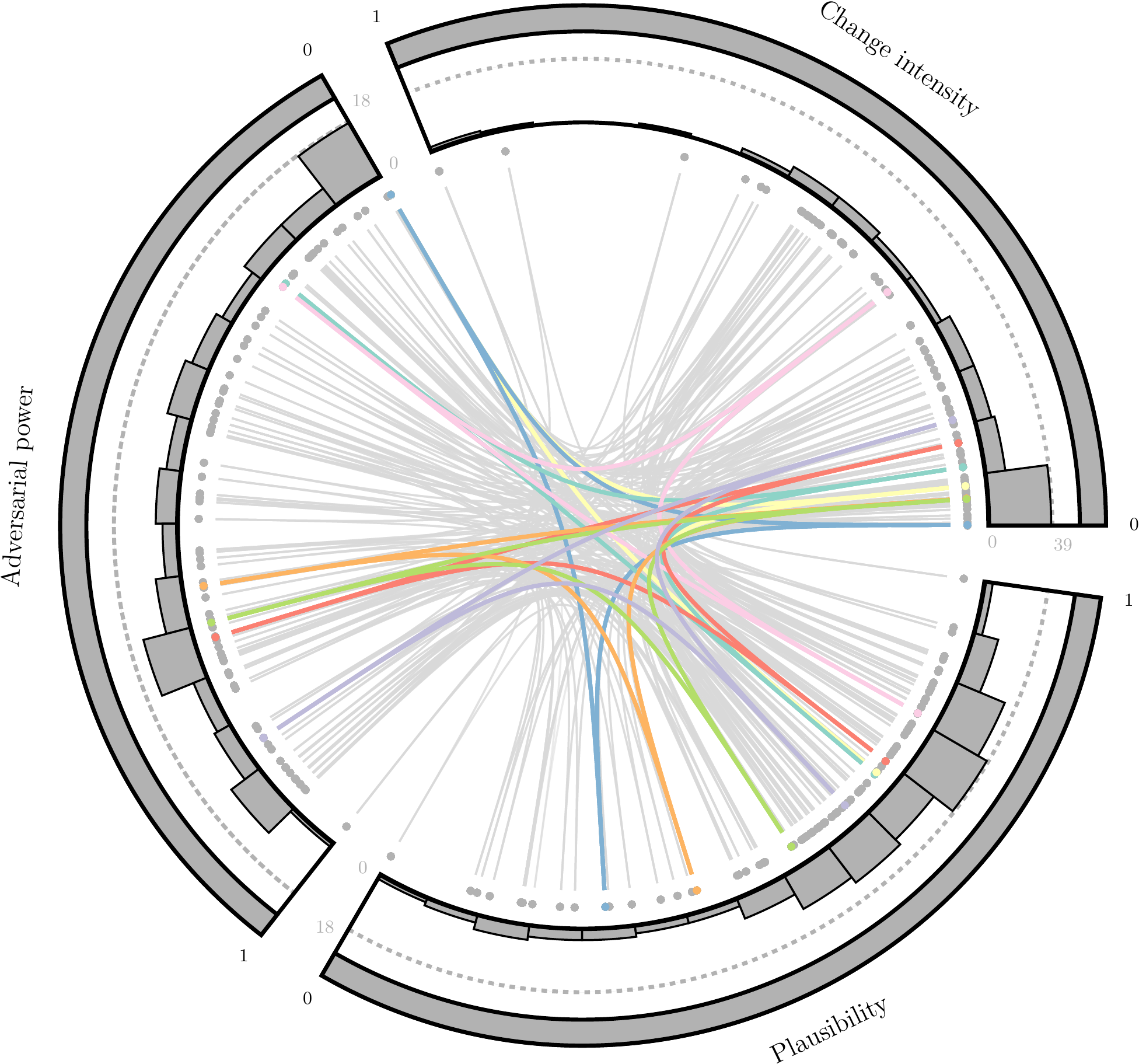}}
	\subfigure[]{\includegraphics[width=0.6\columnwidth]{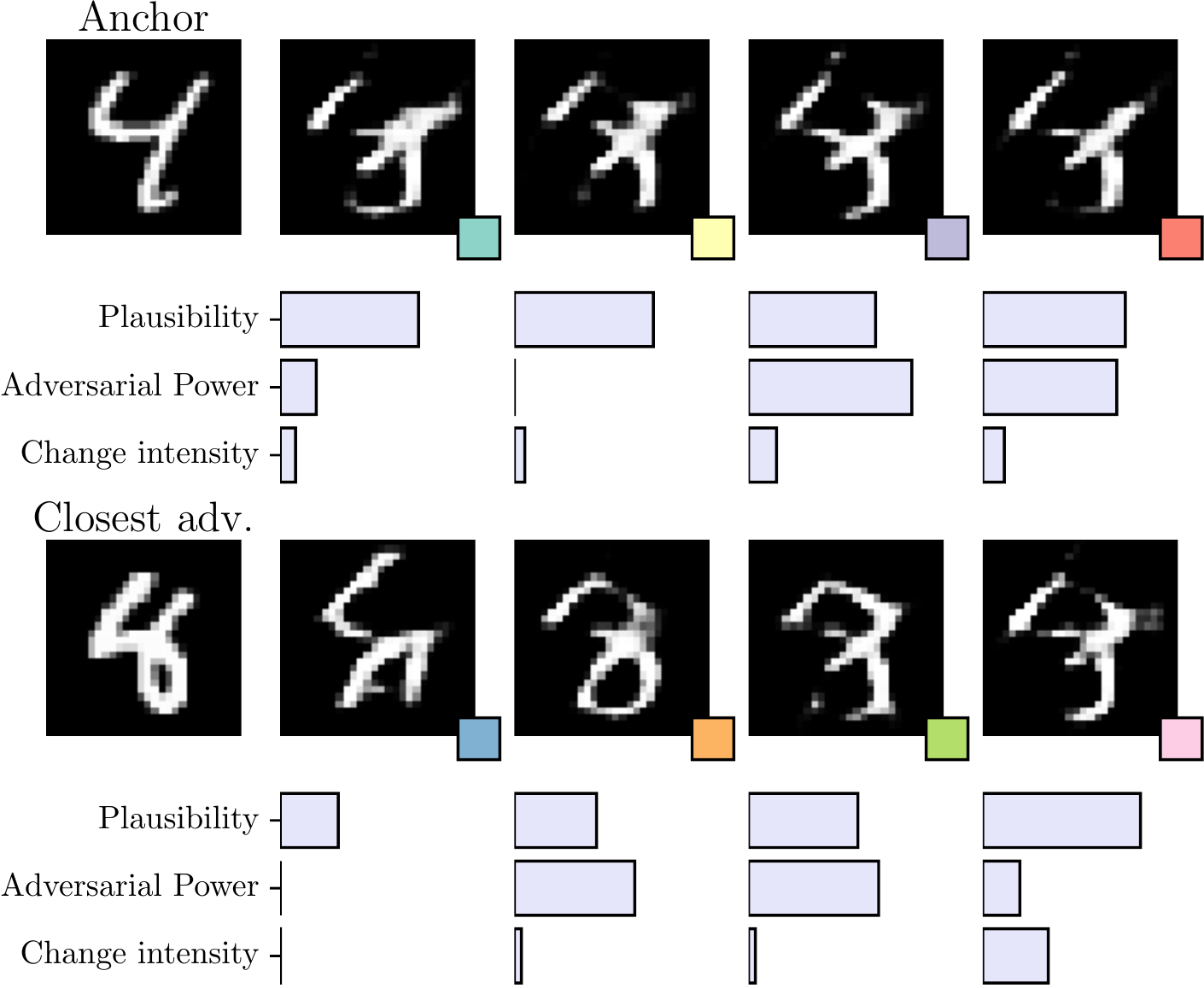}}
	\caption{(a) Pareto front; (b) parallel lines visualization; (c) chord plot; and (d) images of the counterfactual examples generated for an MNIST digit classification model by the proposed framework configured with a CGAN model.}
	\label{fig:cgan_results}
\end{figure}

Figures \ref{fig:cgan_results}.a to \ref{fig:cgan_results}.d portray the output of the framework when generating counterfactuals for an anchor image $\mathbf{x}^{\mathbf{a},\oplus}$ corresponding to digit 4. From what can be observed in the samples extracted from the front, visual information corresponding to digits 4 and 8 appear to be interfering with the capability of the audited model to discriminate among them. This intuition is buttressed by the fact that the closest element is a sample corresponding to digit 8, as displayed in the first bottom image of Figure \ref{fig:cgan_results}.d. Indeed, once again misrepresented visual artifacts in the dataset are opening a path to generate plausible counterfactuals, since most instances generated by the framework are digits with incomplete shapes. This may come from the fact that the MNIST dataset is mostly composed by digits that are correctly written.  

\begin{figure*}[!ht]
	\vspace{-1mm}
	\centering
	\hspace{2.15cm}
	\subfigure[]{%
		\includegraphics[width=0.45\columnwidth]{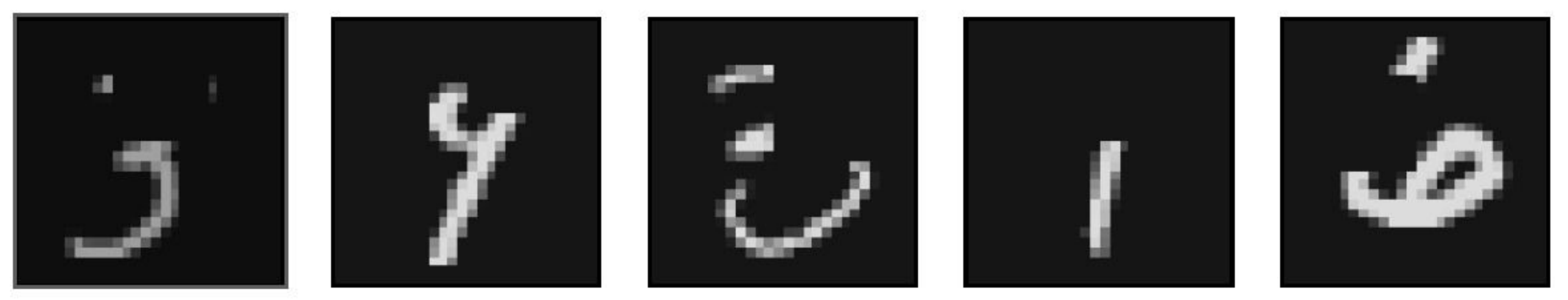}
		}
	\newline
		\subfigure[]{\includegraphics[width=0.34\columnwidth]{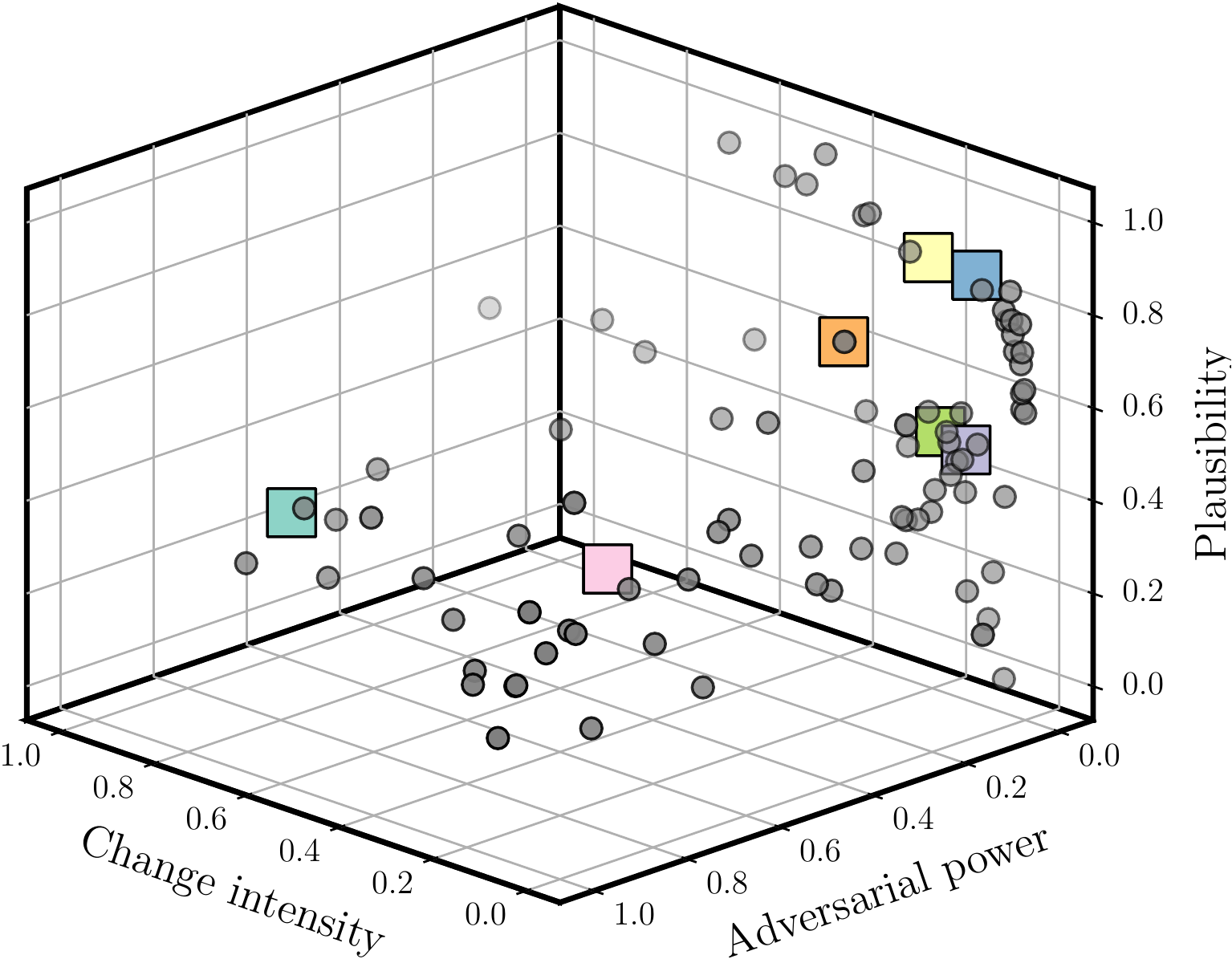}}
		\subfigure[]{\includegraphics[width=0.34\columnwidth]{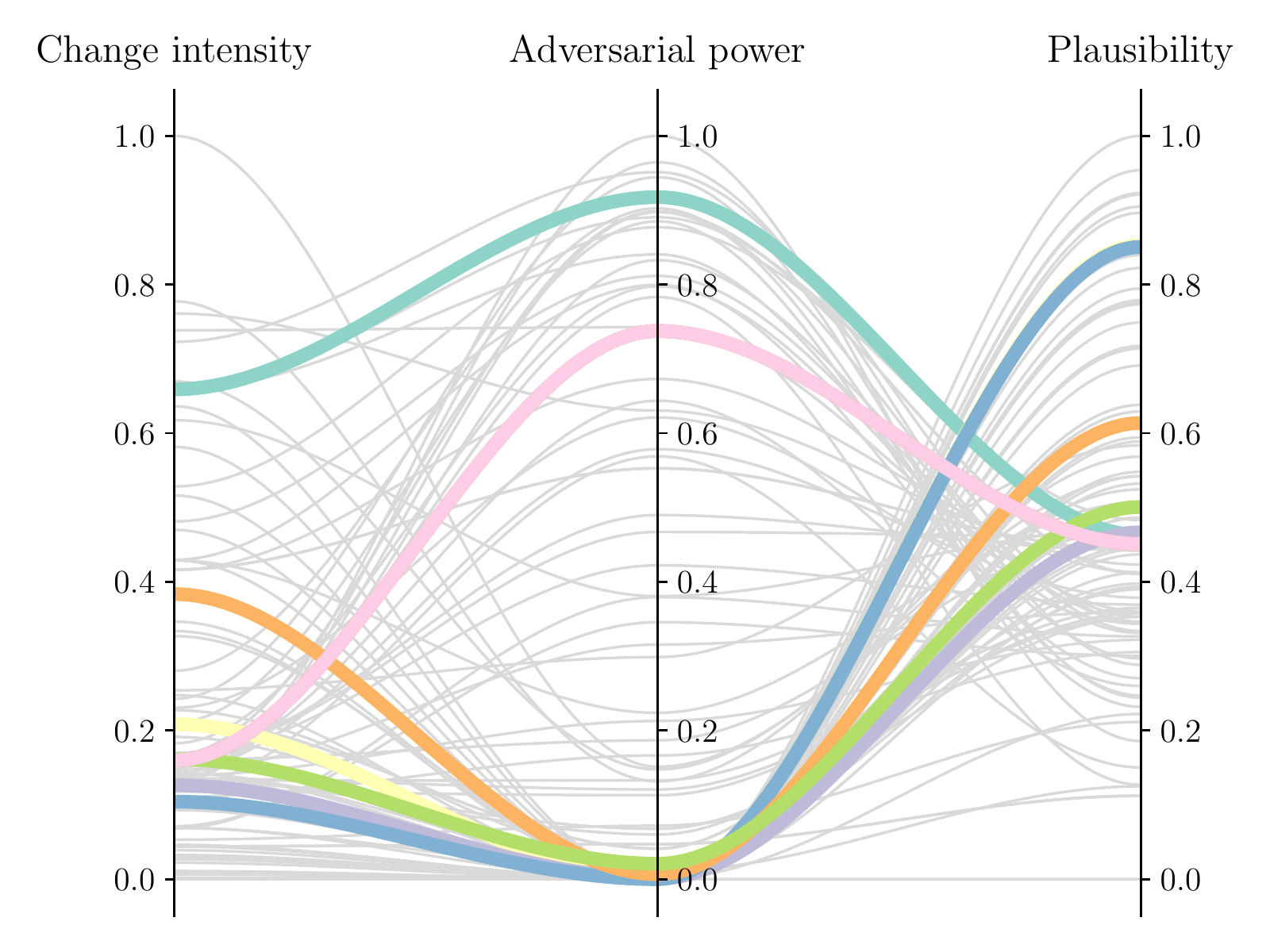}}
		\subfigure[]{\includegraphics[width=0.3\columnwidth]{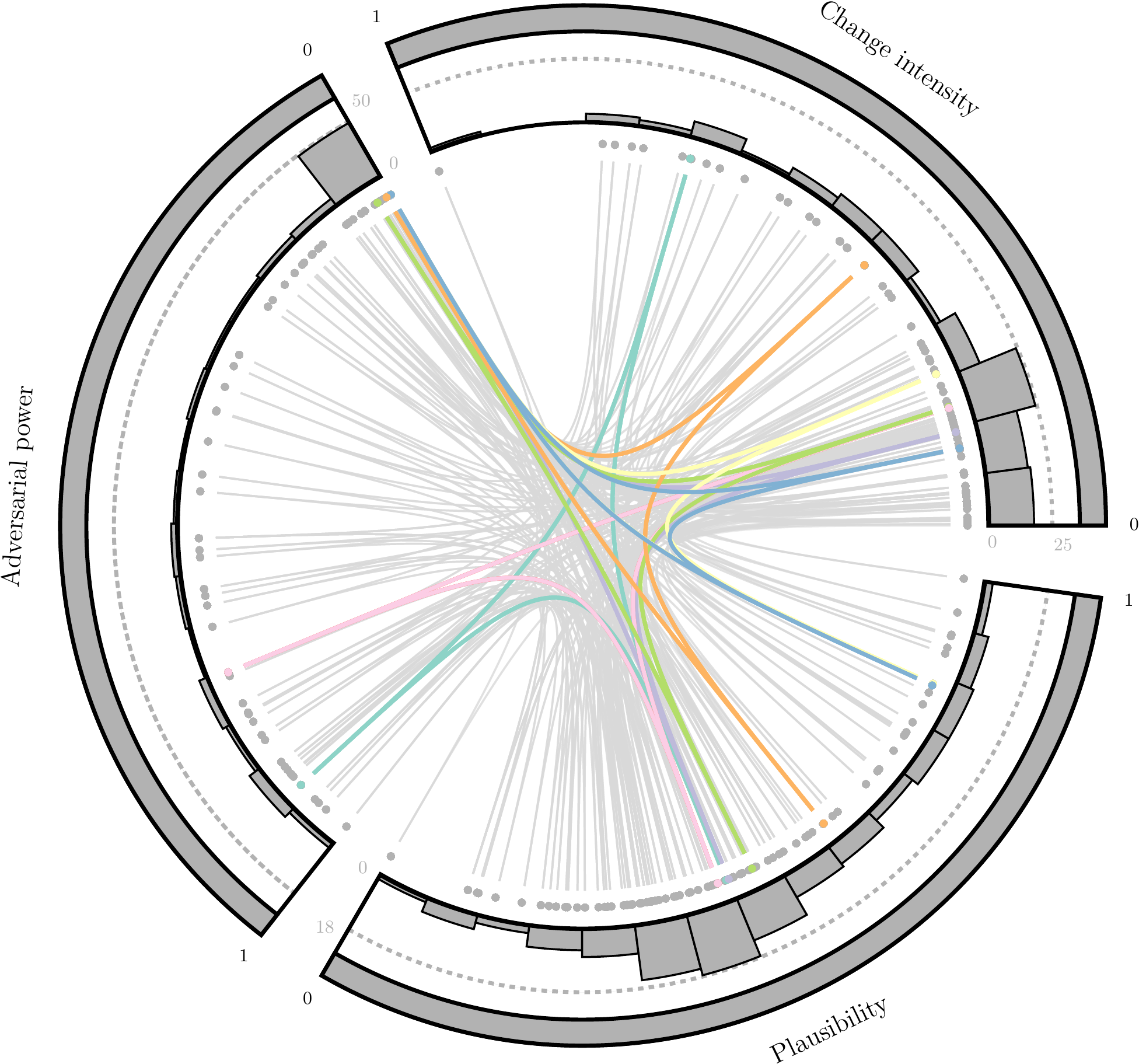}}
		\subfigure[]{\includegraphics[width=0.6\columnwidth]{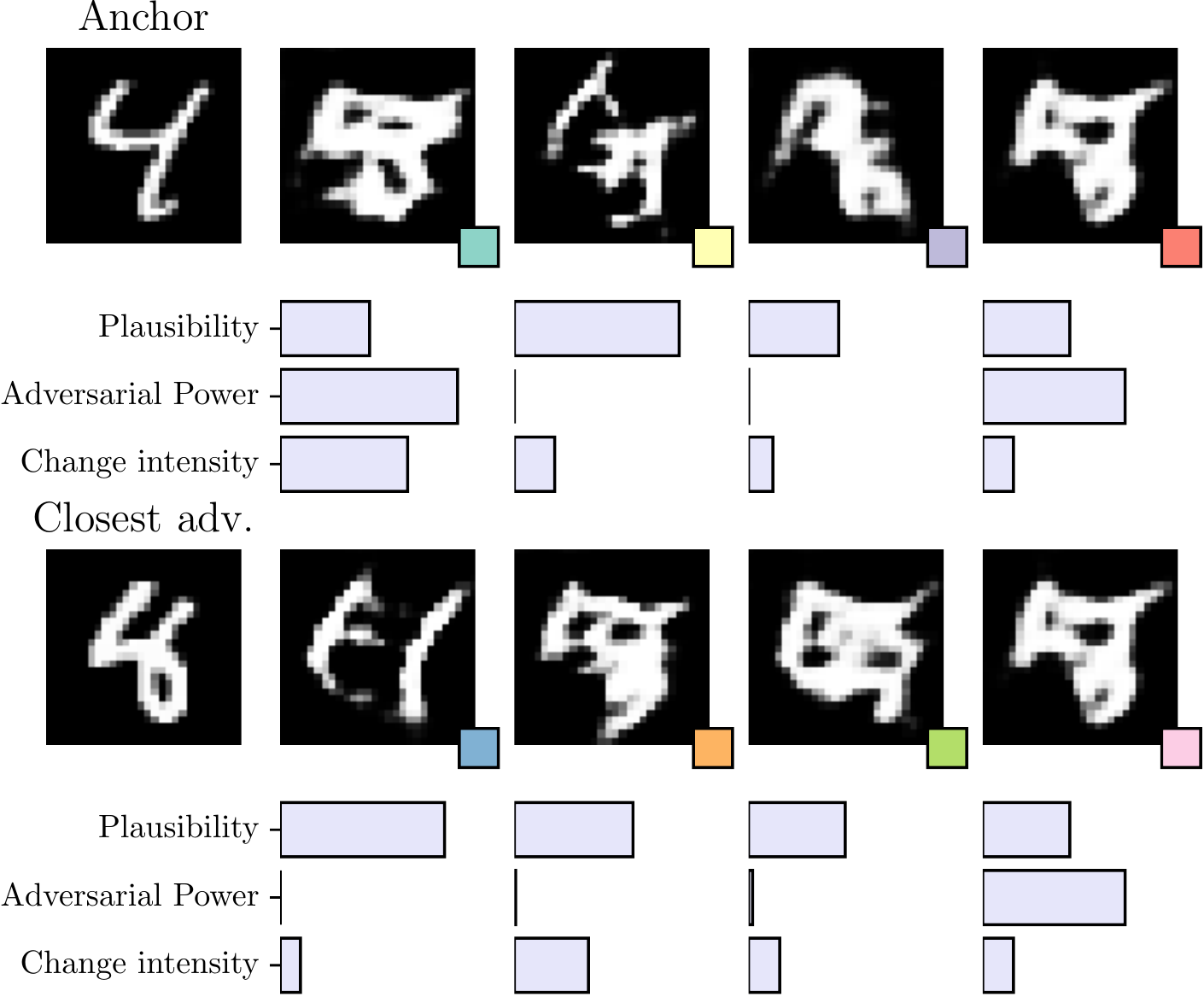}}
	\caption{(a) Sample of the unfinished digits generated for supplementing the MNIST dataset as an additional class. These digits are designed to cover the gap found in the initial phase of experiment \#5, which showed a weakness of the MNIST dataset concerning non-finished digits; (b, c, d and e) output of the framework when auditing the same model trained over the augmented dataset. In this case, images of the produced counterfactual instances do not conform to what human thinking expects to be a digit.}
	\label{fig:cgan_results_ana}
	\vspace{-1mm}
\end{figure*}

We prove the converse to this statement by running again the experiment with an additionally inserted class in the dataset that contains digits of every class over which a part has been erased. This narrows the opportunities for the framework to generate counterfactuals by erasing selected shape fragments of the anchor digit. This is confirmed in Figures \ref{fig:cgan_results_ana}.a to \ref{fig:cgan_results_ana}.d, which depict the output of the framework in this alternative setting: the counterfactual instances generated can be declared to be plausible with respect to this extended dataset, yet a visual inspection of their corresponding digits concludes that they do not resemble a numerical digit. In summary: the output of our framework can tell which domain over the image (color, shape) can be leveraged to make the audited model more robust against input artifacts.

\section{Conclusion}\label{sec:conclusion}

This manuscript has proposed a novel framework that leverages the generative strength of GAN architectures and the efficient exploration capabilities of multi-objective optimization algorithm to traverse search spaces of large dimensionality. Our research hypothesis departs from the need to inform the user with further information beyond the capability of the counterfactual instances to adversarially change the output of the black-box model under study. Counterfactual generation approaches must ensure that the change \emph{can occur}, and inform the user about the intensity of the change of the generated counterfactual, and the severity by which the output of the model would change should the counterfactual actually occur.

The devised framework builds upon this hypothesis to produce multi-criteria counterfactual explanations for a given input example and a black-box model to be audited. Specifically, a GAN model is used to furnish a generative model that characterizes the distribution of input examples which, together with its discriminator module and its conditional dependence on an attribute vector, synthesizes examples that can be considered \emph{plausible}. The trained GAN is therefore used as a proxy evaluator of the plausibility of new data instances that change the output of the audited model (counterfactuals). Our designed framework seeks the set of counterfactual examples that best balance between \emph{plausibility}, and \emph{adversarial power}, incorporating a third objective (\emph{change intensity}) that may be also in conflict, depending on the dataset at hand. 

Six experiments have been run and discussed to answer three research questions aimed to understand the contribution of the framework to the explainability and understanding of the model being audited. The conclusions drawn with respect to such questions are given below:
\begin{itemize}[leftmargin=*]
    \item \textit{Q1. Is counterfactual generation an optimization problem driven by several objectives?}

	\item[] As evinced by the Pareto front approximations obtained for the six experiments, counterfactual explanations are clearly governed by multiple objectives of relevant importance for the search. Depending on the dataset, some of such objectives could not be conflicting with each other. Nevertheless, the task of finding good counterfactual explanations must be approached as a search comprising different goals for the sake of a more enriched interface for the user of the audited model.
	
    \item \textit{Q2. Do the properties of the generated counterfactual examples conform to general logic for the tasks and datasets at hand?}
    
    \item[] Definitely: our discussion on the results obtained for every experiment we have qualitatively inspected images and voxel volumes corresponding to the produced counterfactual instances. Artifacts observed in such adversarial images not only can be explained departing from common sense as per the task addressed by the audited model (e.g. color variations or emphasized structural parts of the voxels), but also exploit differences and similarities found among the data classes feeding the model at hand.
    
    \item \textit{Q3. Do multi-criteria counterfactual explanations serve for broader purposes than model explainability?}

	\item[] Indeed, counterfactual analysis may contribute to the discovery of hidden biases resulting from misrepresentations in the training dataset of the audited model. Our discussions have empirically identified that counterfactual explanations can reflect such misrepresentations which, depending on the context, can be understood as a hidden compositional (attribute-class) bias or a potential vulnerability for adversarial attacks.
\end{itemize}

On a closing note, the framework presented in this work has showcased that counterfactual explanations must be tackled as a multi-faceted challenge due to the diversity of audiences and profiles for which they are generated. Understanding how a black-box behaves within the prediction boundaries of its feature spaces empowers non-expert users and improves their trust in the model's output. However, an advance use of this explanatory interface should regard other aspects to respond the \emph{so much for how much?} question in counterfactual analysis. This is in essence the ultimate purpose of the framework proposed in this paper, as well as the motivation for future research aimed at easing the interpretation of counterfactual explanations issued by the framework in more complex problems, comprising larger input and/or output dimensionalities (e.g. video classification and multi-modal classification tasks). {\color{black}Improving the understandability of the counterfactuals as per the cognitive feedback of the audience will be also actively investigated, for which mechanisms will be devised to bring the human cognitive skills into the algorithmic loop of the counterfactual generation framework.}

\section*{Ethics and Declaration of Competing Interests}

The authors declare that there are no conflict of interests. This work does not raise any ethical issues.

\section*{Acknowledgements}

This work has received funding support from the Basque Government (\emph{Eusko Jaurlaritza}) through the Consolidated Research Group MATHMODE (IT1294-19) and the ELKARTEK funding program (3KIA project, KK-2020/00049). Parts of this work have been funded by the Austrian Science Fund (FWF), Project: P-32554. N. D\'iaz-Rodr\'iguez is currently supported by the Spanish Government Juan de la Cierva Incorporaci\'on contract (IJC2019-039152-I) and the Google Research Scholar Programme 2021.

\bibliographystyle{model1-num-names}
\bibliography{bib}

\end{document}